\documentclass[twoside]{article}
\usepackage{microtype}
\usepackage{xcolor}
\usepackage{graphicx}
\usepackage[utf8]{inputenc}
\usepackage{latexsym}
\usepackage{hyperref}
\usepackage{bbm}
\usepackage{graphicx}
\usepackage{mathptmx}
\usepackage{amsmath}
\usepackage{amssymb}
\usepackage{float}
\usepackage{amsfonts}
\usepackage{amsthm}
\newtheorem{prop}{Proposition}
\newtheorem{theorem}{Theorem}
\newtheorem{lemma}[theorem]{Lemma}
\newtheorem{remark}{Remark}

\usepackage{comment}
\usepackage{tikz}
\usetikzlibrary{positioning}
\usepackage{lipsum,adjustbox} %
\usepackage{subcaption}
\usepackage{cancel}

\usepackage{float}
\usetikzlibrary{shapes.geometric, arrows}
\usetikzlibrary{calc}
\usepackage{algorithm}
\usepackage{xcolor}
\usepackage{booktabs}
\usepackage{url}
\usepackage{bm}
\usepackage{xcolor}
\usepackage{soul}
\usepackage{rotating}

\usepackage{algpseudocode}
\usepackage{mathtools}

\usepackage[accepted]{aistats2023}
\usepackage[round]{natbib}

\begin{document}

%

%

\twocolumn[

\aistatstitle{\{PF\}$^2$ES: Parallel Feasible Pareto Frontier Entropy Search for Multi-Objective Bayesian Optimization}

\aistatsauthor{ Jixiang Qing \And Henry B. Moss \And  Tom Dhaene \And Ivo Couckuyt  }

\aistatsaddress{ Ghent University -imec\\Ghent, Belgium\\ {Jixiang.Qing@Ugent.be}  \And  Secondmind.ai \And Ghent University -imec\\Ghent, Belgium \And Ghent University -imec\\Ghent, Belgium} ]

\begin{abstract}
We present Parallel Feasible Pareto Frontier Entropy Search ($\{\text{PF}\}^2$ES) --- a novel information-theoretic acquisition function for multi-objective Bayesian optimization supporting unknown constraints and batch query. Due to the complexity of characterizing the mutual information between candidate evaluations and (feasible) Pareto frontiers, existing approaches must either employ crude approximations that significantly hamper their performance or rely on expensive inference schemes that 
  substantially increase the optimization's computational overhead. By instead using a variational lower bound, $\{\text{PF}\}^2$ES provides a low-cost and accurate estimate of the mutual information. 
  We benchmark $\{\text{PF}\}^2$ES against other information-theoretic acquisition functions, demonstrating its competitive performance for optimization across synthetic and real-world design problems. 
\end{abstract}

\section{INTRODUCTION}

The problem of optimizing multiple objectives is common across science, machine learning and industry and is known as \textit{Multi-Objective Optimization} (MOO). Due to the conflicting nature of the multiple objectives, there is not one single solution and we instead seek 
the \textit{Pareto Frontier} ---  a set of optimal solutions that provide a trade-off among different objectives. More precisely, the Pareto frontier consists of all solutions from which one cannot improve the performance of a specific objective without degrading another. For constrained MOO problems, the goal is to find the  \textit{feasible Pareto frontier}, i.e., the Pareto frontier containing only points that satisfy the constraints.

\textit{Multi-Objective Bayesian Optimization} (MOBO) \citep{yang2019efficient, daulton2021parallel, feliot2017bayesian, qing2022robust} is a well-established framework for solving expensive MOO problems with strict evaluation budgets. In order to achieve data-efficiency, BO uses cheap probabilistic \textit{surrogate models} to predict the performance of not-yet-evaluated configurations. Heuristic search strategies, known as \textit{acquisition functions}, then use the posterior belief of this model to direct subsequent objective function evaluations into promising areas of the search space. The choice of the acquisition function plays a crucial rule on the performance of BO and so acquisition function design is an active research area. Existing common MOBO acquisition functions include \citep{yang2019efficient, daulton2020differentiable, couckuyt2014fast, abdolshah2018expected}, to name a few.

A new class of acquisition functions for MOBO has recently arisen based on information theory. Motivated by state-of-the-art performance achieved by information-theoretic approaches for single-objective BO \citep{hernandez2014predictive, pmlr-v70-wang17e, moss2021gibbon, hvarfner2022joint}, recent works \citep{hernandez2016predictive, garrido2019predictive, belakaria2019max, suzuki2020multi, belakaria2020max, fernandez2020max} have provided several information-theoretic approaches for MOO. These acquisition functions follow the intuitive goal of seeking to reduce the amount of uncertainty (as quantified by differential entropy \citep{hennig2012entropy}) held by our surrogate model about the Pareto frontier. 

However, the full potential of multi-objective information-theoretic acquisition functions has not yet been realised, with existing work not demonstrating the state-of-the-art performance exhibited by their single-objective counterparts. This poor performance is due to the difficulty in providing a proper and efficient characterization of the mutual information between sampled observation and the Pareto frontier. Existing approaches rely on either computationally expensive approximate inference schemes like expectation propagation \citep{hernandez2016predictive, garrido2019predictive} and density filtering \citep{fernandez2020max}, or on making coarse assumptions about the structure of the Pareto front \citep{suzuki2020multi,belakaria2019max, belakaria2020max}. Indeed, both types of approximation lead to poor empirical performance \citep{daulton2021parallel}.


In this work, we provide $\{\text{PF}\}^2$ES, a new information-theoretic acquisition function for MOBO. Inspired by the recent work of \citep{poole2019variational, takeno2022sequential}, we use a variational lower bound to propose a novel cheap, accurate and explainable approximation to the joint mutual information between (batches of) candidate evaluations and (feasible) Pareto fronts. Our primary contributions are summarised as follows:

\begin{itemize}
    \item  We introduce a new information-theoretic acquisition function for multi-objective optimization that \textbf{supports both constrained and unconstrained problems}.
    \item We propose an efficient parallelisation strategy q-$\{\text{PF}\}^2$ES which provides effective \textbf{batch optimization} across batches of $q$ points.
    \item We provide \textbf{theoretical links} between $\{\text{PF}\}^2$ES and a common MOBO acquisition function: multi-objective Probability of Improvement \citep{yang2019efficient, hawe2007enhanced},  the first such analysis across information-theoretic MOBO.
    \item We provide a new taxonomy of comparing different \textit{output space} focused information theoretic acquisition function (see Section 2.3) through the \textbf{uncertainty calibration} of the Pareto frontier.
    \item We demonstrate $\{\text{PF}\}^2$ES's \textbf{competitive performance} against existing acquisition functions across a range of synthetic and real-life batch optimization problems.

\end{itemize}

\section{PRELIMINARIES}
\label{prelims}
\subsection{Multi-Objective Optimization with Unknown Constraints}  
\begin{equation}
\begin{aligned}
&  \text{Maximize}\ \bm{f}(\bm{x}) =\left(f_1(\bm{x}), f_2(\bm{x}), ..., f_M(\bm{x})\right)\\
& s.t.\ \left(g_1(\bm{x}) \geq 0, ..., g_C (\bm{x})\geq 0 \right)  : = \bm{g}(\bm{x}) \geq \bm{0}\\ & \bm{x} \in \mathcal{X} \in \mathbb{R}^d
\label{Eq: main_express}
\end{aligned}
\end{equation}  

We consider constrained multi-objective optimization (CMOO) problems, formally expressed as finding the \textbf{maximum} of a vector value function: $\bm{f}=\{f_1, ..., f_M\}: \bm{x}\rightarrow \mathbb{R}^M$ in a bounded design space $\mathcal{X} \subset \mathbb{R}^d $ that need to consider a set of \textit{unknown constraints}  $\bm{g}=\{g_1, ..., g_C\}$ (i.e., the constraint function's analytic formulation is unknown), where $M$ and $C$ represents the number of objectives and constraints respectively. A candidate $\bm{x}$ is \textit{feasible} if $\bm{g}(\bm{x}) \geq \bm{0}$. The optimal comparison of different feasible candidates is determined through the following \textit{ranking} mechanism: a feasible candidate $\bm{x}$ is preferable to $\bm{x}'$ in the sense that $\forall j \in M: f_j(\bm{x}) \geq f_j(\bm{x}')$ and $\exists j \in M: f_j(\bm{x}) > f_j(\bm{x}')$. This specific ranking strategy is termed as \textit{dominance} ($\succ$) and described as $\bm{f}(\bm{x})$ dominates $\bm{f}(\bm{x}')$: $\bm{f}(\bm{x}) \succ \bm{f}(\bm{x}')$. A feasible candidate input $\bm{x}$ is called a \textit{Pareto optimal candidate} if there do not exist any other feasible candidates in the design space that are able to dominate it; the set containing all the Pareto optimal candidates is called \textit{Pareto set} and denoted as $\bm{x}_{\mathcal{F}}$. The goal of CMOO is to efficiently identify the \textit{feasible Pareto frontier}  \footnote{This formulation can be easily adapted to MOO problem without constraints $\bm{g}$. Since we aim to tackle both MOO and CMOO problems using our acquisition function,  we overload $\mathcal{F}$ (as well as the term \textit{Pareto frontier}) to represent both feasible Pareto frontier in CMOO and Pareto frontier in the MOO problem.} $\mathcal{F}:=\{\bm{f}(\bm{x}) \vert \forall \bm{x}' \in \mathcal{X}\setminus\bm{x}_{\mathcal{F}} \ s.t.\ \bm{g}(\bm{x}')\geq \bm{0}, \exists \bm{x} \in \bm{x}_{\mathcal{F}}\ s.t.\ \bm{f}(\bm{x}) \succ \bm{f}(\bm{x}') \}$, which is constructed by the Pareto set.  


\subsection{Multi-Objective Bayesian Optimization}  
For many real-world problems, the exact form of $\bm{f}$ and $\bm{g}$ is unknown, and the evaluation of the objective functions and constraints functions $\bm{h}_{\bm{x}}:=\{f_{1_{\bm{x}}}, ..., f_{M_{\bm{x}}}, g_{1_{\bm{x}}}, ..., g_{C_{\bm{x}}}\}$ at an input location $\bm{x}$ is computationally expensive. In these settings, it is crucial to restrict the total number of observations required to find the Pareto frontier $\mathcal{F}$.


In order to achieve data efficiency, BO  \citep{frazier2018tutorial, shahriari2015taking, garnett_bayesoptbook_2022} leverages a probabilistic \textit{surrogate model} as a computationally efficient approximation of the original expensive objective function. Within this research we focus our discussion on the standard Gaussian Process (GP) \citep{rasmussen2003gaussian} framework. Given $N_{tr}$ expensive observations $D=\{\{\bm{x}\}_i, \{\bm{h}\}_i\}_{i=1}^{N_{tr}}:=\{\bm{X}_{tr}, \bm{H}\}$, a GP is able to provide a Gaussian posterior distribution of any not-yet-evaluated $\bm{h}$. Here we follow a common and most generic assumption that each objective (and constraint) are statistically independent. Consequently, the posterior distribution of $i'$th outcome $\bm{h}^i$ at unknown candidate(s) $\bm{x}$ is a (multivariate) Gaussian with mean and (co)variance defined as:  
\begin{equation}
\begin{aligned}
&\bm{m}^i(\bm{x}\vert D) = \bm{k}(\bm{x})^T\bm{K}^{-1}\bm{H}^i\\
&\mathbb{V}^i(\bm{x}\vert D) = \bm{k}(\bm{x}, \bm{x}) - \bm{k}(\bm{x})^T\bm{K}^{-1}\bm{k}(\bm{x})
\end{aligned}
\end{equation}

\noindent where $\bm{k}(\cdot, \cdot)$ represents kernel and $\bm{K}$ is the Gram matrix building upon existing input $\bm{X}_{tr}$, $\bm{H}^i$ represents the $i$th output of training data. See \cite{rasmussen2003gaussian} for a comprehensive introduction to GPs.

To guide the search into promising areas of the search space and provide highly efficient optimization, BO relies on an \textit{acquisition function} that uses the posterior belief to predict the utility of making an evaluation at any candidate input. The original expensive objective functions and constraints are then evaluated at the input with the largest predicted utility and the resulting evaluation is used to update the surrogate model. This model updating, acquisition function building, and objective function evaluation pattern iterates until a predefined stopping criterion has been met. 

\subsection{Information-Theoretic Multi-Objective Bayesian Optimization}
\label{info-moo}
One increasingly popular class of acquisition functions are those based on the now well-established information-theoretic framework. Here, we seek evaluations that provide maximal information about a given \textit{target quantity}. In the context of MOO, the target is to reduce our uncertainty about the set of optimal (feasible) trade-offs (i.e., Pareto set or Pareto frontier), however this can be formulated in two distinct ways. First, we can use the \textit{input-space formulation} \citep{hernandez2016predictive, garrido2019predictive} and calculate our uncertainty over where the optimal trade-offs lie in our search space, i.e., the Pareto set. More recently, \textit{output-space} methods \citep{belakaria2019max, belakaria2020max,fernandez2020max} have been proposed that seek to reduce the uncertainty in the (feasible) Pareto frontier $\mathcal{F}$ directly. As the discrete Pareto frontier is just an $M \cdot \vert \mathcal{F}\vert$-dimensional quantity in contrast to the $   d\cdot \vert \mathcal{F}\vert$-dimensional quantities in the Pareto set, the output-space method enjoys simpler calculations and easier numerical approximations ( at least when $d > M$). Unfortunately, although existing output-based MOO methods are much cheaper than their input-based alternatives \citep{belakaria2019max}, they all employ coarse approximations that hamper their performance and hinder their interpretation. For example, existing approaches employ approximations like assuming factorized conditional probability distributions \citep{fernandez2020max}, or using an overdone heuristic approximation to collapse $\mathcal{F}$ to its outcome-wise max \citep{belakaria2021output}.

\begin{figure*}[t]
\begin{subfigure}[t]{3.8cm}
  \centering
  \includegraphics[width=1.1\linewidth]{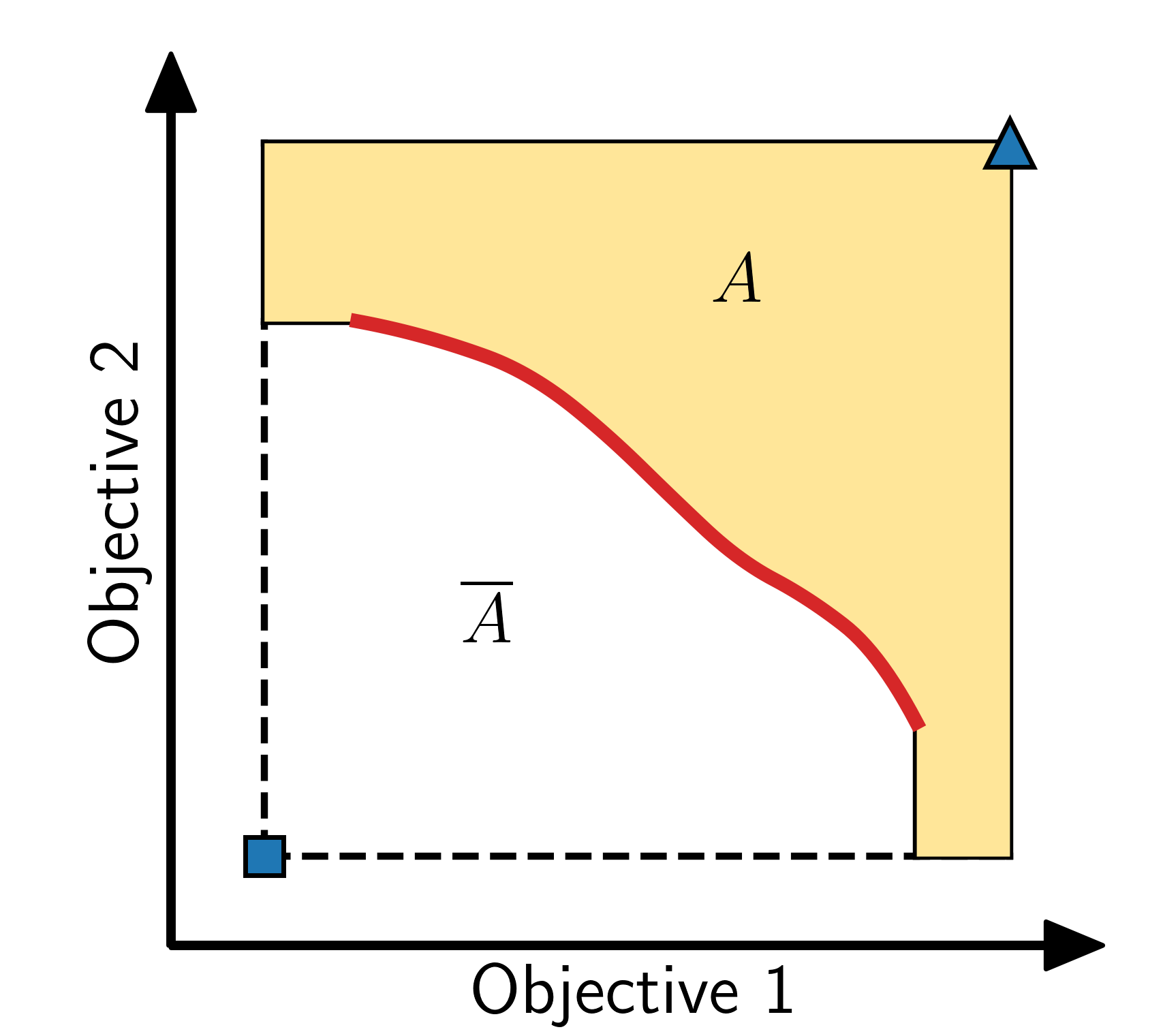}  
  \caption{}
  \label{fig:sub-first}
\end{subfigure}
\begin{subfigure}[t]{3.8cm}
\hspace*{0.1cm}
  \centering
  \includegraphics[width=1\linewidth]{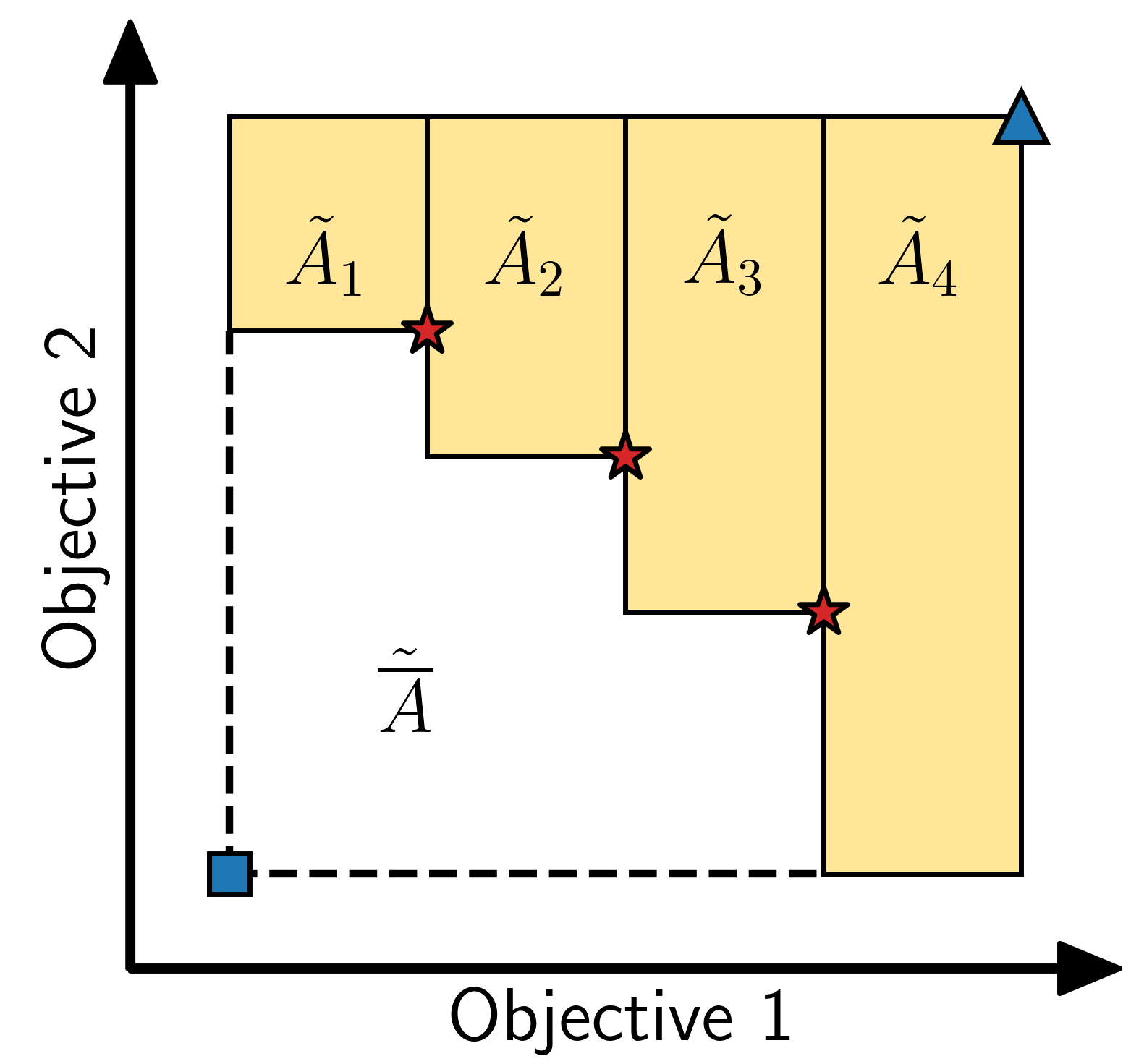}  
  \caption{}
  \label{fig:sub-second}
\end{subfigure}
\begin{subfigure}[t]{3.8cm}
  \centering
  \includegraphics[width=1.3\linewidth]{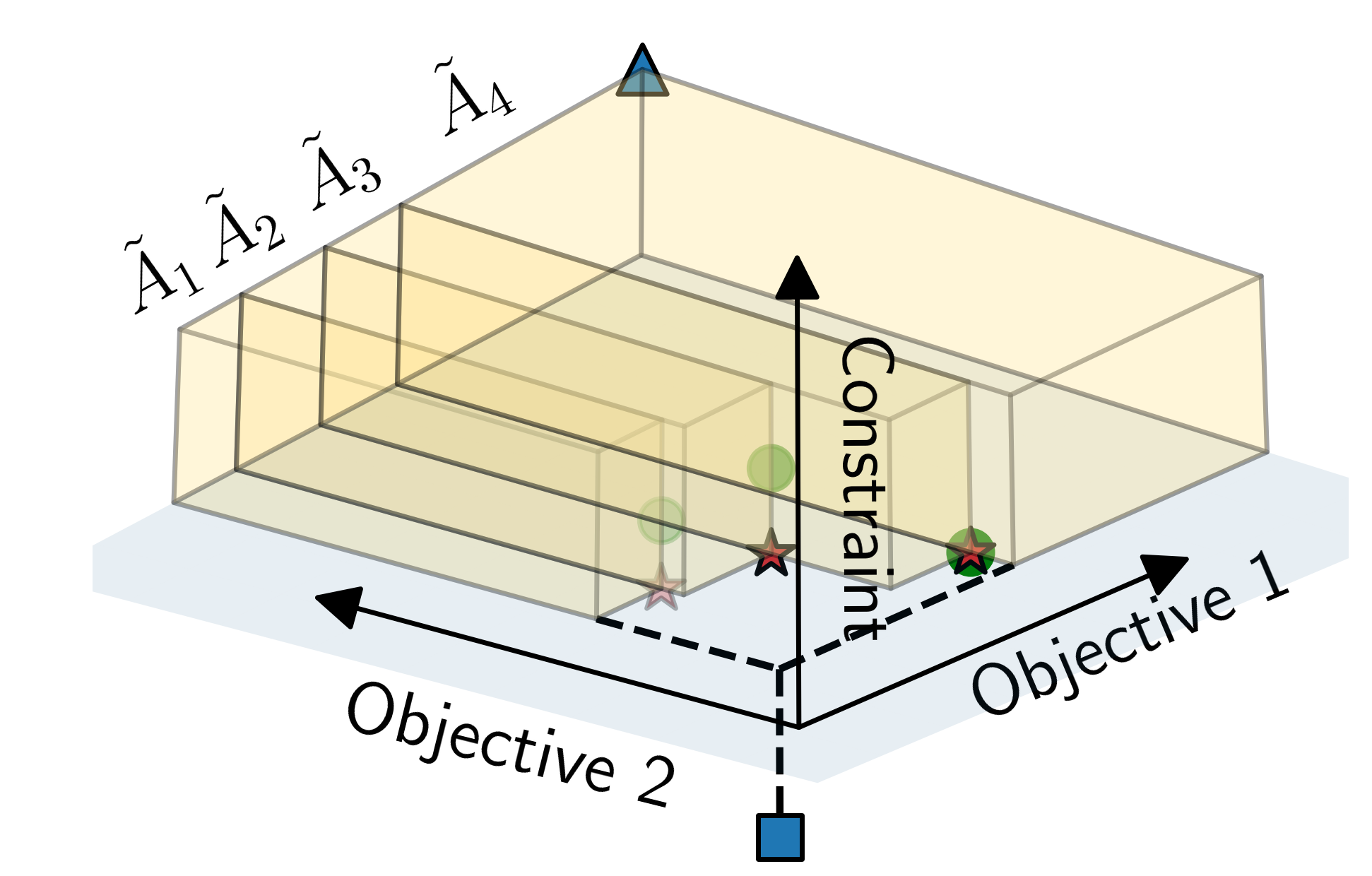}  
  \caption{}
  \label{fig:sub-third}
\end{subfigure}
  \begin{subfigure}[t]{2.5cm}
  \hspace*{1.5cm}
  \centering
  \raisebox{1cm}{\includegraphics[width=1.4\linewidth]{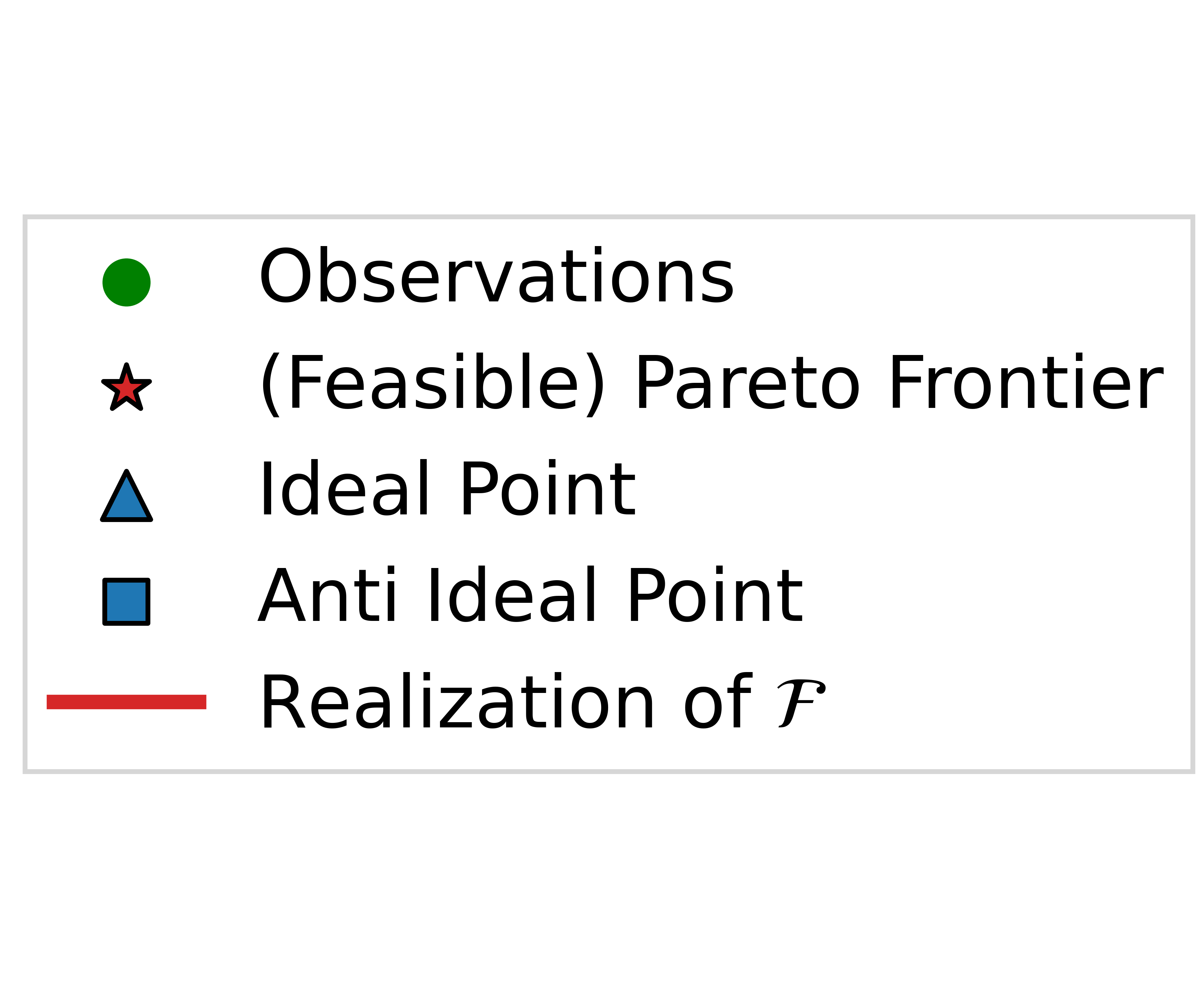}}
  \label{fig:sub-lgd}
\end{subfigure}
\caption{Partitioning the output space (bounded by the ideal point and anti-ideal point) according to a Pareto frontier: a) Given $\mathcal{F}$: Partitioning into a dominated region $\overline{A}$ and an non-dominated region $A$. b) Given a discrete Pareto frontier approximation $\tilde{\mathcal{F}}$: Partitioning the output space into an approximated dominated region $\tilde{\overline{A}}$ and non-dominated region $\tilde{A}$ by hypervolume based decomposition $\mathcal{P}$. c) A partition of the output space given a discretized feasible Pareto frontier  $\tilde{\mathcal{F}}$ for a constrained optimization problem.}
    \label{fig: obj_partition}
\end{figure*}

\section{$\{\text{PF}\}^2$ES FOR CONSTRAINED MULTI-OBJECTIVE OPTIMIZATION}  \label{Sec: FPFES-ILB}

The goal is to determine \textit{where to sample $\bm{x}$ in order to learn as much as possible about the Pareto frontier $\mathcal{F}$?} Under the information-theoretic framework, the act of learning is characterized by the Shannon mutual information $I(\cdot, \cdot)$ \citep{cover1999elements} between our target quantity  $\mathcal{F}$, and the possible distribution of the concatenated objective-constraint observations $\bm{h}_{\bm{x}}$ according to our GP surrogate models: 
\begin{equation}
    I( \mathcal{F}; \bm{h}_{\bm{x}})
    \label{Eq: mutual_info}
\end{equation}

Current output-space methods for information-theoretic MOO \citep{belakaria2020max, suzuki2020multi}  (i.e., those that seek to learn the Pareto Frontier, as discussed in Section \ref{info-moo}) recast the mutual information (Eq.~\ref{Eq: mutual_info}'s) as the reduction in differential entropy $\mathbb{H}$ of $\mathcal{F}$ provided by the candidate evaluation, i.e., using the expansion $I(\mathcal{F}, \bm{h}_{\bm{x}}) = \mathbb{H}(\mathcal{F}) - \mathbb{H}(\mathcal{F}\vert \bm{h}_{\bm{x}})$. Following this formulation, the primary difficulty is providing a reliable and efficient approximation of the differential entropy $\mathbb{H}(\mathcal{F}\vert \bm{h}_{\bm{x}})$.

\subsection{A Variational Lower Bound of the Mutual Information $I(\mathcal{F}; \bm{h}_{\bm{x}} )$}
For $\{\text{PF}\}^2$ES we avoid the difficulties of approximating the differential entropy by directly approximating the mutual information itself (Eq.~\ref{Eq: mutual_info}). In particular, we follow the ideas of \citep{poole2019variational, takeno2022sequential} and replace the mutual information with the following variational lower bound:

\begin{equation}
    \begin{aligned}
        & I(\mathcal{F}; \bm{h}_{\bm{x}} ) \\ &=  \int_{\mathcal{F}} \int_{\bm{h}_{\bm{x}}} p(\bm{h}_{\bm{x}}, \mathcal{F})\text{log} \frac{ p(\bm{h}_{\bm{x}}, \mathcal{F})}{p(\bm{h}_{\bm{x}})p(\mathcal{F})}d\mathcal{F}d\bm{h}_{\bm{x}} \\  & = \mathbb{E}_{\mathcal{F}}\left[\int_{\bm{h}_{\bm{x}}}p(\bm{h}_{\bm{x}} \vert \mathcal{F})\text{log} \frac{ p(\bm{h}_{\bm{x}}\vert \mathcal{F})q(\bm{h}_{\bm{x}} \vert \mathcal{F})}{p(\bm{h}_{\bm{x}})q(\bm{h}_{\bm{x}} \vert \mathcal{F})}d\bm{h}_{\bm{x}}\right]  \\ & = \mathbb{E}_{\mathcal{F}}\Big[\int_{\bm{h}_{\bm{x}}}p(\bm{h}_{\bm{x}} \vert \mathcal{F})\text{log} \frac{ q(\bm{h}_{\bm{x}} \vert \mathcal{F})}{p(\bm{h}_{\bm{x}})}d\bm{h}_{\bm{x}} + \\ &\quad \quad \quad D_{KL}(p(\bm{h}_{\bm{x}} \vert \mathcal{F}) || q(\bm{h}_{\bm{x}} \vert \mathcal{F}))\Big]  \\ & \geq \mathbb{E}_{\mathcal{F}}\left[\int_{\bm{h}_{\bm{x}}}p(\bm{h}_{\bm{x}} \vert \mathcal{F})\text{log} \frac{ q(\bm{h}_{\bm{x}} \vert \mathcal{F})}{p(\bm{h}_{\bm{x}})}d\bm{h}_{\bm{x}} \right]  
    \end{aligned}
\label{Eq: lower_bound_of_mutual_info}
\end{equation}

\noindent where $D_{KL}$ represents the KL-divergence and the density  $q(\bm{h}_{\bm{x}} \vert \mathcal{F})$ is a variational approximation of the ground truth conditional distribution $p(\bm{h}_{\bm{x}} \vert \mathcal{F})$. The inequality in the final line of  Eq.~\ref{Eq: lower_bound_of_mutual_info} is due to the non-negativity of the KL-divergence. As the gap between the true mutual information and our lower bound can be explicitly seen as the KL-divergence between the variational approximation and  ground truth conditional distribution, the suitability of this lower bound depends on our ability to build a reasonable approximation $q(\bm{h}_{\bm{x}} \vert \mathcal{F})\approx p(\bm{h}_{\bm{x}} \vert \mathcal{F})$.

We now explain how to build an effective variational approximation $q(\bm{h}_{\bm{x}} \vert \mathcal{F})$. Given a realization of the Pareto frontier $\mathcal{F}$ (illustrated in Fig.~\ref{fig:sub-first} for an unconstrained problem), the \textit{output space} $\mathbb{R}^{M+C}$ can be partitioned into two complementary regions $A(\mathcal{F})$ and $\overline{A}(\mathcal{F})$, i.e., $ A(\mathcal{F}) \cup \overline{A}(\mathcal{F}) = \mathbb{R}^{M+C}$ \footnote{We note the output space is actually defined in a bounded space formulated by two extreme points: \textit{ideal point} and \textit{anti-ideal point} as shown in Fig.~\ref{fig:sub-first} to allow practical objective space partition, however, in \{PF\}$^2$ES we set these two extreme points a vector of large constant (i.e., 1e20) and omit the notation of them afterwards.}, where  $A(\mathcal{F}) \subset \mathbb{R}^{M+C}$ is the \textit{feasible non-dominated region} $\{ \{\bm{f}_{\bm{x}}, \bm{g}_{\bm{x}}\}: \nexists \bm{x}^* \in  \bm{x}_{\mathcal{F}}\ s.t.\ \bm{f}_{\bm{x}^*}\succ \bm{f}_{\bm{x}},\ \bm{g}_{\bm{x}}\geq \bm{0} \}$ in the output space.  We then set $q(\bm{h}_{\bm{x}} \vert \mathcal{F})$ as (an extension of the concept of \textit{Pareto frontier truncated normal distribution} \citep{suzuki2020multi}):
\begin{equation}
q(\bm{h}_{\bm{x}} \vert \mathcal{F})  = \left\{
\begin{aligned}
&\frac{p(\bm{h}_{\bm{x}} )}{Z_{\overline{A}}} \quad \bm{h}_{\bm{x}} \in \overline{A}\\
&0 \quad\quad\quad\  \text{else}\\
\end{aligned} 
\right.
\label{Eq: conditional_prob}
\end{equation}

\noindent where $Z_{\overline{A}}:= \int_{\overline{A}} p(\bm{h}_{\bm{x}})d\bm{h}_{\bm{x}} = 1 - Z_A$ is the probability of $\bm{h}_{\bm{x}}$ be in $\overline{A}$, which is constructed based on the frontier realization $\overline{A}(\mathcal{F})$ . The choice of variational approximation $q$ is motivated by that, given a fixed $\mathcal{F}$, for any $\bm{x}, \bm{h_x}$; we assign $q(\bm{h_x}|\mathcal{F})=0$ for $\bm{h_x}$ to be feasible and non-dominated (i.e.,  there exists $\bm{x}' \in \bm{x}_\mathcal{F}$ such that $\bm{f_x}  \succ \bm{f(x')}$ and $\bm{g_x > 0} $). This of course assumes that we are able to obtain the expensive observations without noise which is a common scenario (e.g., \citep{vakili2020regret}).

 By substituting our approximate conditional distribution (Eq.~\ref{Eq: conditional_prob}) into Eq.~\ref{Eq: lower_bound_of_mutual_info}, the variational information lower bound has the following simple expression  \footnote{By convention \citep{cover1999elements} $0log0=0$ and so $\int_{\bm{h}_{\bm{x}} \in A}p(\bm{h}_{\bm{x}} \vert \mathcal{F}) log p(\bm{h}_{\bm{x}} \vert \mathcal{F})d\bm{h}_{\bm{x}} = 0$. See Appendix~A of \citep{takeno2022sequential} for a discussion.}

\begin{equation}
\begin{aligned}
   I(\mathcal{F}; \bm{h}_{\bm{x}} )   & \geq   \mathbb{E}_{\mathcal{F}}\left[ \int_{\bm{h}_{\bm{x}} \in \overline{A}(\mathcal{F})}p(\bm{h}_{\bm{x}} \vert \mathcal{F}) d\bm{h}_{\bm{x}}\cdot \text{log} \frac{1 }{Z_{\overline{A}(\mathcal{F})}}\right]\\ &=  - \mathbb{E}_{  \mathcal{F}}\left[\text{log}\left(1 -Z_{A(\mathcal{F})} \right)\right] = \alpha_{\text{\{PF\}}^2\text{ES}}.
\label{Eq: PF2ES_Exact}
\end{aligned}
\end{equation}

\noindent which leads to the proposed \{PF\}$^2$ES acquisition function.  

\textbf{Explainability of \{PF\}$^2$ES}:  Given the \{PF\}$^2$ES's expression in Eq.~\ref{Eq: PF2ES_Exact}, we are able to link it with the common Multi-Objective Probability of Improvement (MOPI) \citep{yang2019efficient, hawe2007enhanced}. MOPI is the multi-objective version of Probability of Improvement \citep{kushner1964new} and measures utility by the probability that a new candidate will locate in the non-dominated region $A(\mathcal{F})$, which can be constructed based on a Pareto frontier $\mathcal{F}$. MOPI can be adapted to the CMOO setting, where it can be multiplied with a Probability of Feasibility (PoF) term ($\alpha_{MOPI} \times \alpha_{PoF}$) following the approach of \citep{hawe2008probability, gardner2014bayesian}. This, under the assumption that $\bm{f}$ and $\bm{g}$ is statistically independent, is equivalent to the probability that a new candidate is feasible and located in the non-dominated region. 

The link between \{PF\}$^2$ES and MOPI can be established through the following remark (see Appendix.~\ref{App: relationship_with_pi} for proof): 

\begin{remark}
 The following acquisition functions lead to the same maximal candidate $\bm{x} = arg\ \underset{\bm{x} \in \mathcal{X}}{max}\ \alpha(\bm{x})$:  
 \begin{enumerate}
  \item MOO
  \begin{enumerate}
    \item  \label{item: MOPI_MOO} $\alpha_{MOPI}$ using $\mathcal{F}$ as Pareto frontier
    \item $\alpha_{\{\text{PF}\}^2\text{ES}}$ with $ \vert \tilde{\mathbf{F}}\vert=1 $ and use the same $\mathcal{F}$ as 1.1.
  \end{enumerate}
  \item Constrained MOO when 
    \begin{enumerate}
    \item  \label{item: MOPI_CMOO} $\alpha_{MOPI}$ $*$ $\alpha_{PoF}$ using $\mathcal{F}$ as reference Pareto frontier
    \item  $\alpha_{\{\text{PF}\}^2\text{ES}}$ with $ \vert \tilde{\mathbf{F}}\vert=1 $ and use the same $\mathcal{F}$ as 2.1. 
  \end{enumerate}
\end{enumerate}
\label{rm: connection}
\end{remark}
where $\tilde{\mathbf{F}}$ is defined in Eq.~\ref{Eq: PF2ES_naive_approx}. We note this linkage provides insights into the empirical performance of  $\{\text{PF}\}^2$ES as well.



\section{PRACTICAL CALCULATION OF $\{\text{PF}\}^2$ES AND ITS PARALLELIZATION}

Practical calculation of the variational lower bound (Eq.~\ref{Eq: PF2ES_Exact}) requires some additional steps. The primary difficulty is that the region $A$ (for any given Pareto frontier $\mathcal{F}$) does not have an analytical form. A popular approach \citep{belakaria2019max, suzuki2020multi, hernandez2016predictive, garrido2019predictive} is to use a finite representation of the Pareto frontier $\tilde{\mathcal{F}}$:= $\{\bm{f} \vert  \bm{f} \in \mathcal{F}\}\ s.t.\  \vert \tilde{\mathcal{F}} \vert  < \infty$. Based on $\tilde{\mathcal{F}}$, a hypervolume-based output space partitioning strategy \citep{lacour2017box, Couckuyt2012}, denoted as $\mathcal{P}$, can be used to partition the output space into $N_p$ disjoint hypercubes, thus obtaining an approximation of $A$ denoted by \textit{hypervolume based approximated non-dominated region}:
\begin{equation}
     A \approx \tilde{A}(\tilde{\mathcal{F}}) = \mathcal{P}(\tilde{\mathcal{F}})= \{\tilde{A}_1, \tilde{A}_2, ..., \tilde{A}_{N_p}\}
\end{equation}

which is the complement of $\tilde{\overline{A}}(\tilde{\mathcal{F}}):= \{\bm{f}_{\bm{x}}\in \mathbb{R}^M \vert \exists \bm{f} \in  \tilde{\mathcal{F}}, \bm{f}_{\bm{x}} \preceq \bm{f}  \}$. Such partitioning is illustrated in Fig.~\ref{fig:sub-second} for unconstrained MOO problems. For CMOO problem, $\tilde{A}(\tilde{\mathcal{F}})$ can be obtained by first partitioning the non-dominated region in the \textit{objective space}: $\mathbb{R}^M$ and then concatenated with the constraint space where $\bm{g}\geq \bm{0}$, see Fig.~\ref{fig:sub-third} for an illustration.

Given $\tilde{A}$, we are able to reach the following analytically tractable approximation of Eq.~\ref{Eq: PF2ES_Exact}:
\begin{equation}
\begin{aligned}
       - \mathbb{E}_{  \mathcal{F}}\left[\text{log}\left(1 -Z_{A(\mathcal{F})} \right)\right]  \approx - \frac{1}{ \vert \tilde{\mathbf{F}}\vert }\sum_{\tilde{\mathcal{F}} \in \tilde{\mathbf{F}}} \left[\text{log}\left(1 -\sum_{i=1}^{N_p}Z_{{\tilde{A}}_i} \right)\right] \quad 
\end{aligned}
\label{Eq: PF2ES_naive_approx}
\end{equation}

\noindent where $\tilde{\mathbf{F}}$ is a set of sampled Pareto frontiers providing a Monte Carlo (MC) approximation of the outer expectation in Eq.~\ref{Eq: PF2ES_Exact}.  We build this MC sample of $\tilde{\mathcal{F}}$ by using a standard multi-objective optimization algorithm (e.g., NSGAII \citep{deb2002fast}) on samples from the GP's spectral posterior (also known as sampling GP posterior via its Fourier features \citep{rahimi2007random}), which is a common strategy utilized in acquisition functions for MOO \citep{belakaria2019max, suzuki2020multi, daulton2022multi}. $Z_{\tilde{A}_i}$ is the probability that $\bm{h}_{\bm{x}}$ is in the $i'$th hypercube $\tilde{A}_i$ and can be calculated as \citep{suzuki2020multi}:
\begin{equation}
    Z_{\tilde{A}_i} = \prod_{k=1}^{M+C}\left(\Phi(\frac{\tilde{A}_{i_u}^k - \bm{m}_{\bm{x}}^k }{\bm{\sigma}_{\bm{x}}^k}) - \Phi(\frac{\tilde{A}_{i_l}^k - \bm{m}_{\bm{x}}^k }{\bm{\sigma}_{\bm{x}}^k}) \right).
\end{equation}

\noindent where $\Phi(\cdot)$ represents the cumulative density function of the standard normal distribution, $\tilde{A}_{i_u}^k, \tilde{A}_{i_l}^k$ represents the $k'$th dimensional upper and lower bound of the $i'$th hypercube, respectively. $\bm{m}_{\bm{x}}^k, \bm{\sigma}_{\bm{x}}^k$ represent the $k'$th dimensional GP posterior mean and standard deviation, respectively.

Unfortunately, due to the subtlety of our employed partitioning strategy, we cannot simply use  the r.h.s of Eq.~\ref{Eq: PF2ES_naive_approx} as $\alpha_{\{\text{PF}\}^2\text{ES}}$.  In particular, the hyper-volume based partition $\mathcal{P}$, operating on the finite representation $\tilde{\mathcal{F}}$, means that  $\tilde{A}$ is an \textbf{overestimation} of the non-dominated region $A$. Therefore, we are no longer guaranteed to satisfy the lower bound property in Eq.~\ref{Eq: PF2ES_Exact} since $Z_{A(\mathcal{F})} < Z_{A(\tilde{\mathcal{F}})} $. More importantly, calculating \{PF\}$^2$ES based on the above $\tilde{A}$ results in the same empirical \textit{clustering issue} observed in MOPI, i.e. that the probability of improvement rewards tiny improvements of the Pareto frontier approximation $\tilde{\mathcal{F}}$ 
if they are achieved with high confidence \citep{emmerich2020infill} (see Appendix.~\ref{App: clustering_issue} for a detailed elaboration). 

In order to mitigate this clustering issue whilst ensuring that our acquisition function remains true to its motivation as a lower bound to the mutual information, we propose a small change to how we construct our approximation of $A$. Inspired by the common empirical strategy employed in MOPI and PI acquisition functions (\citep{wang2016optimization, emmerich2020infill, kushner1964new}), we introduce a small positive penalization vector $ \bm{\varepsilon} = [\varepsilon_1, ..., \varepsilon_M]$ to shift the discrete Pareto frontier approximation:$\tilde{\mathcal{F}_{\bm{\varepsilon}}}:=\tilde{\mathcal{F}} + \bm{\epsilon}$ to avoid any overestimation of $A$ resulting from the hyper-volume decomposition. In Appendix.~\ref{app:alternative_epsilon}, we show that it is possible to construct an $\epsilon$ that ensures that our acquisition function remains a lower bound. However, building such an epsilon is costly, so we also propose the practical solution of setting $\epsilon$ through a simple heuristic that achieves very similar empirical performance (see Appendix.~\ref{app:alternative_epsilon}). In particular, we set epsilon as $\epsilon_k =  c \left(max(\tilde{\mathcal{F}^k}) - min(\tilde{\mathcal{F}^k})\right)$ $\forall k \in M, \tilde{\mathcal{F}} \in \tilde{\mathbf{F}}$. i.e. a proportion of the maximal total variation observed over the Pareto front. We show empirically that \{PF\}$^2$ES's performance is robust to the choice of $c$ through sensitivity analysis (Appendix.~\ref{app:sensitivity_analysis}). We set  $c=0.04$ for all our experiments.  

Combining all these steps, we can finally state our proposed acquisition function as

\begin{equation}
\begin{aligned}
       \tilde{\alpha}_{\{\text{PF}\}^2\text{ES}} = - \frac{1}{ \vert \tilde{\mathbf{F}}\vert }\sum_{\tilde{\mathcal{F}} \in \tilde{\mathbf{F}}} \left[\text{log}\left(1 -\sum_{i=1}^{N}Z_{{\tilde{A}}_i(\mathcal{F}_{\varepsilon})} \right)\right]. \quad 
\end{aligned}
\label{Eq: PF2ES_fix_approx}
\end{equation} See Algorithm \ref{alg: original} in Appendix.~\ref{app:alternative_epsilon} for a high-level algorithmic summary of our calculation strategy.



\subsection{Parallelisation of $\{\text{PF}\}^2$ES}

For many practical MOO problems, it is common to have parallel or distributed evaluation resources available, i.e. we recommend $q>1$ evaluations during each BO step. We now propose an extension of  $\{\text{PF}\}^2$ES that allows it to be used for such batch optimizations. This time, the principled question is \textit{at which $q$ points should we sample simultaneously to learn as much as possible about $\mathcal{F}$}? 

Consider the random variable $\bm{h}_{\bm{X}} = \{\bm{h}_{\bm{x}_1}, ..., \bm{h}_{\bm{x}_q}\}$, i.e., the possible observations that could arise from a parallel evaluation of a batch of $q$ points. In this scenario, we want to allocate a batch of $q$ points $\textbf{X}$ that provide the most mutual info about the (feasible) Pareto frontier. Following the same derivation as Eq.~\ref{Eq: lower_bound_of_mutual_info} - \ref{Eq: PF2ES_Exact}, we are able to have our proposed extension of $\{\text{PF}\}^2$ES for batch design:

\begin{equation}
\begin{aligned}
  & \tilde{\alpha}_{\text{q-}\{\text{PF}\}^2\text{ES}} \approx   - \frac{1}{ \vert \tilde{\mathbf{F}}\vert }\sum_{\tilde{\mathcal{F}} \in \tilde{\mathbf{F}}} \left[\text{log}\left(1 - Z_{\tilde{A}_{q}(\tilde{\mathcal{F}}_{\bm{\varepsilon}})}\right)\right],
\end{aligned}
\label{Eq: qPF^2ES}
\end{equation}

\noindent where $Z_{\tilde{A}_{q}}:= \int_{\tilde{A}_{q}} p(\bm{h}_{\bm{X}})d\bm{h}_{\bm{X}} $  see Eq. \ref{Eq: batch_conditional_prob}). i.e., the probability that there exists at least one batch element $\bm{h}_{\bm{x}_i} \in \bm{h}_{\bm{X}}$ such that $\bm{h}_{\bm{x}_i} \in \tilde{A}(\tilde{\mathcal{F}}_{\bm{\varepsilon}})=\{\tilde{A}_1(\tilde{\mathcal{F}}_{\bm{\varepsilon}}), ..., \tilde{A}_N(\tilde{\mathcal{F}}_{\bm{\varepsilon}})\}$. Calculating the probability of $Z_{\tilde{A}}$ involves a complex interaction between joint batch points within the complex region $\tilde{A}$; we hence propose to approximate this probability through an MC approximation  (a detailed derivation is provided in Appendix.~\ref{app:derive_parallelization}): 

\begin{equation}
\begin{aligned}
  &Z_{\tilde{A}_{q}}  \\ & \approx \frac{1}{N_{MC}}\sum_{j=1}^{N_{MC}}\left(\bigcup_{i=1}^{N_p}\left(\bigcup_{l=1}^{q} \left(\prod_{k=1}^{M+C} \left(  \mathbbm{1} (A_{i_l}^k \leq \bm{h}_{\bm{x}_{l_j}}^k \leq A_{i_u}^k)  \right) \right)\right)\right) \\& \approx \frac{1}{N_{MC}}\sum_{j=1}^{N_{MC}}\left(max_i\left(max_l \left(\prod_{k=1}^{M+C} \left( \sigma(\frac{\bm{h}_{\bm{x}_{l_j}}^k - A_{i_l}^k}{\tau}) \cdot \right.\right.\right.\right.\\ & \left.\left.\left.\left.\quad  \sigma(\frac{A_{i_u}^k - \bm{h}_{\bm{x}_{l_j}}^k}{\tau})  \right) \right)\right)\right),
\end{aligned}
\label{eq: batch_mopi}
\end{equation}




\noindent where $\bm{h}_{{\bm{x}_l}_j}^k$ represents the $k'$th output dimensionality of $j'$th MC sample of $l'$th batch point $\bm{h}_{{\bm{x}_l}}$, and $\sigma(\cdot)$ is the sigmoid function, $N_{MC}$ represents the MC sample size. To ensure the differentiability of our acquisition function, we follow a common strategy in BO \citep{wilson2018maximizing, daulton2020differentiable} to relax the categorical event imposed by $ \mathbbm{1} (\cdot)$ in the first line of Eq.~\ref{eq: batch_mopi} by replacing it with a sigmoid function and a small non-negative temperature parameter $\tau$.  We note Eq.~\ref{eq: batch_mopi} can be explained as calculating the event that \textbf{whether any batch element is within $\tilde{A}$}, averaged through $N_{MC}$  sample numbers.
For joint sampling of the batch outcome $\bm{h}_{\bm{X}}$, we use the reparameterization trick in combination with a sample average approximation (SAA) \citep{balandat2020botorch} to perform a continuous acquisition function optimization process, where the base sample of SAA has been generated through a randomized quasi-Monte Carlo ( details of the approximation and a demonstration of q-\{PF\}$^2$ES is provided in Appendix.~\ref{app:demonstrate_parallelization}).    

We stress the existence of remark.~\ref{rm: connection} confirms that our MC-based parallelisation strategy (Eq.~\ref{eq: batch_mopi}) can be directly applied to the MOPI acquisition function as well.

\textbf{Complexity Analysis}
Lastly, we provide complexity analysis in Appendix \ref{sec:computation_complexity}, A run-time comparison with other acquisition functions utilized in the next section is also conducted and detailed in Appendix \ref{app:runtime}.

\section{EXPERIMENTAL VALIDATION OF  $\{\text{PF}\}^2$ES}
\label{sec:: exp}
We now present the empirical performance of $\{\text{PF}\}^2$ES across constrained and unconstrained synthetic benchmarks and real-life application problems. The primary focus of these experiments is to compare our methods with other output-based entropy search methods, we have also included the golden standard EHVI acquisition functions as a performance reference \footnote{Our code is available at \url{https://github.com/TsingQAQ/trieste/tree/PF2ES_preview_notebook}.}.  

\textbf{Sequential MOO problem} We compare against PFES \citep{suzuki2020multi}, MESMO \citep{belakaria2020max}, EHVI \citep{yang2019efficient}, PESMO \citep{hernandez2016predictive}, as well as providing random search as a common baseline.  

\textbf{Parallel MOO problem} We compare against qEHVI \citep{daulton2020differentiable}, \{PF\}$^2$ES with a fantasizing method (i.e., the Kriging Believer (KB) method \citep{ginsbourger2010kriging}) and random search. 

\textbf{Sequential CMOO problem} We compare against EHVI-PoF (the multiplication of EHVI with PoF as a common strategy for CMOO (e.g., \cite{martinez2016kriging})), MESMOC \citep{belakaria2021output}, MESMOC+ \citep{fernandez2020max} and random search. 

\textbf{Parallel CMOO problem}  We compare against qEHVI \citep{daulton2020differentiable}\footnote{We note that the EHVI acquisition function typically requires a reference point to calculate, since the setting of reference point can introduce bias hence performance difference, in our numerical experiments we assume such \textbf{a reference point is unknown beforehand and need to be calculated dynamically in each iteration}, we refer Appendix.~\ref{app:synthetic_probs_and_ref_pts} on how such a reference point is specified.}, PPESMOC \citep{garrido2020parallel}, random Search and \{PF\}$^2$ES using a fantasize method \citep{ginsbourger2010kriging}.   



\begin{figure*}[t]
\begin{subfigure}[t]{15cm}
  \centering
  \includegraphics[width=1.1\linewidth]{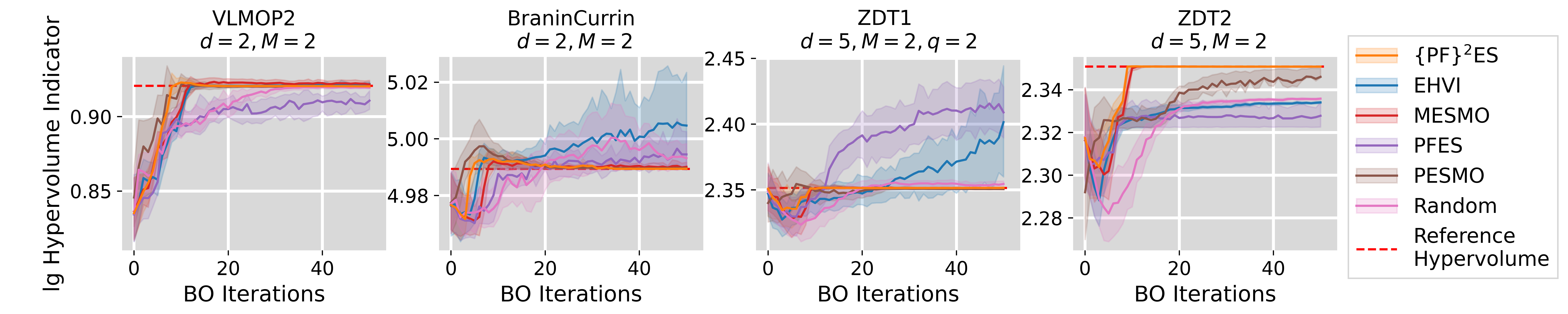}  
  \label{fig:uc_moo}
\end{subfigure}
\begin{subfigure}[t]{15cm}
\vspace*{-0.5cm}
  \centering
  \includegraphics[width=1.1\linewidth]{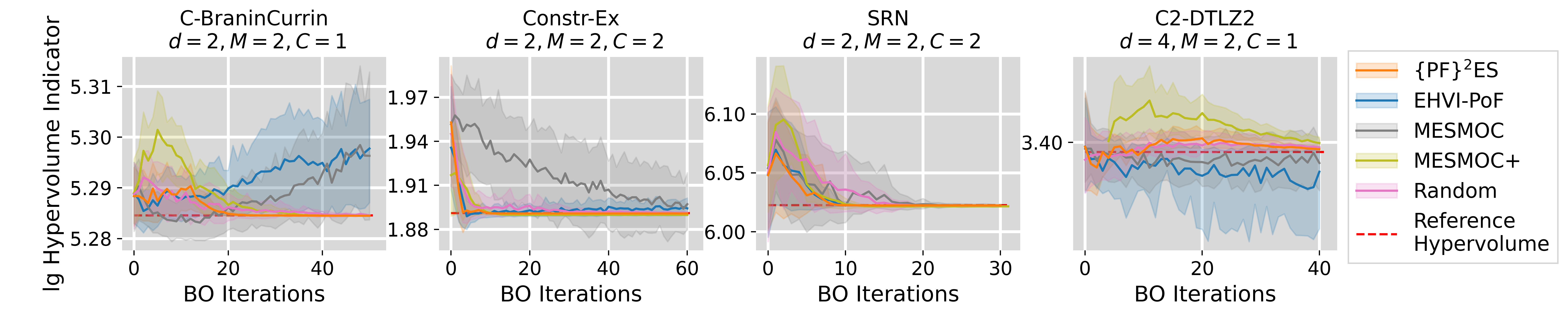}  
  \label{fig:uc_cmoo}
\end{subfigure}
\caption{Uncertainty Calibration of inferred Pareto Frontier during BO iterations.}
\label{fig:UC}
\end{figure*}

For PESMO, MESMOC+ and PPESMOC, we used the public implementation provided by the papers. The settings of all acquisition functions are detailed in Appendix.~\ref{app:acq_setups}. For the surrogate models, we build GPs with Mat\'ern 5/2 kernels using maximum a posteriori estimates for the kernel parameters. For optimizing the acquisition functions, we use a multi-start L-BFGS-B optimizer starting from the $min(10 \times qd, 100)$ \footnote{The upper limit is set to restrict the computation cost of multi-start L-BFGS-B grow continuously with $q$ and $d$, especially for higher input dimensionality and large batch size, see Fig. \ref{fig:Acq_opt_comparison} in Appendix \ref{app:runtime}.} best locations from 5000 random starting locations. For the hyperparameter of the acquisition function, we empirically choose $\tau = 1e-3$ . For sampling the (feasible) Pareto frontier $\tilde{\mathcal{F}}$, we use the open-source NSGAII optimizer in PyMOO \citep{pymoo}, where constraints are handled by the \textit{parameter-less approach} \citep{deb1999niched}. For all our information-theoretic acquisition functions, we use five sampled Pareto frontiers. All of the following experiments start with $2d+1$ random sampled initial points.  For the recommendation of optimal candidates, we use the out-of-sample strategy (performing recommendation based on the GP model) as a common approach in the information-theoretic acquisition function (see Appendix.~\ref{app:out_of_sample} for details). 

First, we consider a suite of popular synthetic benchmarks. Details of the benchmark functions, the reference point settings, and benchmark results on in-sample recommendations are provided in Appendix.~\ref{app:synthetic_probs_and_ref_pts}, \ref{app:in_sample_res} respectively.

\begin{figure*}[t]
\begin{subfigure}[t]{15cm}
  \centering
  \includegraphics[width=1.1\linewidth]{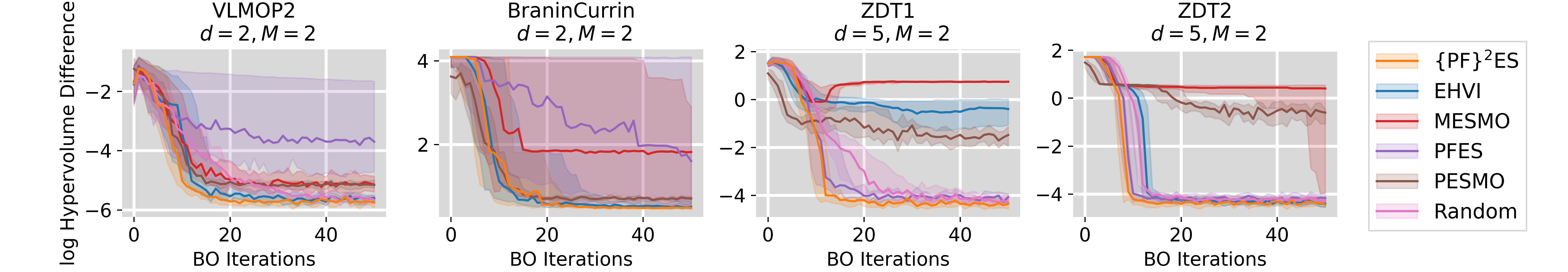}  
\end{subfigure}
\begin{subfigure}[t]{15cm}
\vspace*{-0.2cm}
  \centering
  \includegraphics[width=1.1\linewidth]{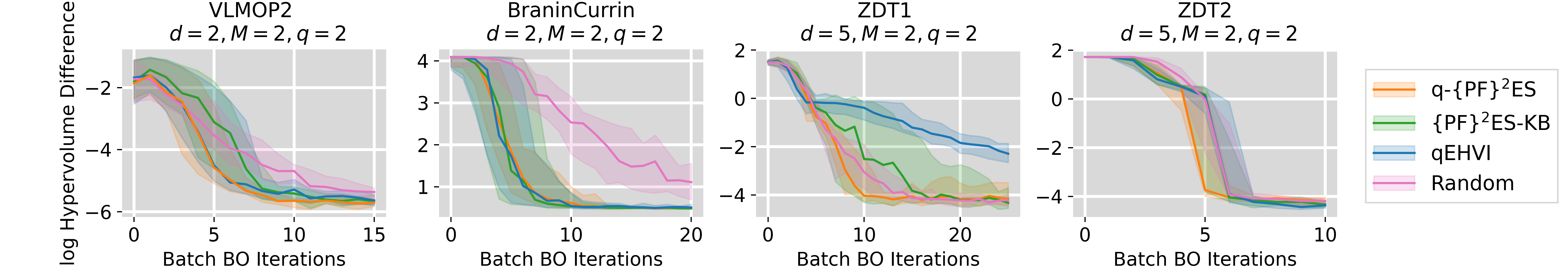}  
\end{subfigure}
\vspace*{-0.2cm}
\begin{subfigure}[t]{15cm}
  \centering
  \includegraphics[width=1.1\linewidth]{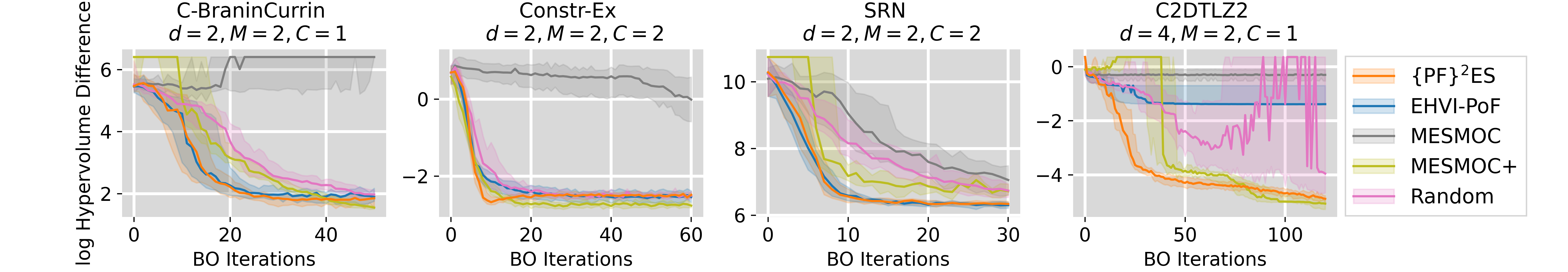}  
\end{subfigure}
\begin{subfigure}[t]{15cm}
\vspace*{-0.2cm}
  \centering
  \includegraphics[width=1.1\linewidth]{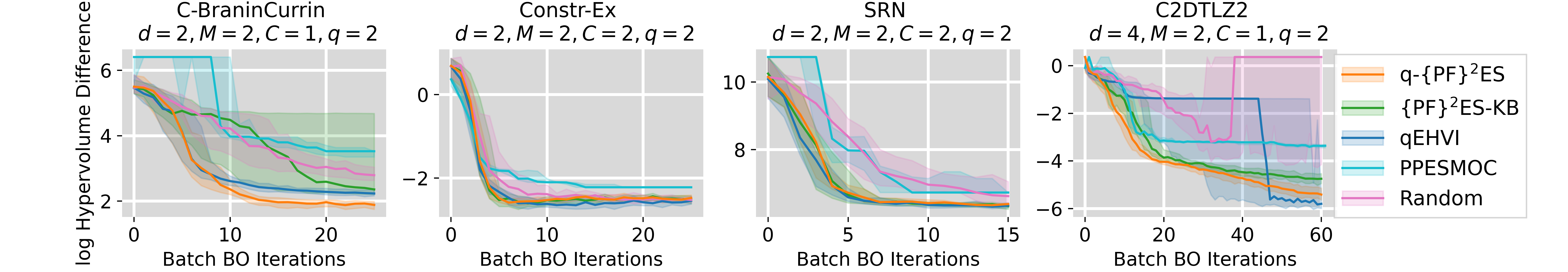}  
\end{subfigure}
\caption{Comparison of different acquisition functions for multi-objective Bayesian optimization. Results are shown through median performances and interquartile ranges.}
\label{fig:os_benchmark}
\end{figure*}

\subsection{Uncertainty Calibration of Inferred Pareto Frontier}

To help discern between \{PF\}$^2$ES and other information-theoretic acquisition functions that choose points with the overarching goal of reducing uncertainty in the Pareto frontier, it is natural to include a performance metric that directly measures this property. 

Therefore, inspired by an approach in active learning \citep{qing2021adaptive}, we propose an uncertainty calibration procedure on a \textit{hypervolume based Pareto frontier indicator distribution} that reflects the uncertainty of GPs w.r.t the Pareto frontier. More specifically, the hypervolume indicator function $I(\cdot)$ \citep{guerreiro2020hypervolume} is applied to each sampled Pareto frontier $\tilde{\mathcal{F}}$ from the GPs. This way, a one-dimensional distribution of hypervolume indicators is obtained. Note that if the sampled Pareto frontiers have converged to the (finite approximation of the)  Pareto frontier $\mathcal{F}$, the distribution will collapse to the hypervolume indicator of the (finite approximation of the)  Pareto frontier $\mathcal{F}$. Hence, the expectation and uncertainty of the hypervolume based Pareto frontier indicator distribution can serve as a representation of the Pareto frontier estimation accuracy as well as the uncertainty based on GPs, respectively.



The proposed uncertainty calibration metric is illustrated in Fig.~\ref{fig:UC}.  For each BO iteration, the Pareto frontier indicator distribution $I(\tilde{\mathcal{F}}(GP))$ is approximated by 10 MC samples of $\tilde{\mathcal{F}}$ and we present the aggregate results across 30 repetitions of experiments by their median and 10-90 percentile. It can be seen that \{PF\}$^2$ES achieved the fastest convergence to the reference hypervolume, implying low uncertainty about the Pareto frontier indicator distribution.  

\subsection{Multi-Objective Bayesian Optimization}
\textbf{Synthetic Problems}

We now present  our main experimental results. The log-hypervolume difference achieved across 30 repetitions for each algorithm is reported in Fig.~\ref{fig:os_benchmark}. We see that \{PF\}$^2$ES leads to competitive performance in both MOO and CMOO problem in the case of sequential and batch sampling. Note that the performance of EHVI, often regarded as the gold standard, relies on setting a reasonable reference point. For problems where there is no prior information known about the reference point location, dynamic reference updating strategies must be used.  For example, EHVI performs poorly on C2DTLZ2 which has disjoint Pareto frontiers \citep{deb2019constrained}. Additional experimental results on larger batch sizes and larger output dimensionalities are provided in Appendix.~\ref{sec: 9.2} and \ref{sec: 9.3} respectively.



\textbf{Four BarTruss Design}

We consider a real-life mechanical design problem (MOO) \citep{tanabe2020easy, dauert1993multicriteria}, where we seek a four elastic truss structure system with small structure volume whilst having small deformation under external forces.  The cross-section areas of the four bars in the truss system are chosen as the design variables for the unconstrained multi-objective optimization problem. Note the scale of the two objectives is significantly different for this problem. 

\begin{figure}[t]
  \includegraphics[width=1\linewidth]{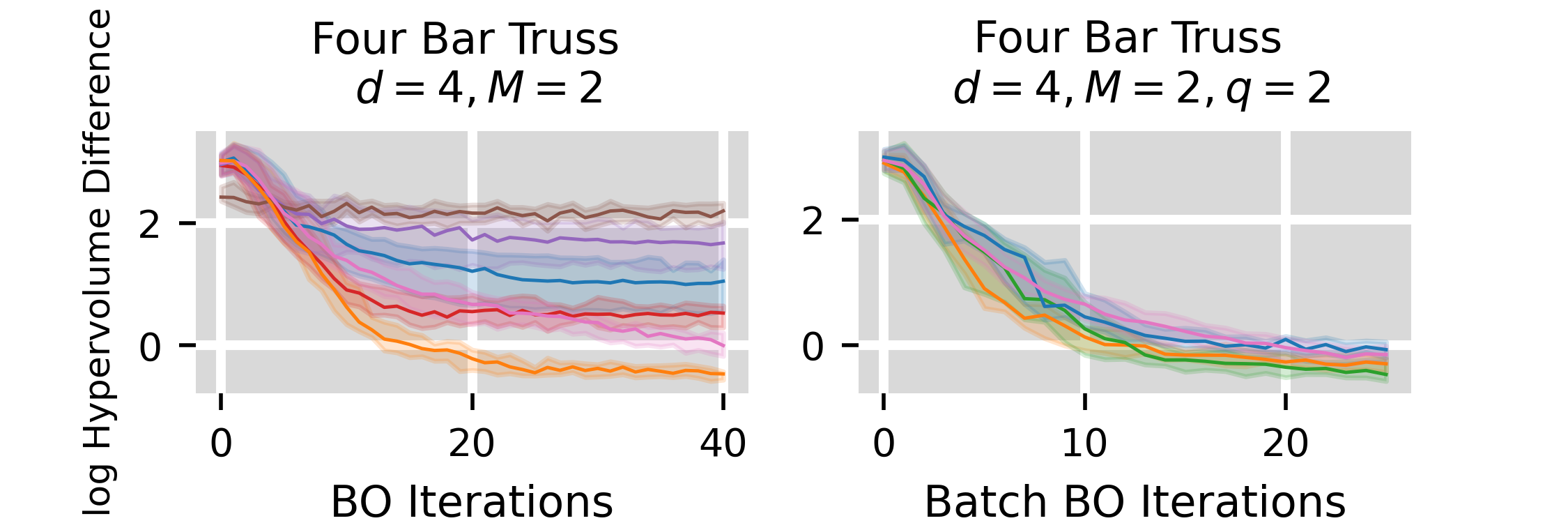} 
\caption{Four Bar Truss experimental results (with the same legend as Fig.~\ref{fig:os_benchmark}).}
\label{fig:fourbartruss}
\end{figure}

The results are depicted in Fig.~\ref{fig:fourbartruss}. For the sequential case, \{PF\}$^2$ES demonstrates the overall fastest convergence speed. While all acquisition functions tend to have similar performance in the batch scenario. 

\textbf{Disc Brake Design}
Finally, we consider a real-life disc brake design problem (CMOO) \citep{tanabe2020easy, ray2002swarm}. Here, the objectives are the mass of the brake and the achieved stopping time. The design variables are the inner radius and outer radius of the discs, the engaging force, and the number of friction surfaces. Four constraints are presented in this problem: the minimum distance between the radii, maximum length of the brake, pressure, temperature, and torque limitations. 

\begin{figure}[t]
  \includegraphics[width=1\linewidth]{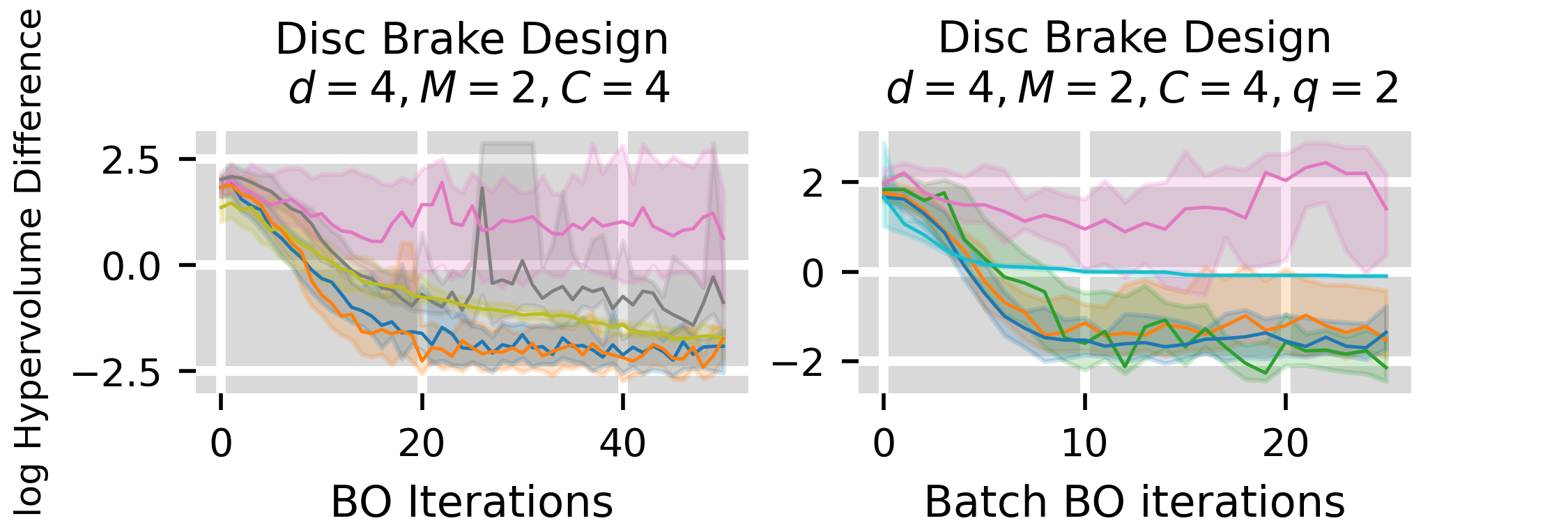} 
\caption{Disc Brake Design experimental results (with the same legend as Fig.~\ref{fig:os_benchmark}).}
\label{fig:discbrakedesign}
\end{figure}

The results are depicted in Fig.~\ref{fig:discbrakedesign}. \{PF\}$^2$ES demonstrate the overall fast convergence speed, while \{PF\}$^2$ES-KB and qEHVI are slightly better than q-\{PF\}$^2$ES in terms of stable performance. 

\section{CONCLUSION}
We have presented a new information-theoretic acquisition function $\{\text{PF}\}^2$ES. By using a variational lower bound to the mutual information, $\{\text{PF}\}^2$ES provides efficient (batch) multi-objective optimization for problems with unknown constraints. Extensive experimentation demonstrates the competitive performance of $\{\text{PF}\}^2$ES over other entropy-based methods. We also advocate the advantages of \{PF\}$^2$ES to other greedy acquisition functions (e.g., EHVI), i.e., it is free from the requirement of a proper configuration of a reference point, which arguably is often unknown for real-life applications and can affect the performance of acquisition functions to a large extent.

\textbf{Limitations and Future Work} The current out-of-sample recommendation is based on the posterior mean of the GP. Hence, its performance depends on the modeling accuracy of the GP itself which means it can suffer from the curse of input dimensionality. Furthermore, the current out-of-sample recommendation strategy can be extended to a Bayesian version. Future research will focus on tackling these aspects and making the parallelization scalable to much larger batch sizes.

\section{SOCIAL IMPACT}
This paper presents a fundamental approach for multi-objective Bayesian Optimization and has no direct societal impact or ethical consequences. It can be used in generic real-life applications from, e.g., machine learning to engineering. Of course, the applications themselves can have an impact. As a Bayesian Optimization technique, \{PF\}$^2$ES aims for identifying the Pareto frontier hopefully with less expensive black-box function evaluations (e.g., expensive biomedical experiments, power-consuming simulations), hence contributing to less energy consumption.

\section*{Acknowledgments}
This research receives funding from the Flemish Government under the “Onderzoeksprogramma Artifciële Intelligentie (AI) Vlaanderen” programme and the “Fonds Wetenschappelijk Onderzoek (FWO)” programme, and Chinese Scholarship Council (grant number 201906290032). We sincerely thank Victor Picheny for the insightful comment on the necessity of considering the effect of spacing between discrete Pareto points, which has inspired proposition 1 and Algorithm 1. We also gratefully thank anonymous reviewers for providing extensive comments for improving the paper. 


{
\renewcommand{\clearpage}{} 
\bibliography{reference.bib}
}

\appendix
\onecolumn

\setcounter{equation}{14}
\setcounter{figure}{4}
\newpage
\section{RELATIONSHIP WITH THE MULTI-OBJECTIVE PROBABILITY OF IMPROVEMENT}\label{App: relationship_with_pi} 

We provide additional insights for $\{\text{PF}\}^2$ES by linking Eq.~6 to the Multi-objective Probability of Improvement (MOPI) \citep{yang2019efficient}. First, we define the following generic formulation of MOO acquisition functions:

\begin{equation}
    \alpha_{MOO} =  \int_{E}g\left(\int_{F}\ell(F, E)p(F)dF\right)p(E)dE,
\label{Eq: gen_moo_def}
\end{equation}
\noindent where $\ell(\cdot)$ and $g(\cdot)$ are arbitrary functions, and $E$ and $F$ are random events. Evidently, $\{\text{PF}\}^2$ES is a special case obtained by setting $E:=\mathcal{F}$, $F:=\bm{h}_{\bm{x}}$, $\ell(\bm{h}, \mathcal{F}) :=  \mathbbm{1}(\bm{h}_{\bm{x}} \in A(\mathcal{F}))$,  and $g(\cdot) := -log(1 - \cdot)$. Here $\mathbbm{1}(\cdot)$ is the indicator function.


Subsequently, by setting $\ell(E, F) :=  \mathbbm{1}(\bm{h}_{\bm{x}} \in \tilde{A}(\mathcal{F}))$ (i.e., $F:=\bm{h}_{\bm{x}}$ and $E := \mathcal{F}$), $g(\cdot)$ as the identity function, and specifying a Dirac delta distribution over the Pareto frontier $\mathcal{F}$, we see that the inner integration $\int_{F}\ell(F, E)p(F)dF$ for both $\{\text{PF}\}^2$ES and MOPI evaluates to $p(\bm{h}_{\bm{x}} \in \tilde{A}(\mathcal{F}))$. Therefore, as $ g_{\{\text{PF}\}^2\text{ES}}(\cdot)$ and $g_{MOPI}(\cdot)$ are both monotonic increasing over $\left(0, 1\right)$, the acquisition functions have the same maximizer, i.e., where $p(\bm{h}_{\bm{x}} \in \tilde{A}(\mathcal{F}))$ is largest. The constraint case (i.e., remark 1.2) is obtained by assuming independency of $\bm{f}$ and $\bm{g}$.

\section{MITIGATING THE CLUSTERING ISSUE} \label{App: clustering_issue}
We first demonstrate the \textit{clustering issue} on the inverted VLMOP2 problem \citep{van1999multiobjective}, i.e., we adopt VLMOP2 for maximization by taking the negative of each objective function. For illustrative purposes, the parameters for extracting the Pareto frontier in  $\tilde{\alpha}_{\{\text{PF}\}^2\text{ES}}(\tilde{A}(\tilde{\mathcal{F}}))$ are set to $\vert \bm{\tilde{F}} \vert  = 5$ and $\vert \tilde{\mathcal{F}} \vert = 5\ \forall \tilde{\mathcal{F}} \in \vert \bm{\tilde{F}} \vert$.

\begin{figure*}[h]
\begin{subfigure}[t]{3.8cm}
  \centering
  \includegraphics[width=1.1\linewidth]{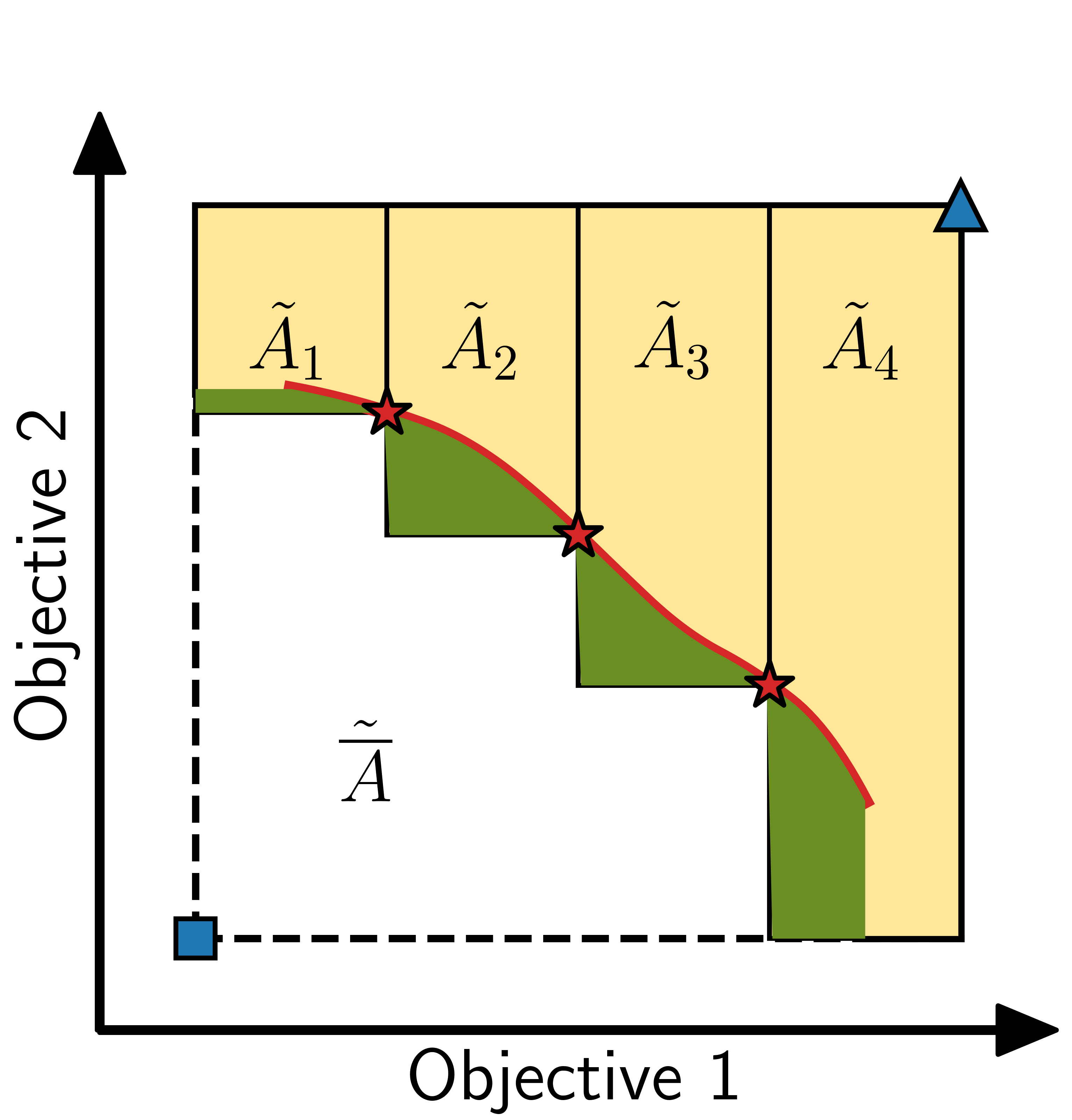}  
  \caption{}
  \label{fig:cluster sub-first}
\end{subfigure}
\begin{subfigure}[t]{3.8cm}
\hspace*{0.4cm}
  \centering
  \includegraphics[width=1.0\linewidth]{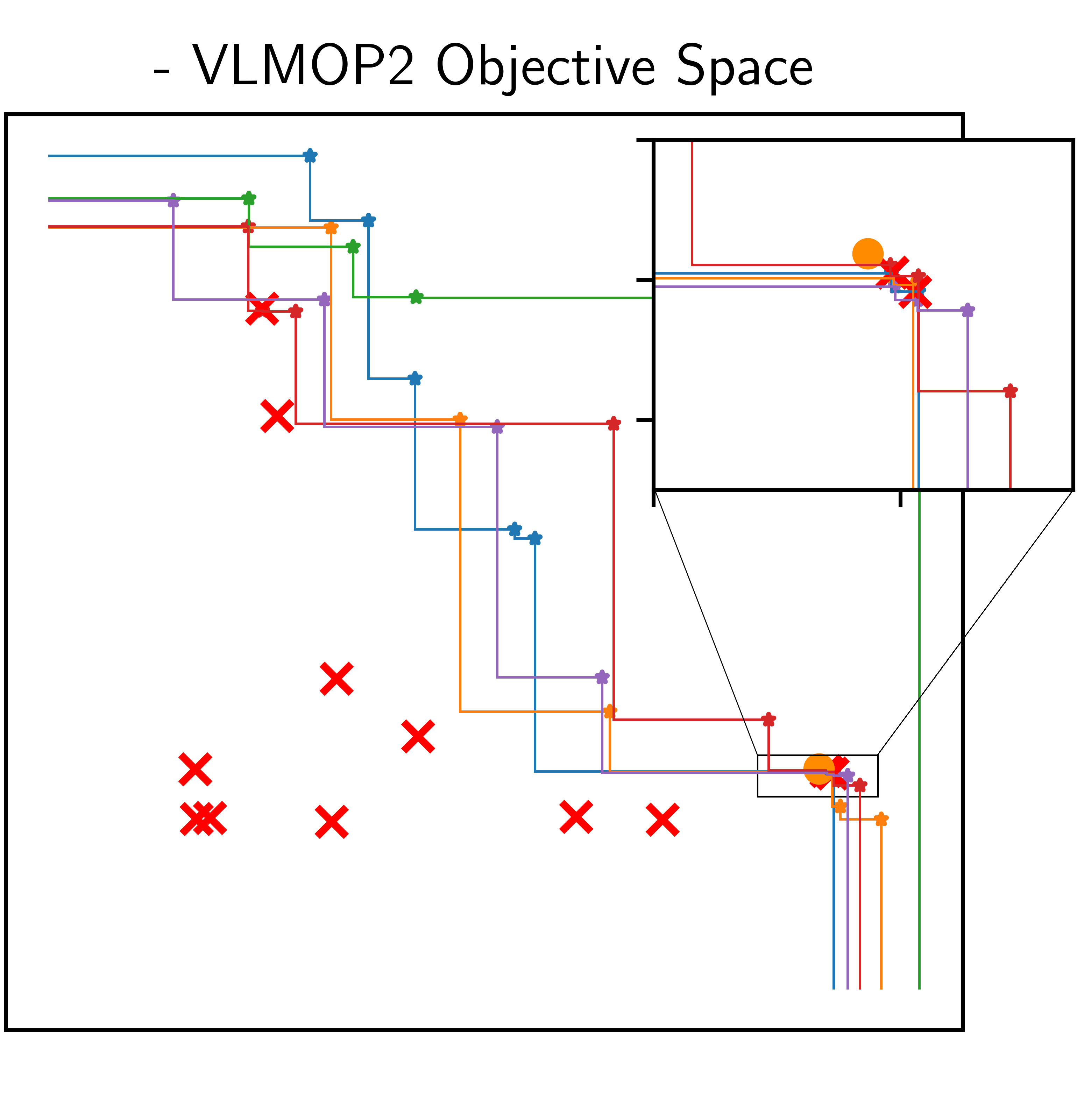}  
  \caption{}
  \label{fig:cluster sub-second}
\end{subfigure}
\begin{subfigure}[t]{3.8cm}
\hspace*{0.7cm}
  \centering
  \includegraphics[width=1.2\linewidth]{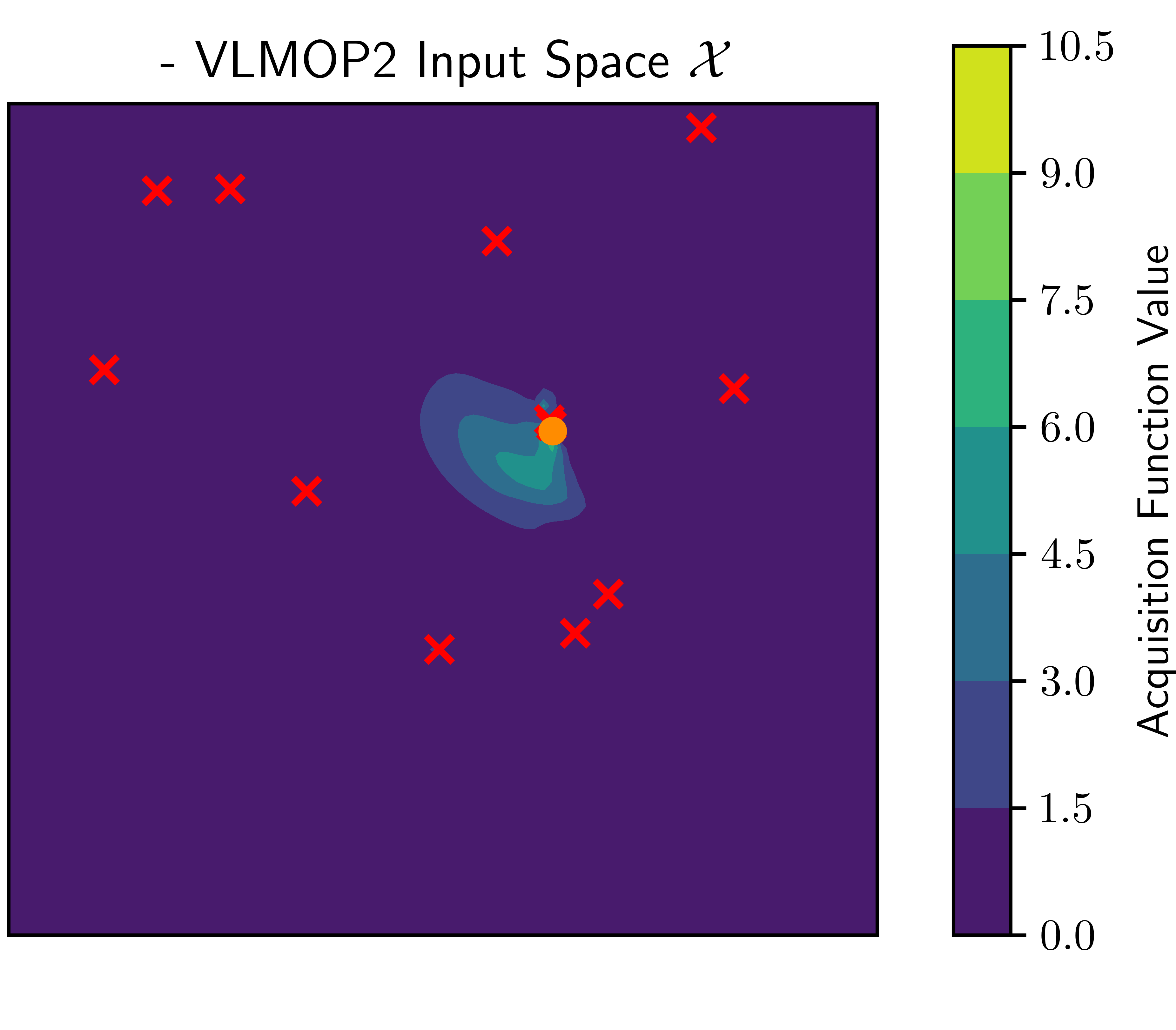}  
  \caption{}
  \label{fig:cluster sub-third}
\end{subfigure}
  \begin{subfigure}[t]{2.5cm}
  \hspace*{1.5cm}
  \centering
  \raisebox{1cm}{\includegraphics[width=1\linewidth]{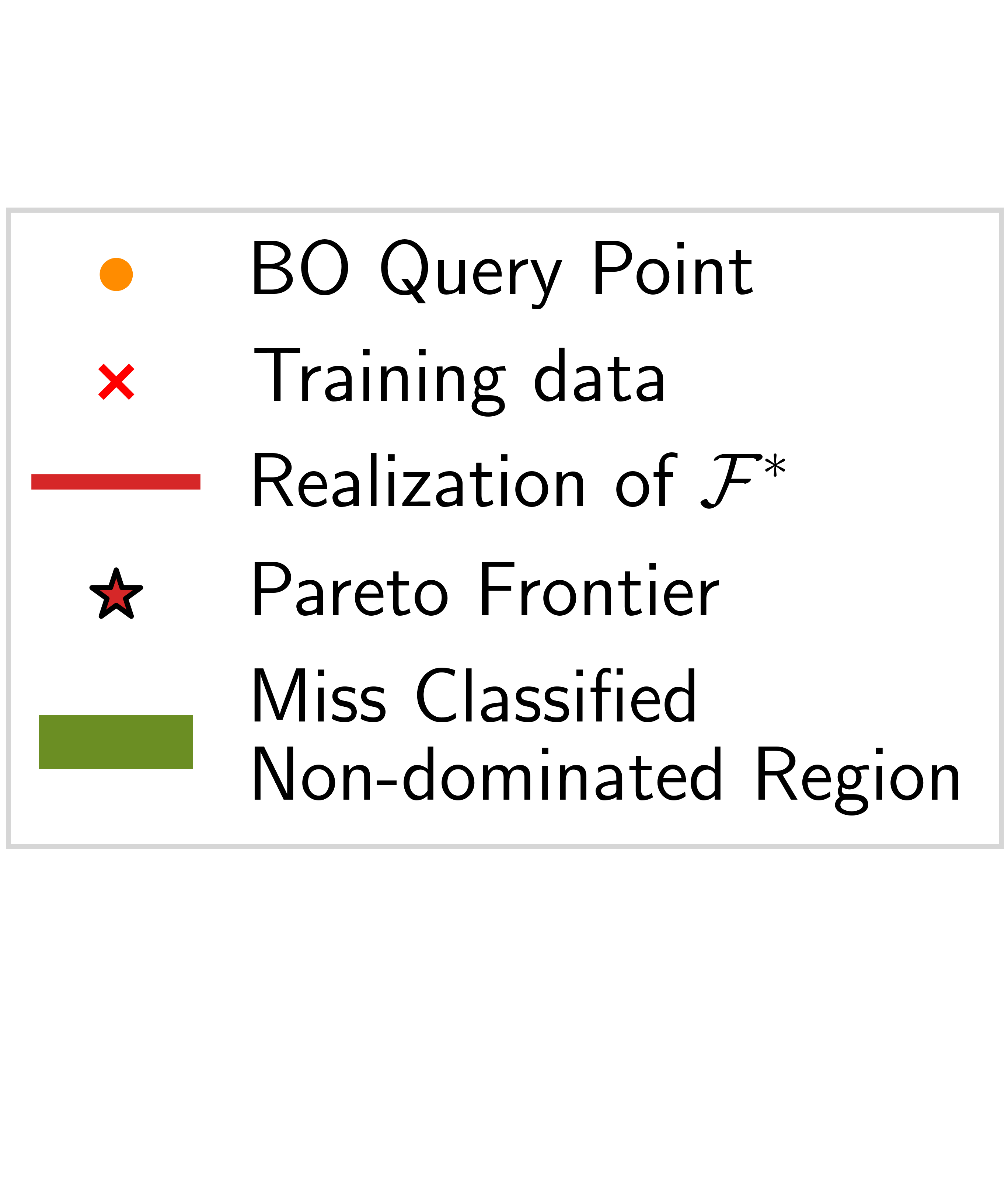}}
\end{subfigure}
\caption{The hypervolume partition strategy for a discrete Pareto frontier $\tilde{\mathcal{F}}$ is an overestimation of the actual non-dominated space $A$. a) Part of the dominated region has been misclassified as non-dominated (green area).  b) This overestimation will make the clustering more severe when calculating \{PF\}$^2$ES based on $\tilde{A}(\tilde{\mathcal{F}})$: For a certain iteration, given the MC samples of $\tilde{\mathcal{F}}$, \{PF\}$^2$ES favors the 'false positive' location in the output space ($\bm{\sigma}_{\bm{x}} = [0.00158, 0.00112]$) that is certain (high confidency) to (marginally) improve. c) This causes a \textit{clustering issue} as candidates very close to existing training data points are preferred.}
\label{fig:greedy_issue}
\end{figure*}

A contour of \{PF\}$^2$ES based on the approximation of Eq.~8 is depicted in Fig.~\ref{fig:greedy_issue}, which shows that the new candidate point is believed to make an improvement with high confidence (due to the small GP posterior variance) in 4 of the 5 Pareto frontier samples. Unfortunately, the improvement is marginal as the predicted observation is located in the (relatively small) \textit{false positive} region introduced by the hypervolume decomposition $\mathcal{P}$, see Fig.~\ref{fig:cluster sub-first}. In practice, \{PF\}$^2$ES will tend to keep sampling in a region that it is confident to make tiny (or no) improvements, causing the sampled points to be densely clustered in a small region of the input domain $\mathcal{X}$.

\begin{figure*}[h]
\begin{subfigure}[t]{3.8cm}
  \centering
  \includegraphics[width=1.1\linewidth]{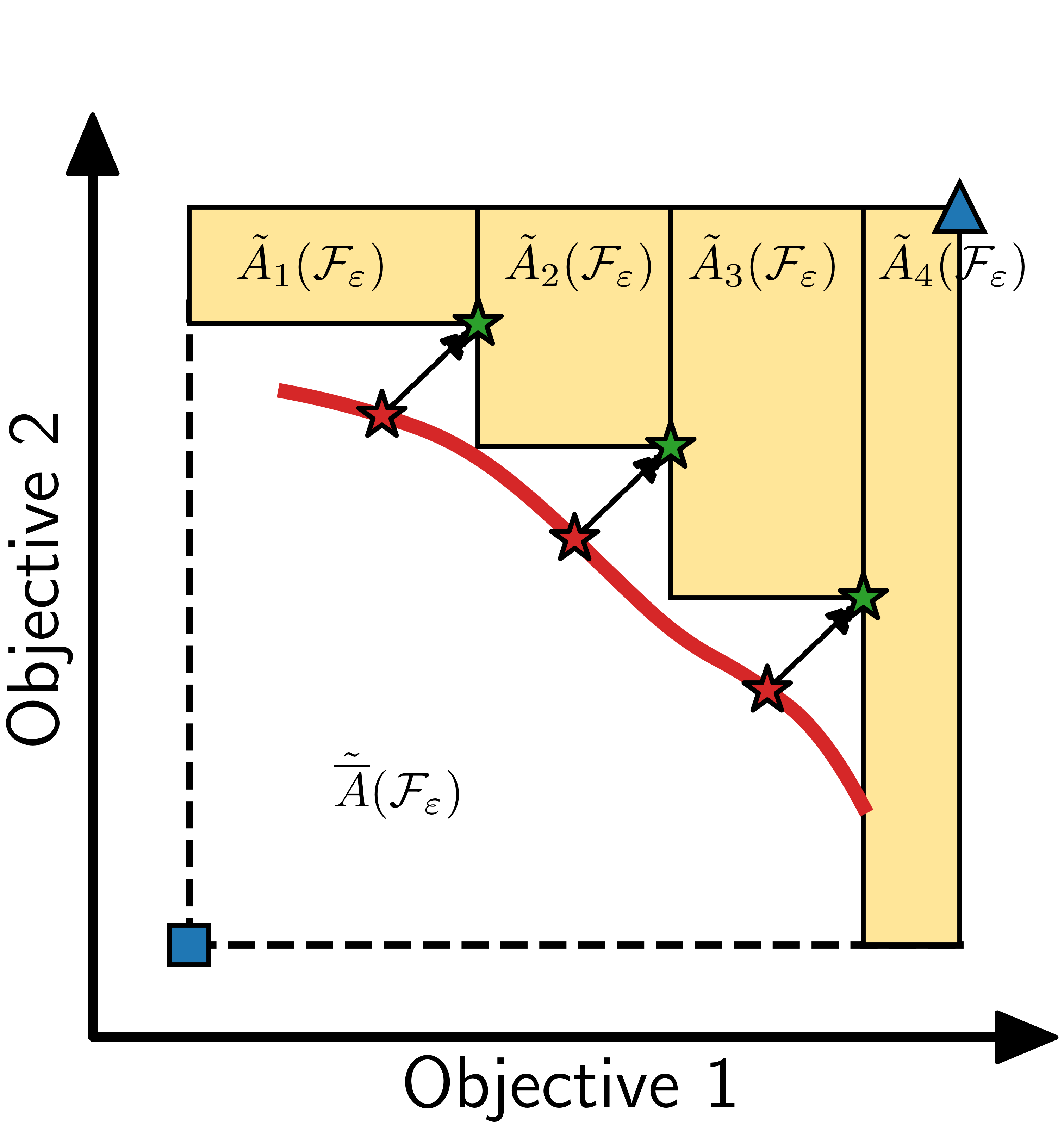}  
  \caption{}
  \label{fig:solve cluster sub-first}
\end{subfigure}
\begin{subfigure}[t]{3.5cm}
\hspace*{0.4cm}
  \centering
  \includegraphics[width=1.0\linewidth]{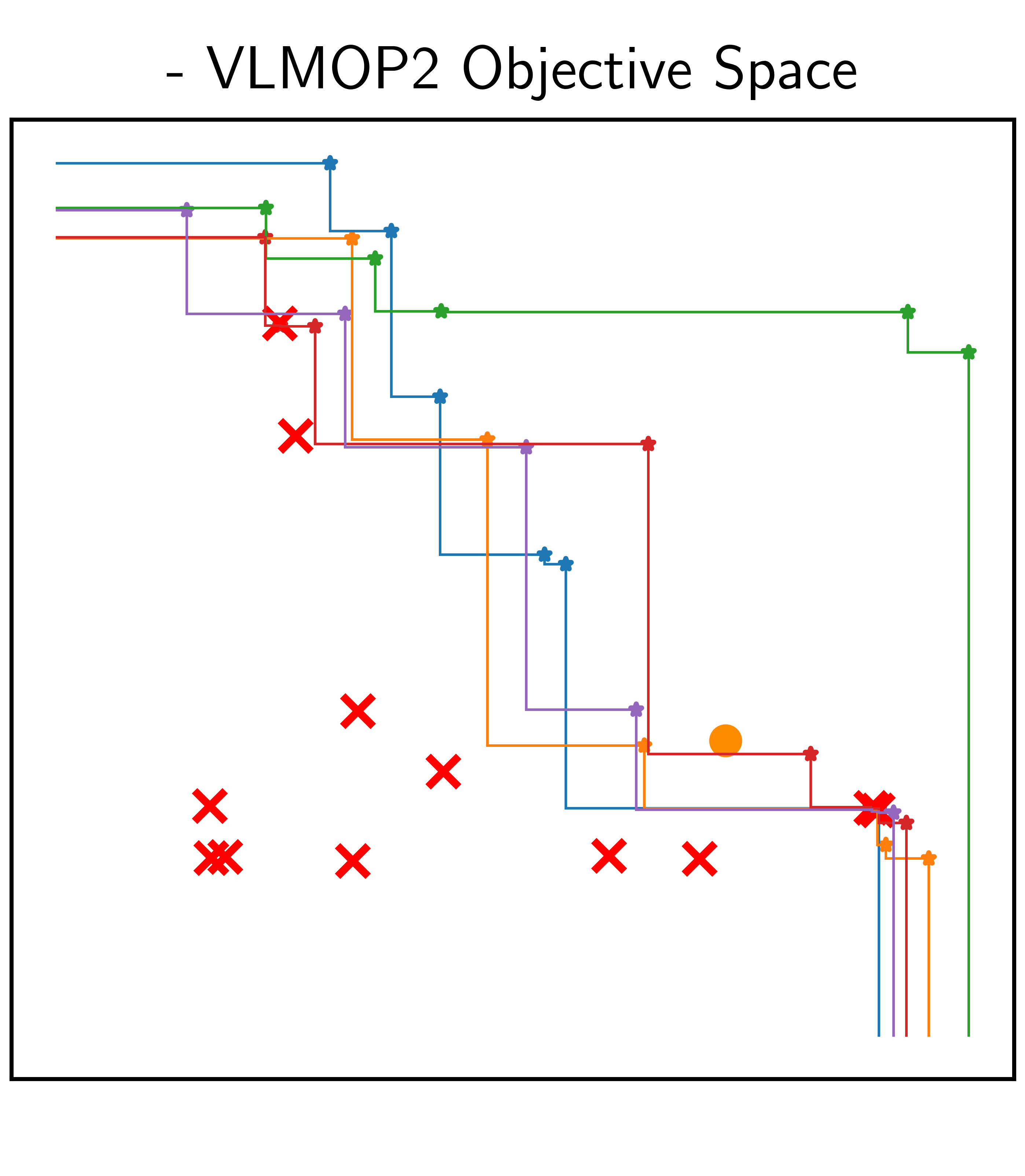}  
  \caption{}
  \label{fig:solve cluster sub-second}
\end{subfigure}
\begin{subfigure}[t]{3.8cm}
\hspace*{0.7cm}
  \centering
  \includegraphics[width=1.2\linewidth]{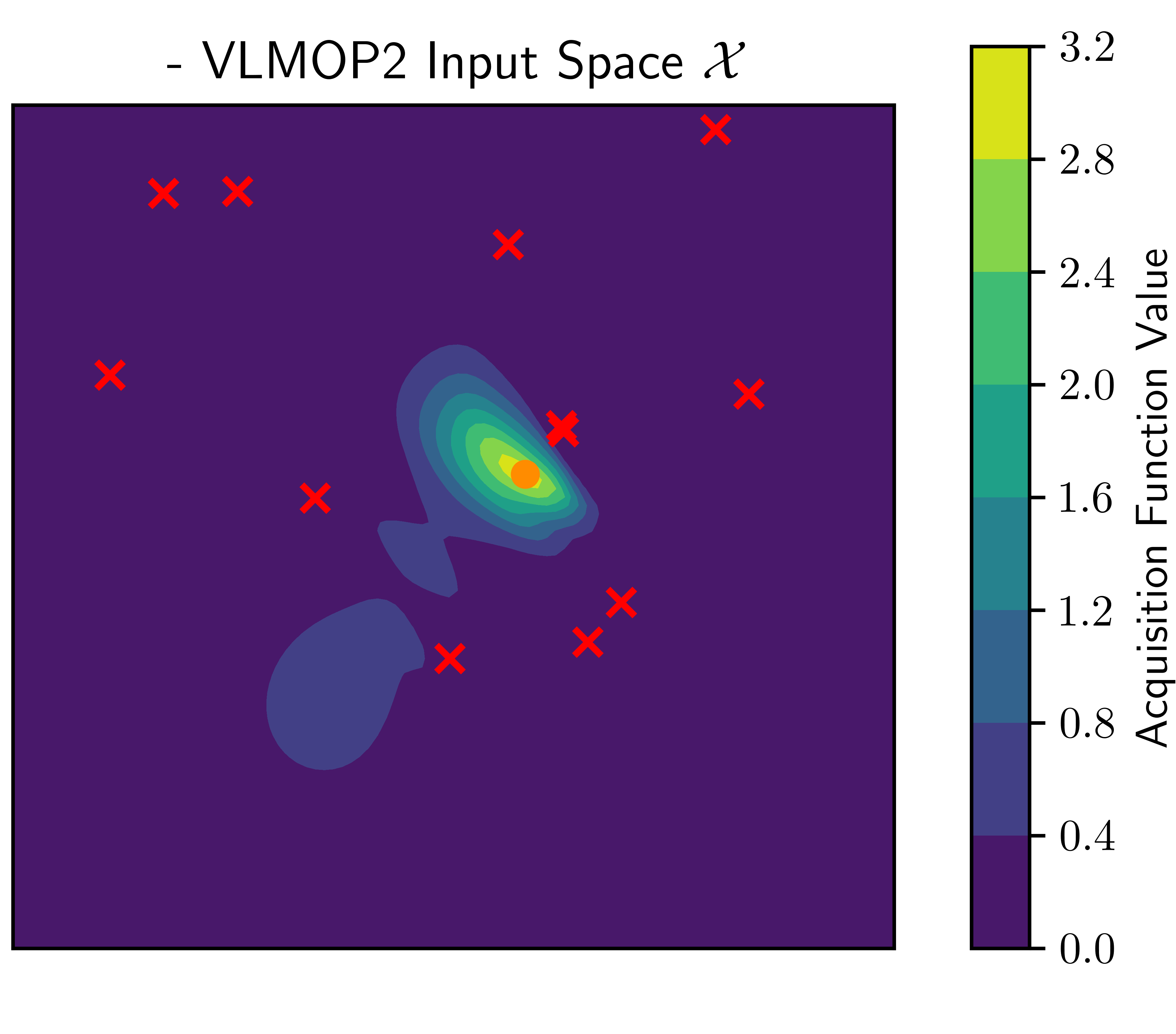}  
  \caption{}
  \label{fig:solve cluster sub-third}
\end{subfigure}
  \begin{subfigure}[t]{2.5cm}
  \hspace*{1.5cm}
  \centering
  \raisebox{0cm}{\includegraphics[width=1\linewidth]{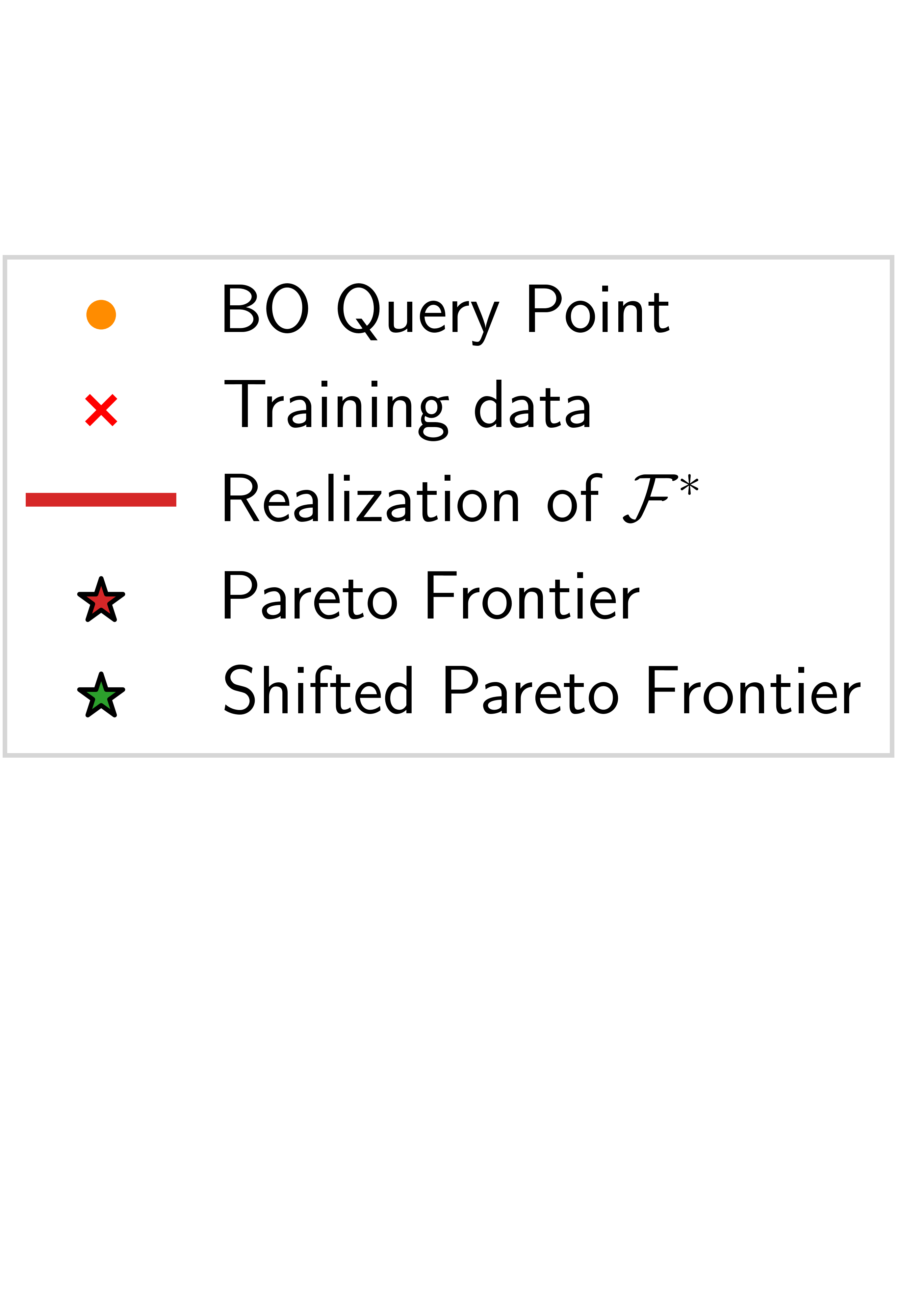}}
\end{subfigure}
\caption{The $\varepsilon$ approach for mitigating clustering issues. a) The non-dominated region is extracted from the partitioning $\mathcal{P}$ based on the shifted Pareto frontier $\tilde{\mathcal{F}}_{\varepsilon}$.  b) Relative to Fig.~\ref{fig:greedy_issue}, the conservative estimation of $\tilde{A}$ has encouraged more exploration, i.e., candidate ($\bm{\sigma}_{\bm{x}} = [0.053400, 0.032628]$). c) The corresponding exploration around the Pareto set in the input space.}
\label{fig:solve_greedy_issue}
\end{figure*}

In order to mitigate this issue and encourage the acquisition function to focus more on where the Pareto frontier is uncertain, we utilize a parameter $\varepsilon$ to shift the Pareto frontier approximation $\mathcal{\tilde{F}}$ and shrink the false positive region. The effect is demonstrated in Fig.~\ref{fig:solve_greedy_issue}. It can be seen that the modified acquisition function $\tilde{\alpha}_{\{\text{PF}\}^2\text{ES}}(\tilde{A}(\tilde{\mathcal{F}}_{\varepsilon}))$ tend to be more explorative than the naive approach $\tilde{\alpha}_{\{\text{PF}\}^2\text{ES}}(\tilde{A}(\tilde{\mathcal{F}}))$ (i.e., based on Eq.~8).

Finally, this significant performance improvement is validated by the sensitivity analysis of $\varepsilon$ in Fig.~\ref{fig:sensitivity_epsilon}, where the performance of the original approach $\tilde{\alpha}_{\{\text{PF}\}^2\text{ES}}(\tilde{A}(\tilde{\mathcal{F}}))$ is denoted as $\varepsilon = 0$. 

\section{PARALLEL VERSION q-\{PF\}$^2$ES}\label{app:parallelization}
\subsection{Derivation of q-\{PF\}$^2$ES}\label{app:derive_parallelization}

\begin{figure}[h]
  \centering
  \includegraphics[width=0.35\linewidth]{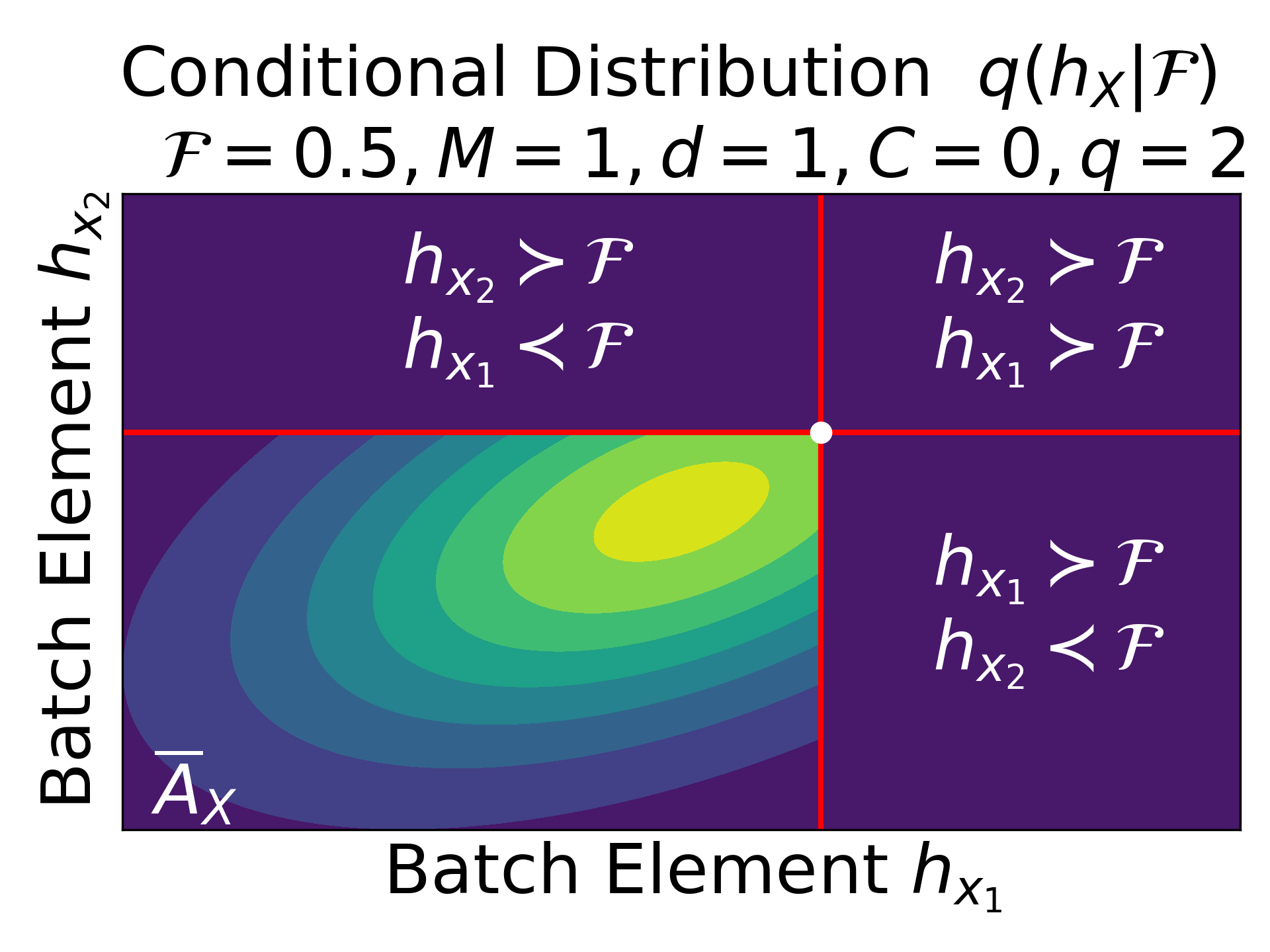}  
\caption{The batch conditional probability density $q(\bm{h}_{\bm{X}} \vert \mathcal{F})$ for $M=1$, $q=2$ and $C = 0$. The problem degenerates to a batch single-objective optimization problem, where $\mathcal{F}$ collapses to a scalar value (the maximum) and the dominance operation is reduced to $>$.  In this scenario, by conditioning on $\mathcal{F}$ (white dot), the objective space can be partitioned into 2 different subspaces: the union of the non-dominated feasible regions $A_{q}$ and its complement $\overline{A}_{q}$, see bottom-left plot. The condition of being in the region of $\overline{A}_{q}$ can be interpreted as: \textit{no point inside the batch is larger than $\mathcal{F}$}. The concept can be generalized for arbitrary $M$, $q$ and $C$.}
\label{fig:demo_of_batch_conditional_prob}
\end{figure}

Consider the random variable $\bm{h}_{\bm{X}} = \{\bm{h}_{\bm{x}_1}, ..., \bm{h}_{\bm{x}_q}\}$, i.e., the objective-constraint observations that could arise from a parallel evaluation of a batch of $q$ points. In this scenario, we want to allocate a batch of $q$ points $\textbf{X}$ that provide the most mutual information about the (feasible) Pareto frontier. With Eq.~4, we have:

\begin{equation}
\begin{aligned}
    I(\mathcal{F}; \bm{h}_{\bm{X}}) \geq \mathbb{E}_{\mathcal{F}}\left[\int_{\bm{h}_{\bm{x}}}p(\bm{h}_{\bm{X}} \vert \mathcal{F})\text{log} \frac{ q(\bm{h}_{\bm{X}} \vert \mathcal{F})}{p(\bm{h}_{\bm{X}})}d\bm{h}_{\bm{X}} \right]
\end{aligned}
\end{equation}
Given the Pareto frontier $\mathcal{F}$, we set the variational approximation of the ground-truth conditional distribution $p(\bm{h}_{\bm{X}} \vert \mathcal{F})$ as:

\begin{equation}
q(\bm{h}_{\bm{X}} \vert \mathcal{F})  = \left\{
\begin{aligned}
&\frac{p(\bm{h}_{\bm{X}} )}{Z_{{\overline{A}_{q}}}} \quad \bm{h}_{\bm{X}} \in \overline{A}_{q}\\
&0 \quad\quad\quad\  \text{otherwise,}\\
\end{aligned} 
\right.
\label{Eq: batch_conditional_prob}
\end{equation}

\noindent where $A_{q} \cup \overline{A}_{q} = \mathbb{R}^{q(M+C)}$,  $A_{q} \in \mathbb{R}^{q(M+C)}$ is the union of the non-dominated feasible regions:
$\{\bm{h}_{\bm{X}} \in \mathbb{R}^{q(M+C)} \vert \exists i \in \{1, ..., q\}, \bm{h}_{{\bm{x}}_i} \in A\}$ and $\overline{A}_{q} \in \mathbb{R}^{q(M+C)}$ is its complement $\{\bm{h}_{\bm{X}} \in \mathbb{R}^{q(M+C)} \vert \forall i \in \{1, ..., q\}, \bm{h}_{{\bm{x}}_i} \in \overline{A}\}$. An intuitive illustration of $A_{q}$ and $q(\bm{h}_{\bm{X}} \vert \mathcal{F})$ is provided in Fig.~\ref{fig:demo_of_batch_conditional_prob}. Note, the practical interpretation of the conditional probability Eq.~\ref{Eq: batch_conditional_prob} is that, given $\mathcal{F}$, the outcome of the objective-constraint observation is expected to \textbf{have zero probability of being in the feasible non-dominated region $A$ for any candidate point within the batch}, as otherwise $\mathcal{F}$ is not a valid Pareto frontier.

Following the same steps to motivate $\{\text{PF}\}^2$ES, we arrive at the following MC estimate of the resulting variational bound and the proposed batch extension of $\{\text{PF}\}^2$ES:

\begin{equation}
\begin{aligned}
  & \alpha_{\text{q-}\{\text{PF}\}^2\text{ES}} \approx   - \frac{1}{ \vert \tilde{\mathbf{F}}\vert }\sum_{\tilde{\mathcal{F}} \in \tilde{\mathbf{F}}} \left[\text{log}\left(1 - Z_{\tilde{A}_{q}(\tilde{\mathcal{F}}_{\bm{\varepsilon}})}\right)\right],
\end{aligned}
\end{equation}

\noindent where $Z_{\tilde{A}_{q}}:= \int_{\tilde{A}_{q}} p(\bm{h}_{\bm{X}})d\bm{h}_{\bm{X}} $. i.e., the probability that \textbf{there exists at least one batch element $\bm{h}_{\bm{x}_l} \in \bm{h}_{\bm{X}}$ such that $\bm{h}_{\bm{x}_l} \in \tilde{A}=\{\tilde{A}_1, ..., \tilde{A}_{N_p}\}$}. We omit the notion $\tilde{\mathcal{F}}_{\bm{\varepsilon}}$ here as well as in Eq.~\ref{eq: even},\ref{eq: batch_mopi} for notational simplicity.

To demonstrate the calculation of $Z_{\tilde{A}_{q}}$, it is convenient to utilize the notion of events. Define $\omega(\tilde{A}, h_{\bm{X}})$ representing the event that there exists at least one batch element $\bm{h}_{\bm{x}_l} \in \bm{h}_{\bm{X}}$ such that $\bm{h}_{\bm{x}_l} \in \tilde{A}$ , it can be seen that $Z_{\tilde{A}_{q}} = p(\omega(\tilde{A}, h_{\bm{X}}))$. Now, since $\tilde{A} = \{\tilde{A}_1, ..., \tilde{A}_{N_p}\}$, we know that $\omega(\tilde{A}, h_{\bm{X}}) = \bigcup_{i=1}^{N_p}\omega(\tilde{A}_i, h_{\bm{X}})$, where $\omega(\tilde{A}_i, h_{\bm{X}})$ represents there exists at least one batch element $\bm{h}_{\bm{x}_{l}} \in \bm{h}_{\bm{X}}$ such that $\bm{h}_{\bm{x}_{l}} \in \tilde{A}_i$. It can be also seen that $\omega(\tilde{A}_i, h_{\bm{X}}) = \bigcup_{l=1}^q\omega(\tilde{A}_i, h_{{\bm{x}_l}})$ where $\omega(\tilde{A}_i, h_{{\bm{x}_l}})$ is the $l$th batch element $h_{{\bm{x}_l}}$ is in $\tilde{A}_i$. Combine the two unions, and the fact that $\omega(\tilde{A}_i, h_{{\bm{x}_l}}) = \prod_{k=1}^{M+C} \left(  \mathbbm{1} (A_{i_l}^k \leq \bm{h}_{\bm{x}_{l_j}}^k \leq A_{i_u}^k)  \right) $ representing $\bm{h}_{\bm{x}_{l}}$ is within the decomposed hypercube $\tilde{A}_i$, the event of $\omega(\tilde{A}, h_{\bm{X}})$ can be written as:

\begin{equation}
    \omega(\tilde{A}, h_{\bm{X}}) = \bigcup_{i=1}^{N_p}\left(\bigcup_{l=1}^{q} \omega(\tilde{A}_i, h_{{\bm{x}_l}})\right) = \bigcup_{i=1}^{N_p}\left(\bigcup_{l=1}^{q} \left( \prod_{k=1}^{M+C} \left(  \mathbbm{1} (\tilde{A}_{i_l}^k \leq \bm{h}_{\bm{x}_{l_j}}^k \leq \tilde{A}_{i_u}^k) \right) \right) \right)
    \label{eq: even}
\end{equation}

Finally, we are able to MC approximate the probability of this event:

\begin{equation}
\begin{aligned}
  Z_{\tilde{A}_{q}} & = p\left(\omega(\tilde{A}, \bm{h_X})\right) \\ & \approx \frac{1}{N_{MC}}\sum_{j=1}^{N_{MC}}\left(\bigcup_{i=1}^{N_p}\left(\bigcup_{l=1}^{q} \left(\prod_{k=1}^{M+C} \left(  \mathbbm{1} (\tilde{A}_{i_l}^k \leq \bm{h}_{\bm{x}_{l_j}}^k \leq \tilde{A}_{i_u}^k)  \right) \right)\right)\right) \\& \approx \frac{1}{N_{MC}}\sum_{j=1}^{N_{MC}}\left(max_i\left(max_l \left(\prod_{k=1}^{M+C} \left( \sigma\left(\frac{\bm{h}_{\bm{x}_{l_j}}^k - \tilde{A}_{i_l}^k}{\tau}\right) \cdot   \sigma\left(\frac{\tilde{A}_{i_u}^k - \bm{h}_{\bm{x}_{l_j}}^k}{\tau}\right)  \right) \right)\right)\right),
\end{aligned}
\label{eq: mc_batch_mopi}
\end{equation}

Note that the additional approximation introduced in Eq.~\ref{eq: mc_batch_mopi} is caused by the relaxation of the categorical event (imposed by the indicator function) to a continuous approximation: $\mathbbm{1} (\tilde{A}_{i_l}^k \leq \bm{h}_{\bm{x}_{l_j}}^k \leq \tilde{A}_{i_u}^k) \approx \sigma\left(\frac{\bm{h}_{\bm{x}_{l_j}}^k - \tilde{A}_{i_l}^k}{\tau}\right) \cdot   \sigma\left(\frac{\tilde{A}_{i_u}^k - \bm{h}_{\bm{x}_{l_j}}^k}{\tau}\right)$. This makes the MC approximation suitable for gradient-based optimizers.
\subsection{Monte Carlo Approximation of q-\{PF\}$^2$ES and its Demonstration}\label{app:demonstrate_parallelization}

We elaborate on how the MC approximation has been implemented here. Letting $\mathcal{\ell}: \mathbb{R}^{q(M+C)} \times A \rightarrow \mathbb{R}$  represents the \textit{utility function} that calculate $Z_{\tilde{A}_{q}}$ as in Eq. \ref{eq: batch_mopi}, we can write the MC approximated q-\{PF\}$^2$ES in terms of $\mathcal{\ell}$ as:

\begin{equation}
\begin{aligned}
  \alpha_{\text{q-}\{\text{PF}\}^2\text{ES}}  \approx \tilde{\alpha}_{\text{q-}\{\text{PF}\}^2\text{ES}} =   - \frac{1}{ \vert \tilde{\mathbf{F}}\vert }\sum_{\tilde{\mathcal{F}} \in \tilde{\mathbf{F}}} \left[\text{log}\left(1 - \frac{1}{N_{MC}}\sum_{j=1}^{N_{MC}}\mathcal{\ell}({\bm{h_X}}_j, \tilde{A}_{q}(\tilde{\mathcal{F}}_{\bm{\varepsilon}}))\right)\right],
\end{aligned}
\end{equation}

where $\tilde{\alpha}_{\text{q-}\{\text{PF}\}^2\text{ES}}$ represents the MC approximated acquisition function, we note $\tilde{A}_{q}(\tilde{\mathcal{F}}_{\bm{\varepsilon}})$ is independent of $\bm{X}$. With the aim of a consistent MC acquisition function gradient, we leverage the reparameterization trick \citep{kingma2013auto} to generate samples of $\bm{h_X}$: since $\bm{h_X} \sim \mathcal{N}(\bm{m_X}, L_{\bm{X}}L_{\bm{X}}^T)$ where $\bm{m_X} \in \mathbb{R}^{q(M+C)}$, Cholesky factor $L_{\bm{X}} \in \mathbb{R}^{q(M+C) \times q(M+C)}$ , it can be generated via ${\bm{h_X}} = \bm{m_X} + L_{\bm{X}}\lambda$, where $\lambda \sim \mathcal{N}(\bm{0}, \bm{I})$ and the \textit{base sample} sets: $\{\lambda_j\}_{j=1}^{N_{MC}}, \lambda_j \in \mathbb{R}^{q(M+C)}$ is holding through the whole process of acquisition function maximization, the largely preserved consistency can be shown via the chain rule of derivative as $\nabla_{\bm{X}}\tilde{\alpha}_{\text{q-}\{\text{PF}\}^2\text{ES}} =  - \frac{1}{ \vert \tilde{\mathbf{F}}\vert }\sum_{\tilde{\mathcal{F}} \in \tilde{\mathbf{F}}} \left[\frac{- \frac{1}{N_{MC}}\sum_{j=1}^{N_{MC}}\nabla_{\bm{X}}\mathcal{\ell}({\bm{h_X}}_j, \tilde{A}_{q}(\tilde{\mathcal{F}}_{\bm{\varepsilon}}))}{\left(1 - \frac{1}{N_{MC}}\sum_{j=1}^{N_{MC}}\mathcal{\ell}({\bm{h_X}}_j, \tilde{A}_{q}(\tilde{\mathcal{F}}_{\bm{\varepsilon}}))\right)}\right]$, where each element within the $max$ function $\left(\prod_{k=1}^{M+C} \left( \sigma\left(\frac{\bm{h}_{\bm{x}_{l_j}}^k - \tilde{A}_{i_l}^k}{\tau}\right) \cdot   \sigma\left(\frac{\tilde{A}_{i_u}^k - \bm{h}_{\bm{x}_{l_j}}^k}{\tau}\right)  \right) \right)$ is differentiable. We use quasi-Monte Carlos to generate the base samples and refer \cite{balandat2020botorch, daulton2020differentiable} for details of such an approximation.

\begin{figure}[h]
    \centering
    \begin{subfigure}[t]{0.195\textwidth}
        \centering
        \includegraphics[width=1.1\linewidth]{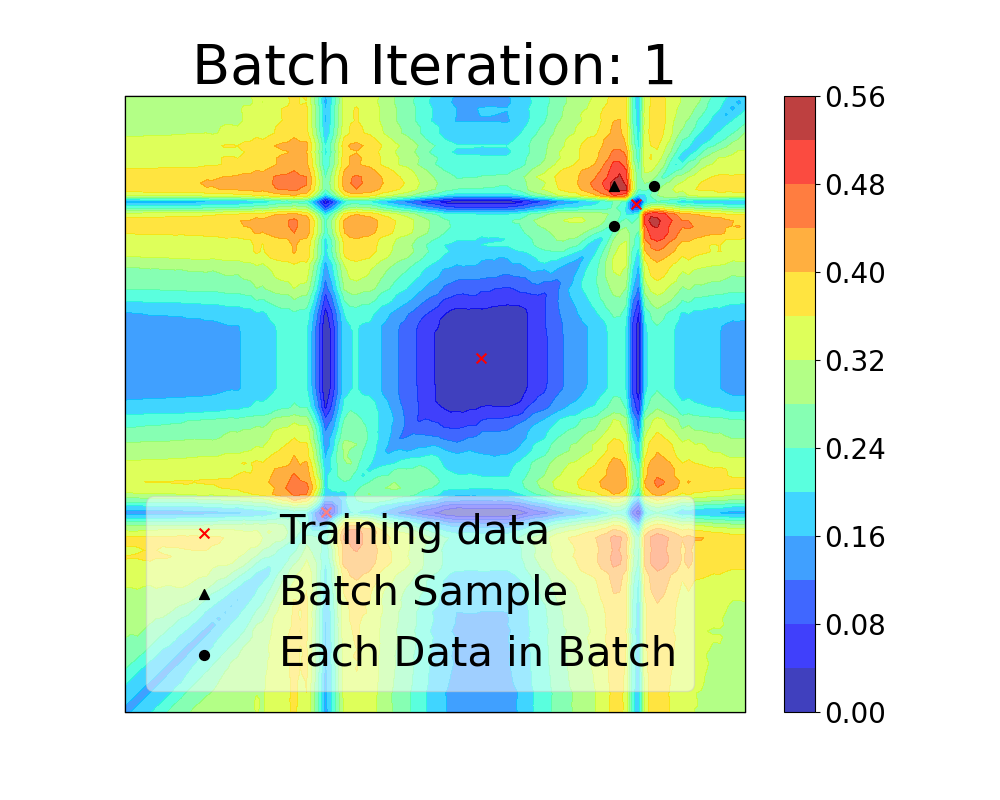} 
    \end{subfigure}
    \hfill
    \begin{subfigure}[t]{0.195\textwidth}
        \centering
        \includegraphics[width=1.1\linewidth]{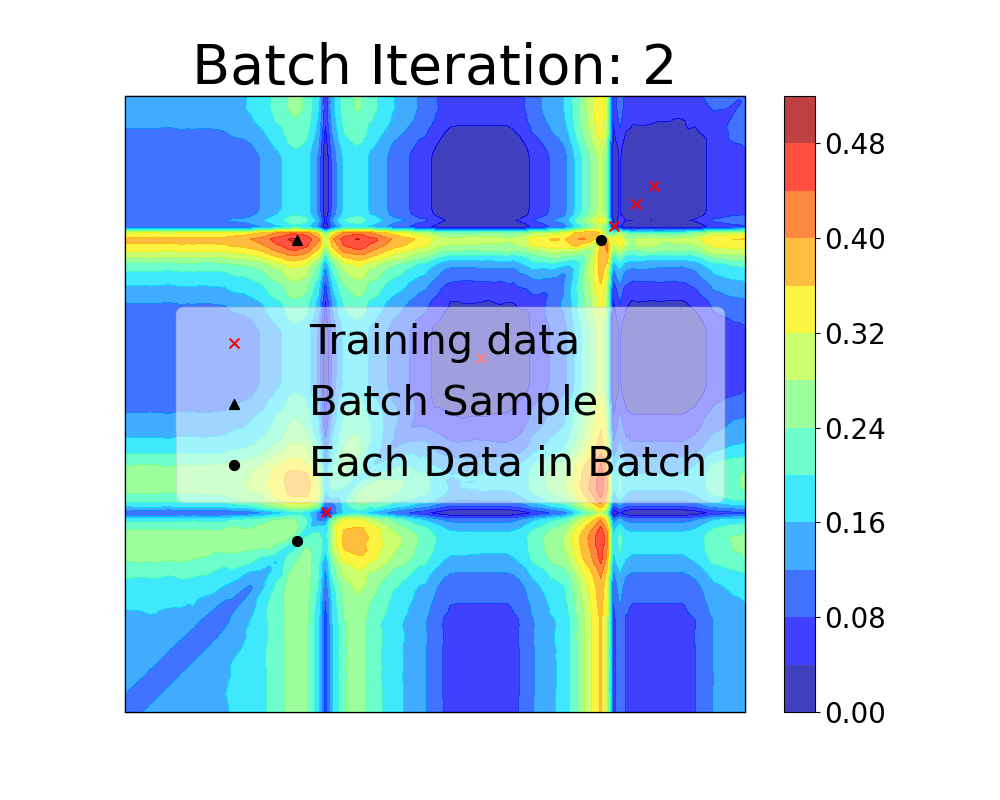} 
    \end{subfigure}
        \hfill
    \begin{subfigure}[t]{0.195\textwidth}
        \centering
        \includegraphics[width=1.1\linewidth]{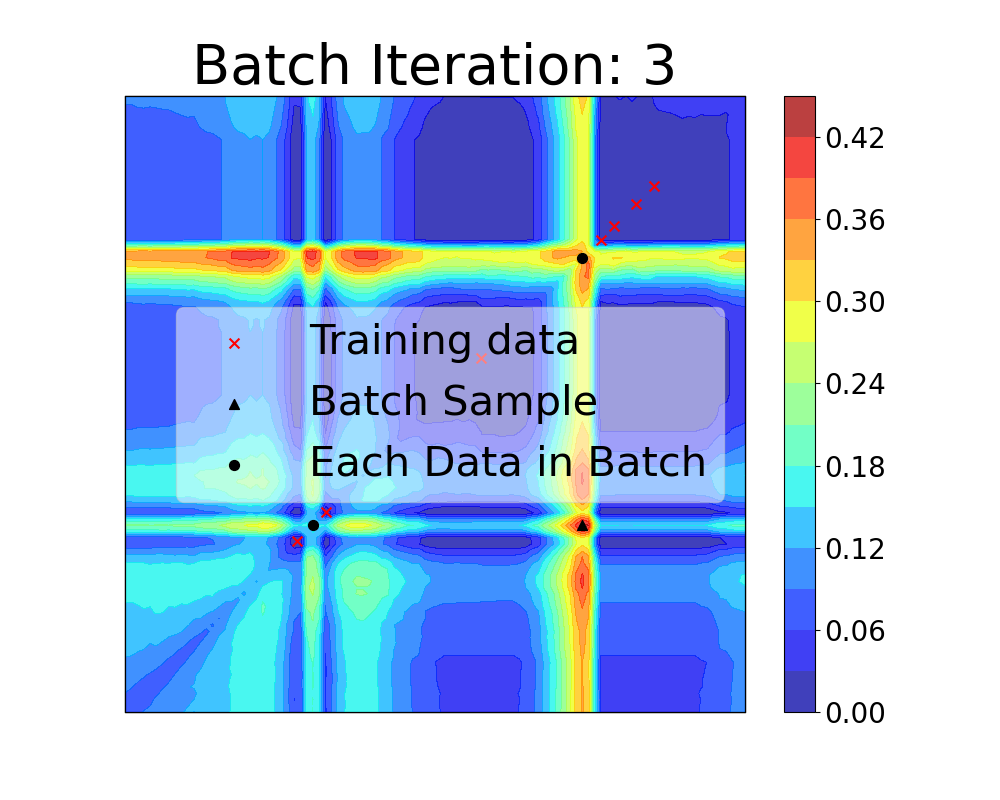} 
    \end{subfigure}
    \hfill
        \begin{subfigure}[t]{0.195\textwidth}
        \centering
        \includegraphics[width=1.1\linewidth]{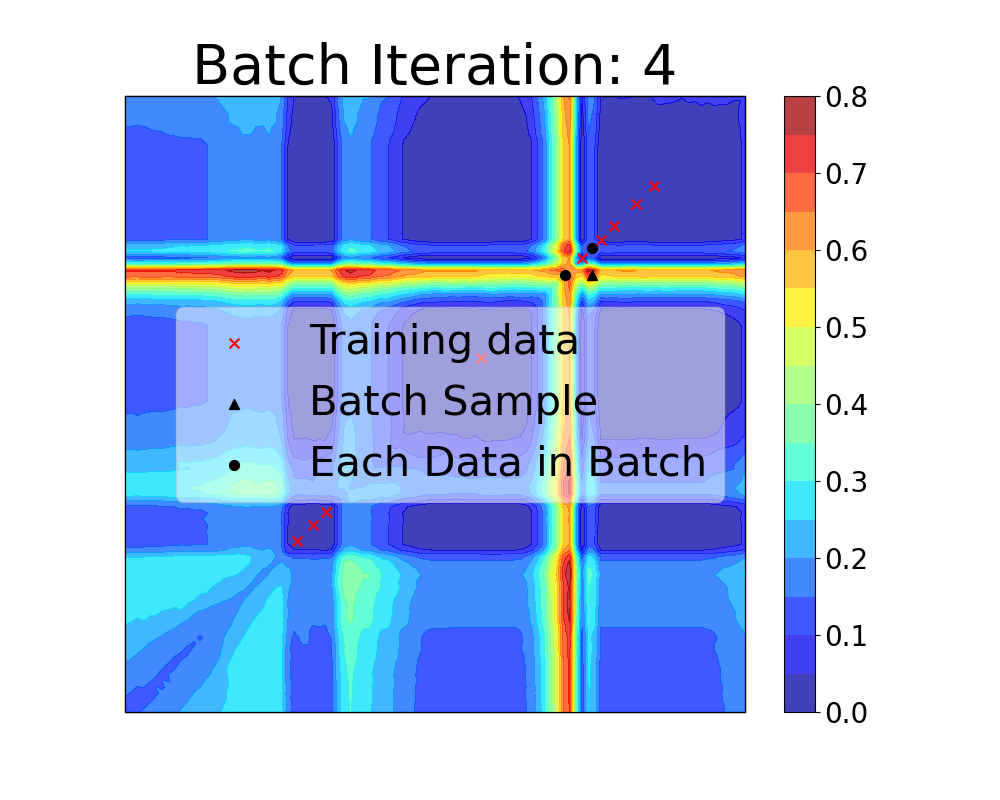} 
    \end{subfigure}
        \hfill
        \begin{subfigure}[t]{0.195\textwidth}
        \centering
        \includegraphics[width=1.1\linewidth]{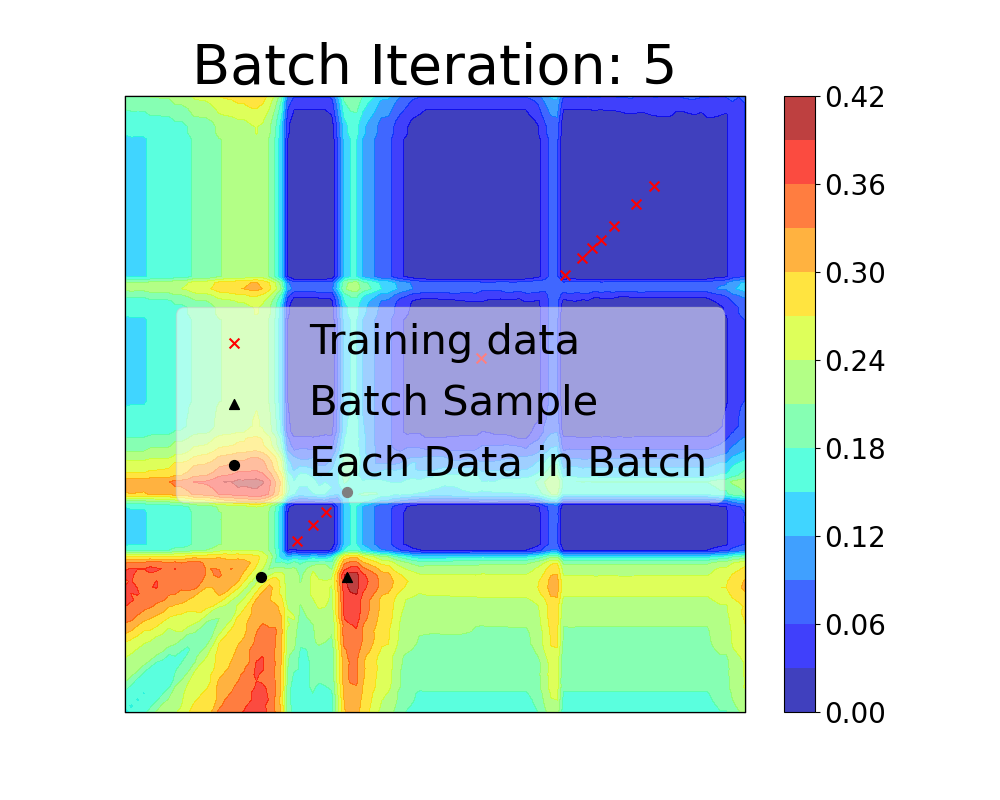} 
    \end{subfigure}
        \begin{subfigure}[t]{0.195\textwidth}
            \hspace{-5mm}
        \includegraphics[width=1.1\linewidth]{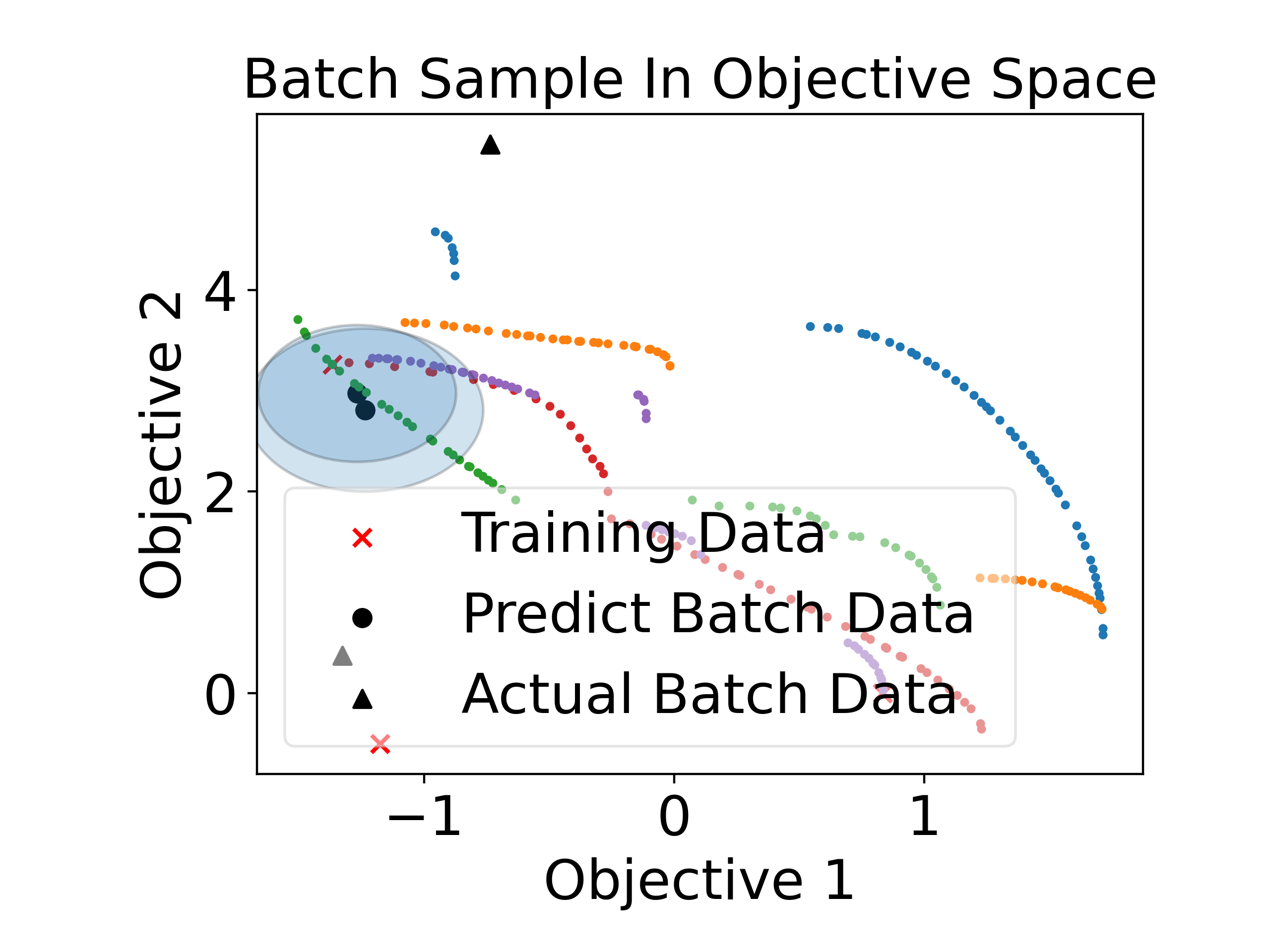} 
    \end{subfigure}
            \begin{subfigure}[t]{0.195\textwidth}
        \hspace{-5mm}
        \includegraphics[width=1.1\linewidth]{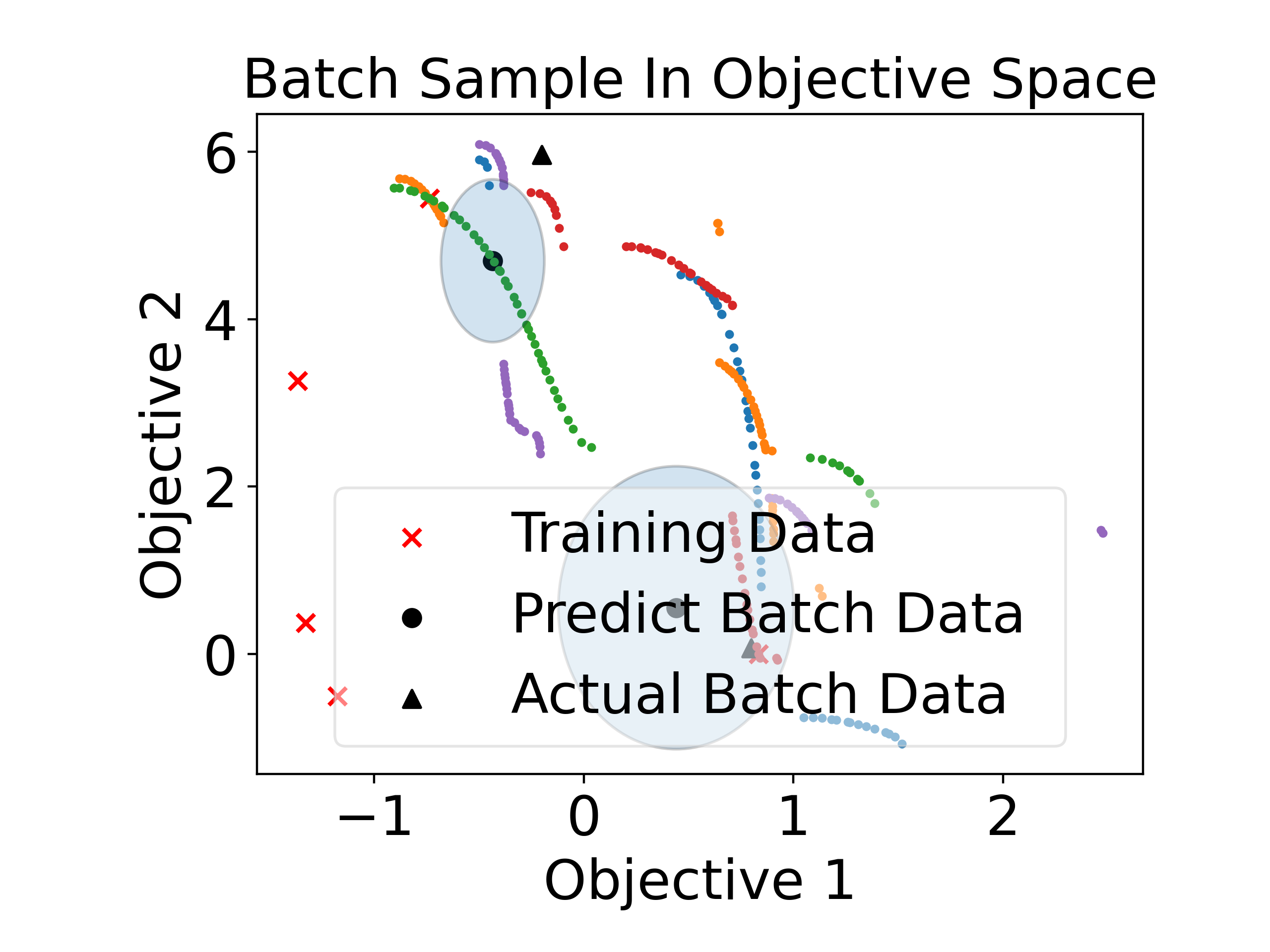} 
    \end{subfigure}
            \begin{subfigure}[t]{0.195\textwidth}
        \hspace{-5mm}
        \includegraphics[width=1.1\linewidth]{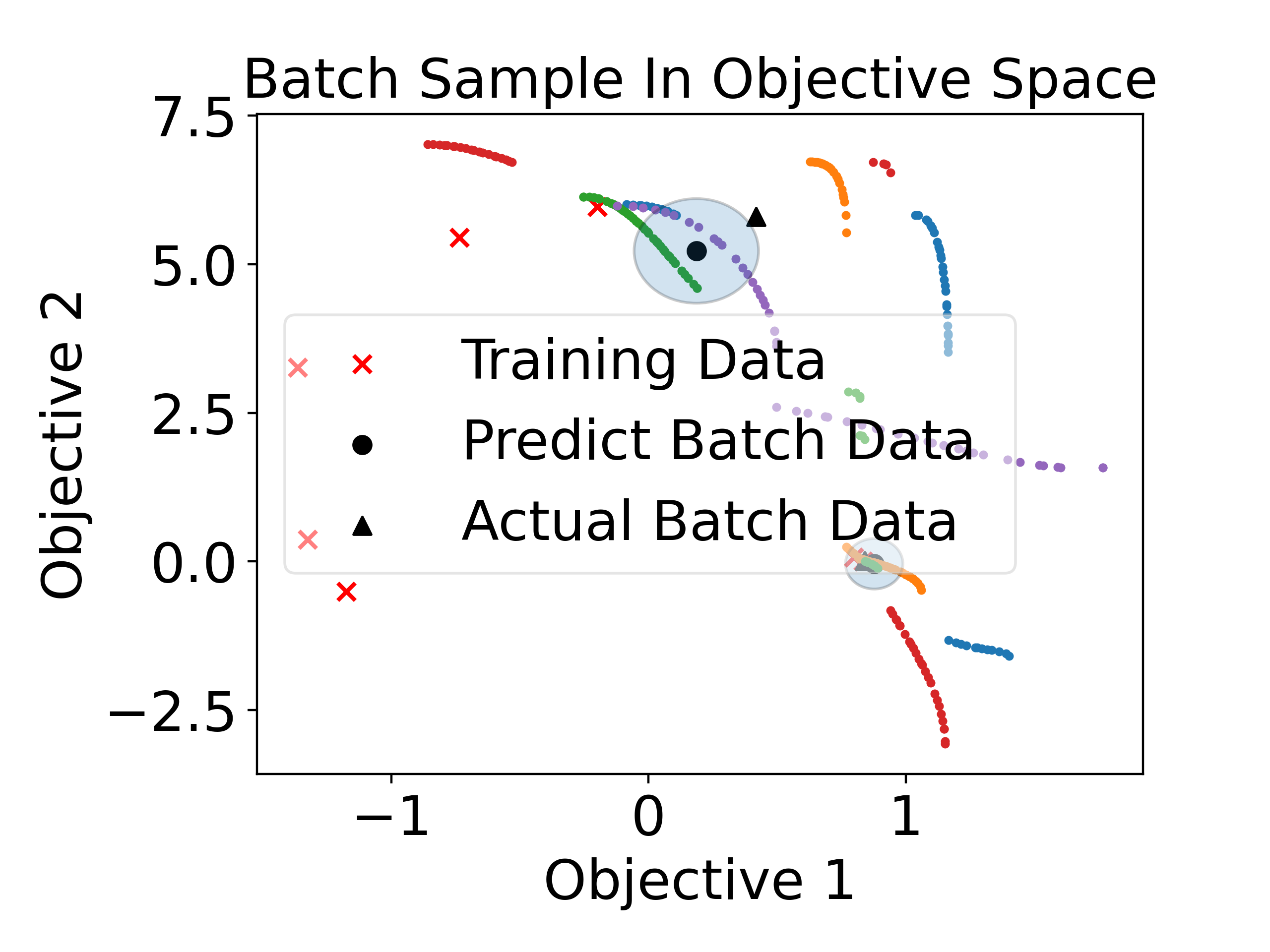} 
    \end{subfigure}
                \begin{subfigure}[t]{0.195\textwidth}
        \hspace{-5mm}
        \includegraphics[width=1.1\linewidth]{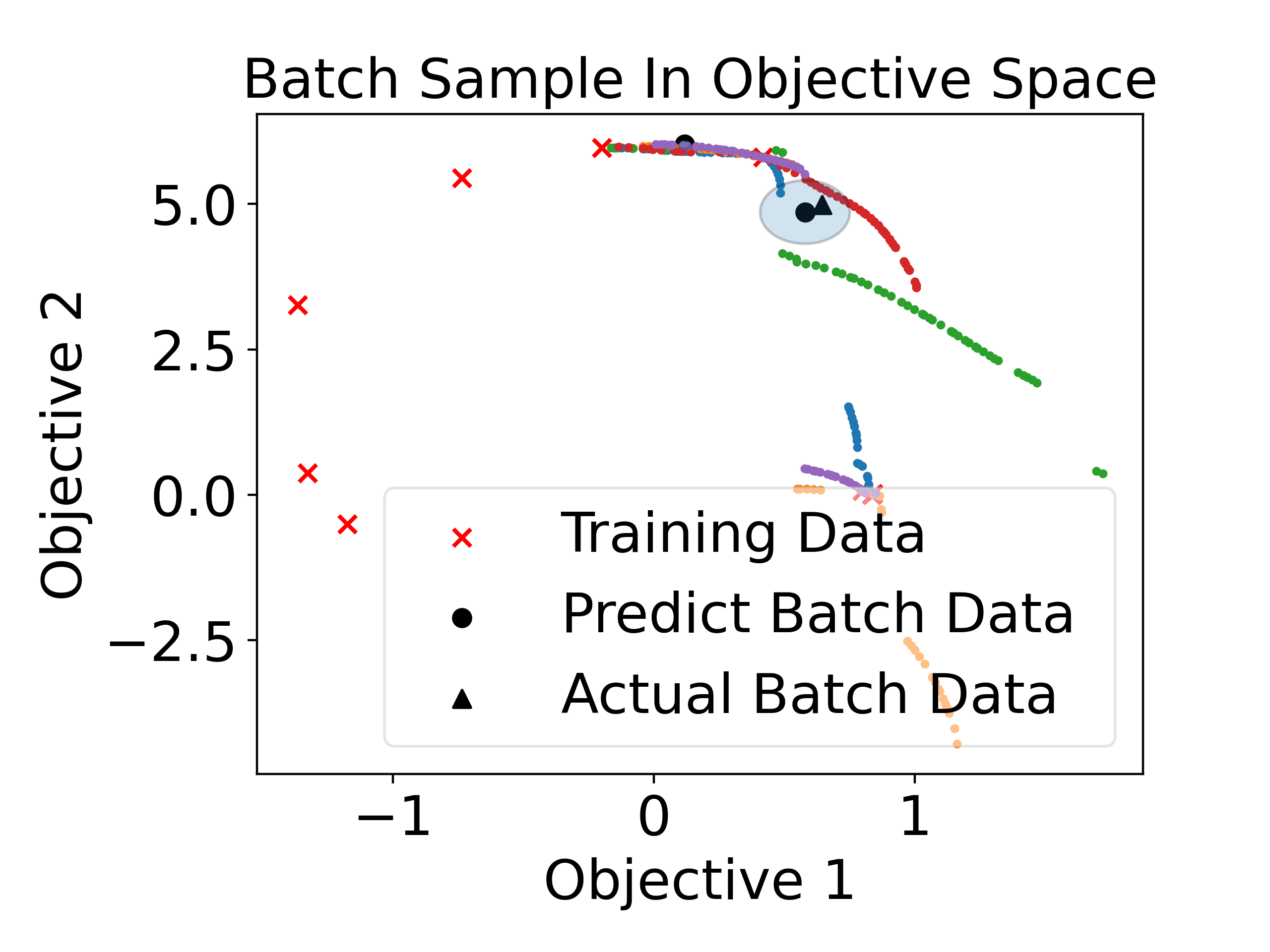} 
    \end{subfigure}
                    \begin{subfigure}[t]{0.195\textwidth}
        \hspace{-5mm}
        \includegraphics[width=1.1\linewidth]{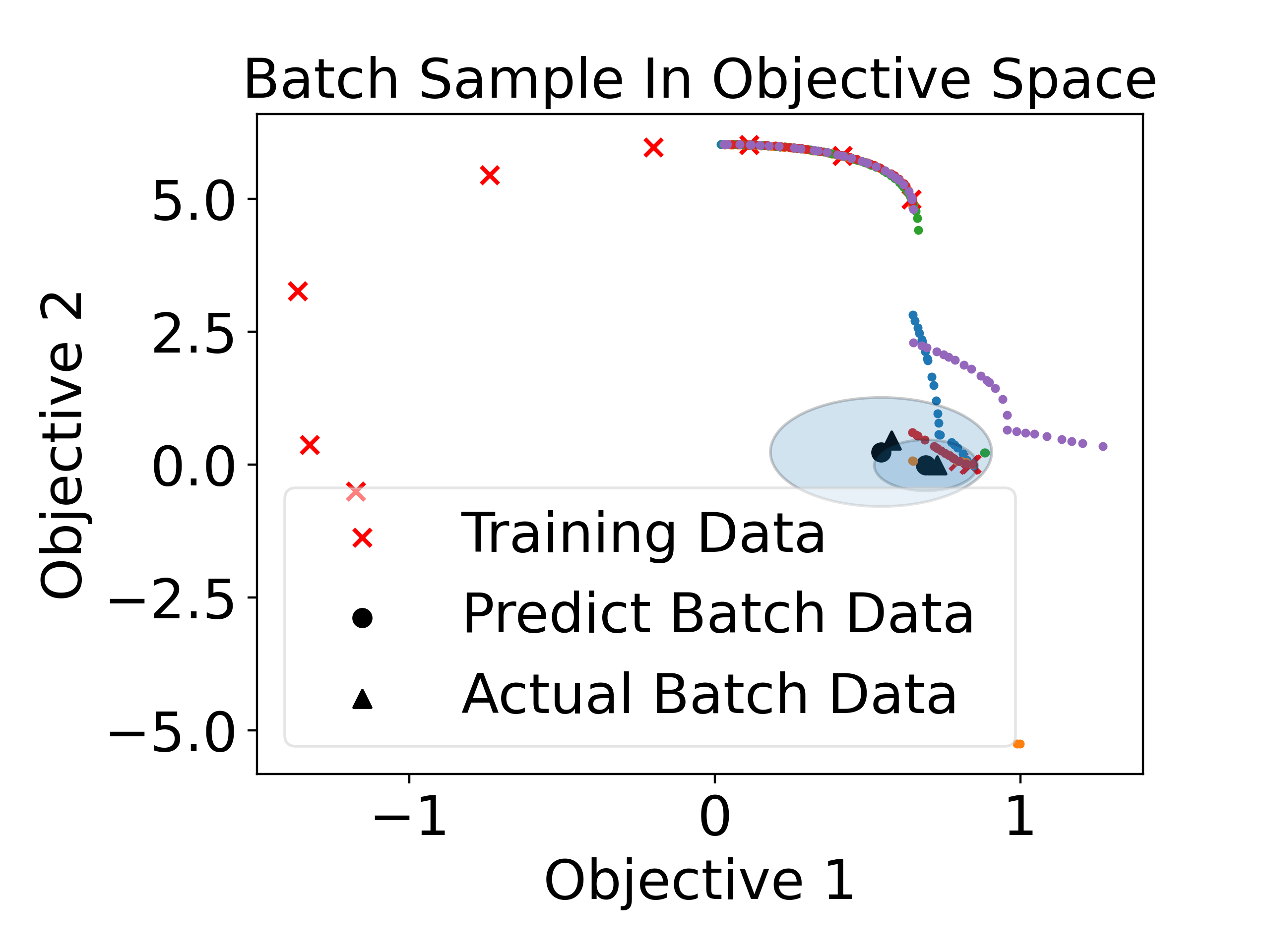} 
    \end{subfigure}
    \hfill
            \begin{subfigure}[t]{0.195\textwidth}
        \hspace{-5mm}
        \includegraphics[width=1.1\linewidth]{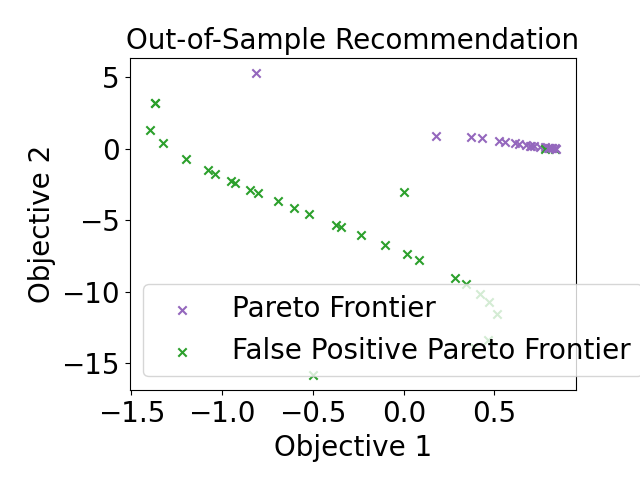} 
    \end{subfigure}
    \hfill
                \begin{subfigure}[t]{0.195\textwidth}
        \hspace{-5mm}
        \includegraphics[width=1.1\linewidth]{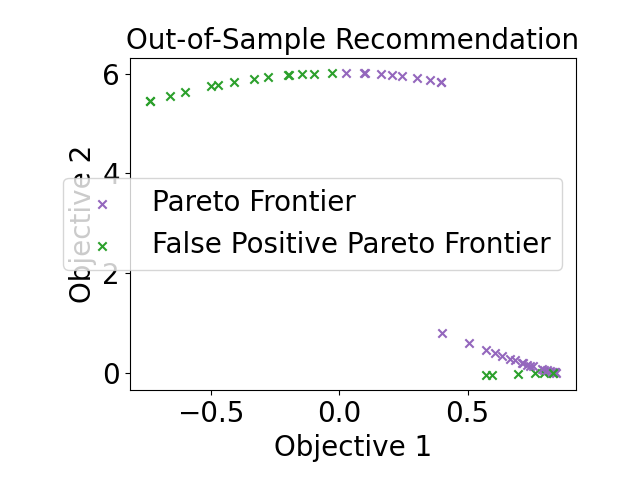} 
    \end{subfigure}
    \hfill
                \begin{subfigure}[t]{0.195\textwidth}
        \hspace{-5mm}
        \includegraphics[width=1.1\linewidth]{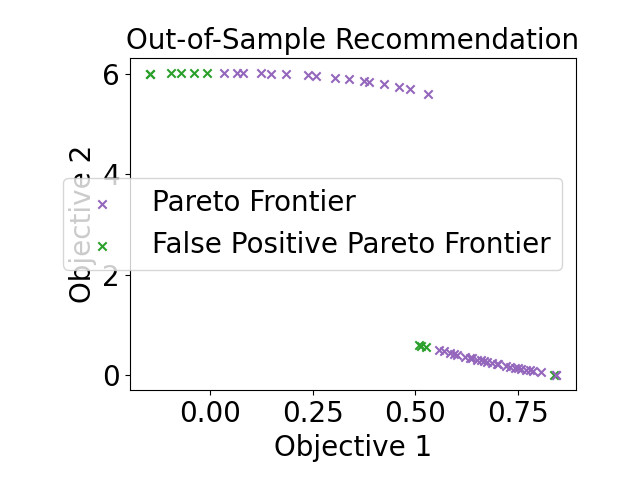} 
    \end{subfigure}
    \hfill
                    \begin{subfigure}[t]{0.195\textwidth}
        \hspace{-5mm}
        \includegraphics[width=1.1\linewidth]{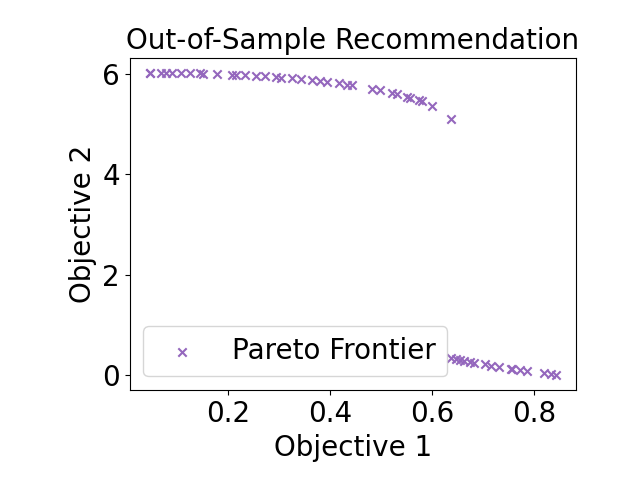} 
    \end{subfigure}
        \hfill
                    \begin{subfigure}[t]{0.195\textwidth}
        \hspace{-5mm}
        \includegraphics[width=1.1\linewidth]{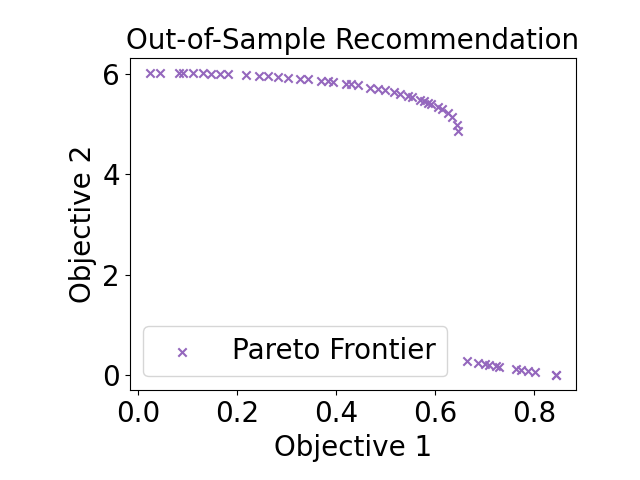} 
    \end{subfigure}
    \hfill
    \caption{q-\{PF\}$^2$\text{ES} on the inverted Sinlinear-Forrester problem ($d = 1, M=2, C=0$), i.e., for maximization we take the negative of the objective functions, with $q = 2$ starting from three training data points. The first row is contour of q-\{PF\}$^2$\text{ES} in the concatenated batch input space. The second row represents the sampled Pareto frontiers $\bm{\tilde{F}}$ and the queried batches in the objective space (the shaded oval area is the $2 \bm{\sigma}$ of the GP uncertainty). The third row shows the out-of-sample recommended Pareto frontiers after the batch query, with false positives denoted in green. }
    \label{fig: batch_query_demo}
\end{figure}

As a proof of concept, we illustrate the effectiveness of q-\{PF\}$^2$\text{ES} (with batch size $q = 2$, and qMC sample size 128) on a one-dimensional Sinlinear-Forrester function (Table~3 of \cite{qing2022robust}). The progress plot of batch queries is provided in Fig.~\ref{fig: batch_query_demo}. It can be seen that the out-of-sample strategy is able to accurately recommend a Pareto frontier without false positives after five batches BO iterations.

\section{ALTERNATIVE STRATEGY FOR SETTING $\varepsilon$}\label{app:alternative_epsilon}

In the case of common bi-objective optimization ($M=2$) scenarios, $\varepsilon$ can also be specified using an alternative strategy. The motivation for this approach is that, under mild additional assumptions, $\varepsilon$ is closer to the infimum of the \textit{mutual information lower bound set} (i.e., an ordered set of $\varepsilon$ with the lower bound property of the mutual information).

\begin{prop}
For a continuous Pareto Frontier $\mathcal{F}$ with $M = 2$, given its finite approximation $\tilde{\mathcal{F}}$:= $\{\bm{f}^* \vert  \bm{f}^* \in \mathcal{F}\}$ that satisfy $\forall k \in M$: $\exists \bm{f}^* \in \tilde{\mathcal{F}} s.t. \bm{f}_k^* = \underset{\bm{x} \in \mathcal{X}}{max}(\bm{f}_k(\bm{x}))$ and $\vert \tilde{\mathcal{F}}\vert \geq 2$, define its corresponding set for $k$th output $\tilde{\mathcal{F}^k}_{ord}$ ordered with $\leq$. Define $\bm{\varepsilon_t} := [\varepsilon_1, ..., \varepsilon_M]$, where $\forall k \in M, \varepsilon_k  = \underset{1 \leq i \leq \vert\tilde{\mathcal{F}}\vert - 1}{max}(\tilde{\mathcal{F}^k}_{{ord}_{i+1}} - \tilde{\mathcal{F}^k}_{{ord}_{i}})$. $\forall \bm{\varepsilon} \geq \bm{\varepsilon_t}$, with the hypervolume partition $\mathcal{P}(\tilde{\mathcal{F}_{\bm{\varepsilon}}})$ based on $\tilde{\mathcal{F}_{\bm{\varepsilon}}}:= \{ \bm{f} + \bm{\varepsilon} \vert \bm{f} \in \tilde{\mathcal{F}}\}$, $I(\mathcal{F}; \bm{h}_{\bm{x}} ) \geq \underset{\vert \tilde{\mathbf{F}}\vert \rightarrow \infty}{\textup{lim}} \tilde{\alpha}_{\{\text{PF}\}^2\text{ES}}(\tilde{A}(\tilde{\mathcal{F}}_{\bm{\varepsilon}}))$.
\label{prop: chosen_epsilon}
\end{prop}

Proposition~\ref{prop: chosen_epsilon} states that under the \textit{extrema assumption} (i.e., the discrete Pareto frontier preserves all extrema points in $\mathcal{F}$), specifying $\varepsilon$ as the maximum inner spacing between Pareto points within $\tilde{\mathcal{F}}$ per output dimension. One can subsequently define the non-dominated region $\tilde{A}$ based on the shifted Pareto frontier $\tilde{\mathcal{F}_{\bm{\varepsilon}}}$, which guarantees the acquisition function to be a lower bound of mutual information estimation (Eq.~8) when the Pareto frontier samples $\vert \tilde{\mathbf{F}}\vert \rightarrow \infty$. 
Here we elaborate on the proof of Proposition~\ref{prop: chosen_epsilon}. We start with assuming an unconstrained multi-objective optimization problem. Note that the following proof also utilizes the well-established definition of \textit{strict} and \textit{weak dominance} (i.e., $\succ$ and $\succeq$ respectively) and the property of transitivity of dominance (i.e., $a \succeq b \succeq c \Rightarrow a \succeq c $), for which we refer to standard multi-objective optimization textbooks, e.g., \citet{emmerich2018tutorial}. 

\begin{lemma} 
For any discrete Pareto frontier $\tilde{\mathcal{F}^*}$ when $M=2$, with $\bm{\varepsilon}_t$ defined as in proposition~\ref{prop: chosen_epsilon}, $\forall \bm{f} \in \tilde{A}(\tilde{\mathcal{F}^*}) \cap \{\bm{f}\vert \forall k \in M: \bm{f}_k \leq max(\tilde{\mathcal{F}_k^*})\} $ , $\exists \bm{f}^* \in \tilde{\mathcal{F}^*}$ s.t. $\bm{f}^* \preceq \bm{f} + \bm{\varepsilon}_t$ and $\bm{f}^*$ is not unique.
\end{lemma} 

\begin{proof}
For any $k$th output, given that $\bm{f}_k \leq max(\tilde{\mathcal{F}_k^*})$, we know that $\exists k_i \in [1, \vert\tilde{\mathcal{F}}\vert - 1] $: $\tilde{\mathcal{F}}^k_{{ord}_{k_i}} \leq \bm{f}_k \leq \tilde{\mathcal{F}^k}_{{ord}_{k_{i+1}}}$. $1^\circ$: if for $l  \neq k$, $\tilde{\mathcal{F}}^l_{{ord}_{l_{\vert \tilde{F} \vert - i}}} \leq \bm{f}_l \leq \tilde{\mathcal{F}^l}_{{ord}_{l_{\vert \tilde{F} \vert - i+1}}}$, then we know that $\bm{f}_k + \varepsilon_k \geq \tilde{\mathcal{F}^k}_{{ord}_{k_{i+1}}}$ and $\bm{f}_l + \varepsilon_l \geq \tilde{\mathcal{F}^l}_{{ord}_{l_{\vert \tilde{F} \vert - i+1}}}$, hence $\bm{f} + \bm{\varepsilon} \succeq (\tilde{\mathcal{F}^k}_{{ord}_{k_{i+1}}}, \tilde{\mathcal{F}^l}_{{ord}_{l_{\vert \tilde{F} \vert - i+1}}})$. Since $(\tilde{\mathcal{F}^k}_{{ord}_{k_{i+1}}}, \tilde{\mathcal{F}^l}_{{ord}_{l_{\vert \tilde{F} \vert - i+1}}}) \succeq (\tilde{\mathcal{F}^k}_{{ord}_{k_{i}}}, \tilde{\mathcal{F}^l}_{{ord}_{l_{\vert \tilde{F} \vert - i+1}}})$ and  $(\tilde{\mathcal{F}^k}_{{ord}_{k_{i+1}}}, \tilde{\mathcal{F}^l}_{{ord}_{l_{\vert \tilde{F} \vert - i+1}}}) \succeq (\tilde{\mathcal{F}^k}_{{ord}_{k_{i+1}}}, \tilde{\mathcal{F}^l}_{{ord}_{l_{\vert \tilde{F} \vert - i}}})$, the lemma holds; $2^\circ$: if for $l  \neq k$, $ \bm{f}_l > \tilde{\mathcal{F}^l}_{{ord}_{l_{\vert \tilde{F} \vert - i+1}}}$. Finally it can be shown that $(\tilde{\mathcal{F}^k}_{{ord}_{k_{i+1}}}, \tilde{\mathcal{F}^l}_{{ord}_{l_{\vert \tilde{F} \vert - i+1}}}) \succeq (\tilde{\mathcal{F}^k}_{{ord}_{k_{i}}}, \tilde{\mathcal{F}^l}_{{ord}_{l_{\vert \tilde{F} \vert - i+1}}})$ and  $(\tilde{\mathcal{F}^k}_{{ord}_{k_{i+1}}}, \tilde{\mathcal{F}^l}_{{ord}_{l_{\vert \tilde{F} \vert - i+1}}}) \succeq (\tilde{\mathcal{F}^k}_{{ord}_{k_{i+1}}}, \tilde{\mathcal{F}^l}_{{ord}_{l_{\vert \tilde{F} \vert - i}}})$.
\end{proof}

\begin{lemma}
For a given discrete Pareto frontier $\tilde{\mathcal{F}}$ with $\bm{\varepsilon}_t$ as defined in proposition~\ref{prop: chosen_epsilon}, $\tilde{A}(\tilde{\mathcal{F}_{{\varepsilon}_t}}) \subseteq A(\mathcal{F})$.
\label{lm:lemma2}
\end{lemma}

\begin{proof}
Note the necessary condition for $\tilde{A}(\tilde{\mathcal{F}}_{\bm{\varepsilon}_t}) \subseteq A(\mathcal{F})$ is $\forall \bm{f} \in \tilde{A}(\tilde{\mathcal{F}}_{\bm{\varepsilon}_t})$: $\bm{f} \in A(\mathcal{F})$. Proof by contradiction: suppose $\exists \bm{f} \in \tilde{A}(\tilde{\mathcal{F}}_{\bm{\varepsilon}_t})$ s.t. $\bm{f} \notin A(\mathcal{F})$, hence $\bm{f} \in \overline{A}(\mathcal{F})$. This implies $  \exists \bm{f}' \in \mathcal{F}: \bm{f}' \succeq  \bm{f} \Rightarrow \bm{f}'+\bm{\varepsilon}_t \succeq  \bm{f} +\bm{\varepsilon}_t$. $1^\circ$: Suppose $\forall k \in M$: $\bm{f}_k \leq max(\tilde{\mathcal{F}}_{{\bm{\varepsilon}_t}_k})$, with lemma~1, by setting $\tilde{\mathcal{F}^*} = \tilde{\mathcal{F}}_{\bm{\varepsilon}_t}$ we know $\exists \bm{f}'' \in \tilde{\mathcal{F}}_{\bm{\varepsilon}_t}$ s.t.  $\bm{f}''  \preceq \bm{f} +\bm{\varepsilon}_t $, then $\bm{f}'+\bm{\varepsilon}_t \succeq  \bm{f} +\bm{\varepsilon}_t \succeq \bm{f}'' 	\Rightarrow \bm{f}' \succeq \bm{f}'' - \bm{\varepsilon}_t$, since  $\bm{f}'' - \bm{\varepsilon}_t \in \mathcal{F}$, due to the definition of a Pareto frontier this can only be hold when $\bm{f}'' - \bm{\varepsilon}_t = \bm{f}'$, which contradicts with its non-unique property, hence lemma holds; $2^\circ$: Suppose $\exists k \in  M\ s.t.\  \bm{f}_k > max(\tilde{\mathcal{F}}_{{\bm{\varepsilon}_t}_k})$, since according to the definition of $\tilde{\mathcal{F}}$, $max(\tilde{\mathcal{F}}_{{\bm{\varepsilon}_t}_k}) = max(\mathcal{F}_{k}) + {\bm{\varepsilon}_t}_k$, this results $\bm{f}_k > max(\mathcal{F}_{k})$ and hence $\bm{f} \in A(\mathcal{F})$, contradicting with $\bm{f} \in \overline{A}(\mathcal{F})$.



\end{proof}


Eventually, the proof of Proposition~\ref{prop: chosen_epsilon} becomes straightforward: with lemma~\ref{lm:lemma2} and Eq.~6 we have $I(\mathcal{F}; \bm{h}_{\bm{x}} ) \geq  - \mathbb{E}_{  \mathcal{F}}\left[\text{log}\left(1 -Z_{A(\mathcal{F})} \right)\right]  = \underset{\vert \mathbf{F}\vert \rightarrow \infty}{\text{lim}} - \frac{1}{ \vert \mathbf{F}\vert }\sum_{\mathcal{F} \in \mathbf{F}} \left[\text{log}\left(1 -Z_{A(\mathcal{F})} \right)\right]  \geq \underset{\vert \tilde{\mathbf{F}}\vert \rightarrow \infty}{\text{lim}} \tilde{\alpha}_{\{\text{PF}\}^2\text{ES}}(\tilde{A}(\tilde{\mathcal{F}}_{\bm{\varepsilon}}))$ since $Z_{A(\mathcal{F})} \geq Z_{\tilde{A}(\tilde{\mathcal{F}}_{\bm{\varepsilon}})}$.

Remark that the proposition~\ref{prop: chosen_epsilon} is also valid for CMOO when $\mathcal{F} \neq \phi$ (the extrema condition in CMOO needs to be generalized for constraints). In reality, $\mathcal{F}$ is not continuous for all problems, unfortunately. Clustering techniques (e.g., \cite{schubert2017dbscan}) can be utilized as a pre-processing step to satisfy the condition of the proposition. Nevertheless, this does increase the complexity of the method. For $M>2$, proposition~\ref{prop: chosen_epsilon} does not hold since $\bm{\varepsilon}$, as defined using the maximum inner spacing, can not guarantee that $\tilde{A}(\tilde{\mathcal{F}_{\varepsilon}}) \subseteq A(\mathcal{F})$, and the choice of $\bm{\varepsilon}$ can be correlated with the choice of partitioning strategy $\mathcal{P}$.  


The practical implementation of the original acquisition function \{PF\}$^2$ES-c, as well as using the proper $\bm{\varepsilon}$ approach based on proposition~\ref{prop: chosen_epsilon} (\{PF\}$^2$ES-lb) is detailed in Algorithm \ref{alg: original} and \ref{alg: prop1}, respectively. Note that lines 2-9 in both algorithms only need to be calculated once per BO iteration.



\subsection{Comparing strategies for setting $\varepsilon$}
\begin{figure}[h]
  \centering
  \includegraphics[width=1.0\linewidth]{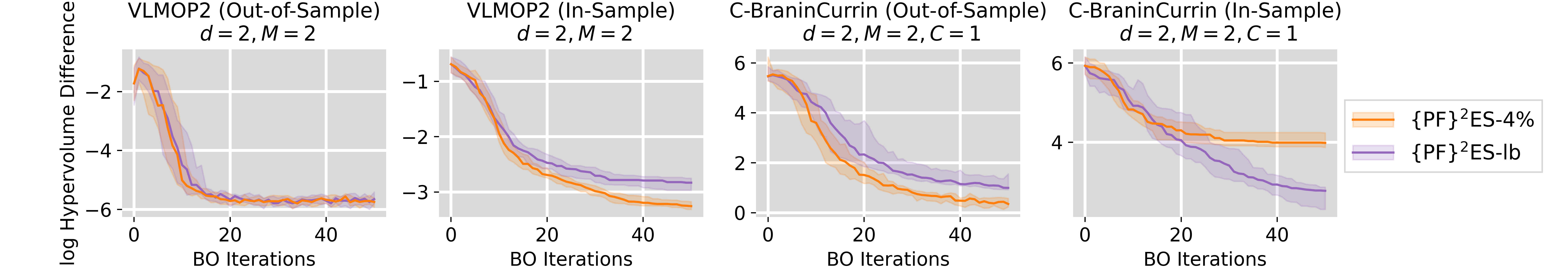}  
\caption{Comparing strategies for setting $\bm{\varepsilon}$ for two synthetic benchmark functions.}
\label{fig:epsilon_approach_comparison}
\end{figure}

We investigate the two different approaches for setting $\bm{\varepsilon}$ for two bi-objective optimization problems. Namely, VLMOP2 and C-BraninCurrin as provided in Table~\ref{Tabular: reference_pts}, both with continuous Pareto frontiers. The results are shown in Fig.~\ref{fig:epsilon_approach_comparison} where we refer to the heuristical approach as \{PF\}$^2$ES-4\% (i.e.,$c = 0.04$) and the approach above as \{PF\}$^2$ES-lb. From the in-sample regret, it can be seen that \{PF\}$^2$ES-lb can be more greedy than \{PF\}$^2$ES-4\%, although it is less competitive in terms of out-of-sample recommendations for C-BraninCurrin. As conclusion, both strategies lead to comparable performance. Hence, we propose to simply utilize the heuristic approach which does not depend on extra assumptions and generalizes better with respect to $M$.

\begin{minipage}[t]{0.46\textwidth}
\begin{algorithm}[H]
    \centering
    \caption{Practical implementation of \{PF\}$^2$ES-c}\label{algorithm}
    \begin{algorithmic}[1]
        \State \textbf{Input}: a candidate $\bm{x}$; GP models for objectives (and constraints): $\mathcal{M}$, heuristic shifting percentage $c$.
        \For{$i= 1,...,\vert \mathcal{F} \vert $}
        \State generate $i$th GP parametric trajectory approximations (see e.g., Eq.~3, 4 of \cite{qing2022spectral}).  
        
        \State Apply NSGA2 to GP trajectories to obtain Pareto frontier samples $\tilde{\mathcal{F}}$
        \For {$k= 1,...,M $}:
        \State $\epsilon_k =  c \cdot \left(max(\tilde{\mathcal{F}^k}) - min(\tilde{\mathcal{F}^k})\right)$
        \EndFor
        \State $\tilde{\mathcal{F}}_{\varepsilon} = \tilde{\mathcal{F}} + \varepsilon$
        \State Partition $\tilde{A}: \tilde{A} = \mathcal{P}(\tilde{\mathcal{F}}_{\varepsilon}) $
        \EndFor
        \State \textbf{Return}  \{PF\}$^2$ES-c (Eq.~10).
    \end{algorithmic}
\label{alg: original}
\end{algorithm}
\end{minipage}
\hfill
\begin{minipage}[t]{0.46\textwidth}
\begin{algorithm}[H]
    \centering
    \caption{Practical implementation of \{PF\}$^2$ES-lb}\label{algorithm}
    \begin{algorithmic}[1]
        \State \textbf{Input}: a candidate $\bm{x}$; GP models for objectives (and constraints): $\mathcal{M}$.
        \For{$i= 1,...,\vert \mathcal{F} \vert $}
        \State generate $i$th GP parametric trajectory approximations (see, e.g., Eq.~3, 4 of \cite{qing2022spectral}).  
        
        \State Apply NSGA2 to GP trajectories to obtain Pareto frontier samples $\tilde{\mathcal{F}}$
        \For {$k= 1,...,M $}:
        \State $\varepsilon_k  = \underset{1 \leq i \leq \vert\tilde{\mathcal{F}}\vert - 1}{max}(\tilde{\mathcal{F}^k}_{{ord}_{i+1}} - \tilde{\mathcal{F}^k}_{{ord}_{i}})$ as in proposition~\ref{prop: chosen_epsilon}
        \EndFor
        \State $\tilde{\mathcal{F}}_{\varepsilon} = \tilde{\mathcal{F}} + \bm{\varepsilon}$
        \State Partition $\tilde{A}: \tilde{A} = \mathcal{P}(\tilde{\mathcal{F}}_{\varepsilon}) $
        \EndFor
        \State \textbf{Return}  \{PF\}$^2$ES-lb (Eq.~10).
    \end{algorithmic}
\label{alg: prop1}
\end{algorithm}
\end{minipage}

\section{EXPERIMENTAL DETAILS} \label{app: exp_detail}  

\subsection{Synthetic problems and setting the reference point} \label{app:synthetic_probs_and_ref_pts}
We benchmark on synthetic problems regularly used within the literature. Details are given in Table~\ref{Tabular: reference_pts}, together with the reference point used in the uncertainty calibration \footnote{Note that since the Pareto frontier samples from the GP posterior, in case of an inaccurate model, it can significantly deviate from the real Pareto frontier (e.g., at the initial stages of the BO process). Hence, the reference point utilized in the hypervolume indicator calculation is set more conservatively than in the regret calculation.} (i.e., Fig.~3) and for performance metric (i.e., Fig.~4, 5). Note that the ideal hypervolume utilized to calculate the log hypervolume difference performance metric is set optimistically to guarantee a well-behaved performance metric.


\begin{sidewaystable}
\caption{Details of the synthetic problems and reference point settings}
\label{sample-table}
\begin{center}
\begin{tabular}{llllllll}
\toprule
Problem  &  Formulation &$d$ &  $M$ &  $C$& Reference point & Reference point & Ideal   \\
 &  & & & & (uncertainty  & (regret) & Hypervolume \\
  & & & & & calibration) && (regret) \\
\midrule
VLMOP2 & Table~1 of \cite{van1999multiobjective} & 2& 2& 0 &  $[3, 3]$ & $[1.2, 1.2]$& 0.77157 \\
BraninCurrin & Appendix E.2 of \citep{daulton2020differentiable} & 2& 2& 0&  $[2000, 50]$ & $[18.0, 6.0]$ &  60.0000\\
ZDT1 & Eq.~7 of \cite{zitzler2000comparison} & 5 & 2& 0&  $[15, 15]$ & $[2.5, 2.5]$& 5.90453\\ 
ZDT2 & Eq.~8 of \cite{zitzler2000comparison} &5 & 2& 0&   $[15, 15]$  & $[2.5, 2.5]$ & 5.57236\\ 
Four Bar Truss & Eq.~21 of \cite{dauert1993multicriteria}& 4 & 2 & 0 & NA & $[3400, 0.05]$ & 81.9590 \\
DTLZ4 (App~\ref{sec: 9.2}) & Eq.~24 of \cite{deb2005scalable}& 8 & 2 & 0 & NA & $[2.5, 2.5]$ & $5.45327$ \\
DTLZ4 (App~\ref{sec: 9.3}) & Eq.~24 of \cite{deb2005scalable}& 5 & 3 & 0 & NA & $[2.5, 2.5, 2.5]$ & $15.0080$ \\
\midrule
C-BraninCurrin & Appendix E.2 of \cite{daulton2020differentiable} & 2& 2& 1&  $[2000, 100]$ & $[80.0, 12.0]$ & 606.752\\
Constr-EX & Eq.~7.2 of \cite{constr_ex_prob} & 2& 2& 2 & $[2, 50]$&  $[1.1, 10.0]$& 5.30186\\
SRN & Eq.~10 of \cite{SRN_prob} & 2 & 2& 2& $[1500, 500]$ &$[250, 50]$ & 46205.0 \\ 
C2-DTLZ2 & Section 4.6 of \cite{deb2019constrained} & 4 & 2& 1& $[10, 250]$& $[2.5, 2.5]$  & 5.43594\\
Disc Brake Design & Section 3.4 of \cite{ray2002swarm} & 4 & 2 & 5 & NA & $[8.0, 4.0]$ & 17.6882 \\
OSY (App~\ref{sec: 9.2}) & Sec~4.3 of \cite{osyczka1995new}& 6 & 2 & 6 & NA & $[50, 100]$ & $25344.9$ \\
Marine Design (App~\ref{sec: 9.3}) & \href{https://github.com/ryojitanabe/reproblems/blob/28845742cc72910e301d5c9d1d806ef54185c074/reproblem_python_ver/reproblem.py#L1109}{CRE32} of \cite{tanabe2020easy} & 6 & 3 & 9& NA& $[-700, 13000, 3500]$ & $6991669$ \\
\bottomrule
\end{tabular}
\end{center}
\label{Tabular: reference_pts}
\end{sidewaystable}
  
\subsection{Out-of-sample recommendation strategy} \label{app:out_of_sample}
For both MOO and CMOO problems, the feasible Pareto optimal inputs $\bm{X}_r$ is recommended using a conservative model-based approach similar to \cite{garrido2020parallel, ungredda2021bayesian}. The recommendation task can be formulated as a MOO problem:

\begin{equation}
\begin{aligned}
    &\underset{\bm{x} \in \mathcal{X}}{maximize}\  \bm{m}^1, \bm{m}^2, ..., \bm{m}^M \\ 
    &s.t., 1 - \Phi(\frac{\bm{m}^{M+1} -  \eta^1}{\bm{\sigma}^{M+1}}) \geq C_{Fea},..., 1 - \Phi(\frac{\bm{m}^{M+C} -  \eta^C}{\bm{\sigma}^{M+C}}) \geq C_{Fea} 
\end{aligned}
\end{equation}

\noindent where $m$ and $\sigma$ represent the GP poster mean and standard deviation, respectively. We expect all the recommendations to be feasible in our empirical experiments, hence $C_{Fea}$ is set to 0.95 and is decreased with 0.05 in case no feasible solution exists (i.e., $\bm{X}_r = \phi$), $\eta^C = 0.5\% \cdot \left(max(\bm{H}^C) - min(\bm{H}^C)\right)$. In practice, we empirically find that while an aggressive recommendation strategy (e.g., $C_{Fea}=0.5, \eta = 0$) can possibly contain infeasible candidates, the recommended candidates are able to cover the real Pareto frontier faster and the constraints are often only violated with a tiny amount (e.g., $1e-3$ level), which is tolerable for some applications. Hence, the practitioner is recommended to choose $C_{Fea}$ and $\eta$ according to the tolerance to constraint violations. 

\subsection{Setup details of different acquisition functions}\label{app:acq_setups}
We provide configuration details for different information-theoretic acquisition functions used in the comparison. 

\textbf{EHVI} is using the Trieste \citep{Berkeley_Trieste_2022} implementation. The dynamic reference point specification strategy \citep{knudde2017gpflowopt} is utilized:
\begin{equation}
r_{ref} =   max(\tilde{\mathcal{F}}) + 2 \cdot \frac{max(\tilde{\mathcal{F}}) - min(\tilde{\mathcal{F}})}{\vert \tilde{\mathcal{F}} \vert }
\end{equation}

\textbf{qEHVI} is based on the Trieste \citep{Berkeley_Trieste_2022} implementation and extended with a random quasi-Monte Carlo (qMC) method for batch reparameterization sampling(see Eq.~15 of \cite{daulton2020differentiable})). 128 qMC samples are used to approximate the hypervolume improvement. 

The qEHVI paper does not provide a dynamic reference point strategy for CMOO. Denoting the outcome of objective function on training data as $\bm{F}$, the following strategy is used:
\begin{equation}
r_{ref} =      \left\{
\begin{array}{ll}
       max(\tilde{\mathcal{F}}) + 2 \cdot \frac{max(\tilde{\mathcal{F}}) - min(\tilde{\mathcal{F}})}{\vert \tilde{\mathcal{F}} \vert } & \tilde{\mathcal{F}} \neq  \phi  \\
      max(\bm{F}) + 2 \cdot \frac{max(\bm{F}) - min(\bm{F})}{\vert \bm{F} \vert } & \tilde{\mathcal{F}}  = \phi   \\
\end{array} 
\right. 
\end{equation}

\textbf{EHVI-PoF} is a common strategy for handling CMOO. In case there are no predicted feasible observations on the training data, the probability of feasibility function will be used for locating a feasible observation first. For EHVI-PoF, the same reference point setting as EHVI is utilized, where $\tilde{\mathcal{F}}$ is the \textbf{feasible} Pareto frontier. 

\textbf{Information-theoretic acquisition functions}
For PESMO, MESMOC+ and PPESMOC, the implementations of the accompanying papers are used \footnote{We are not able to use a continuous optimizer for out-of-sample recommendations for MESMOC+ and PPESMOC.} and implement the remaining acquisition functions ourselves. While we aim to use similar settings for all algorithms (e.g., the Pareto frontier number, the population sizes for NSGA2), the provided implementation of PESMO, MESMOC+, and PPESMOC still have different settings, see Table~\ref{tab: acq_configuration_difference}\footnote{NSGA2* is a parameter-less NSGA2 approach \citep{deb1999niched} to perform CMOO. We use the available training data to initialize NSGA2, which leads to improved performance.}. 


\begin{table*}[h]
\caption{Configurations of different information-theoretic acquisition functions.}
\begin{center}
\begin{tabular}{ l|l|l|l } 
\toprule
 & \begin{tabular}{cc} Pareto Frontier \\ Sample Strategy \end{tabular} &\begin{tabular}{cc} Pareto Frontier \\ Samples \end{tabular}&  \begin{tabular}{cc} Out-of-sample \\  Recommendation Strategy  \end{tabular}  \\
 \midrule
 \{PF\}$^2$ES (for MOO)& NSGA2 &5 & NSGA2\\
 PFES &NSGA2 &5  & NSGA2\\
 MESMO &NSGA2& 5 & NSGA2\\
 PESMO &Monte Carlo &1  & NSGA2\\ 
   q-\{PF\}$^2$ES (for MOO) &NSGA2 &5  & NSGA2\\
   \midrule
    \{PF\}$^2$ES (for CMOO)& NSGA2*  &5 & NSGA2*\\
 MESMOC &NSGA2* & 5 & NSGA2*\\
 MESMOC+  &Monte Carlo & 1 & Monte Carlo \\  
    q-\{PF\}$^2$ES (for CMOO) &NSGA2* &5  & NSGA2*\\
PPESMOC &Monte Carlo  & 1 & Monte Carlo \\ 
\bottomrule
\end{tabular}
\label{tab: acq_configuration_difference}
\end{center}
\end{table*}



\section{SENSITIVITY ANALYSIS OF THE HYPERPARAMETERS OF \{PF\}$^2$ES AND q-\{PF\}$^2$ES} \label{app:sensitivity_analysis}
\subsection{Parameter $\varepsilon$}
\begin{figure}[h]
  \centering
  \includegraphics[width=0.8\linewidth]{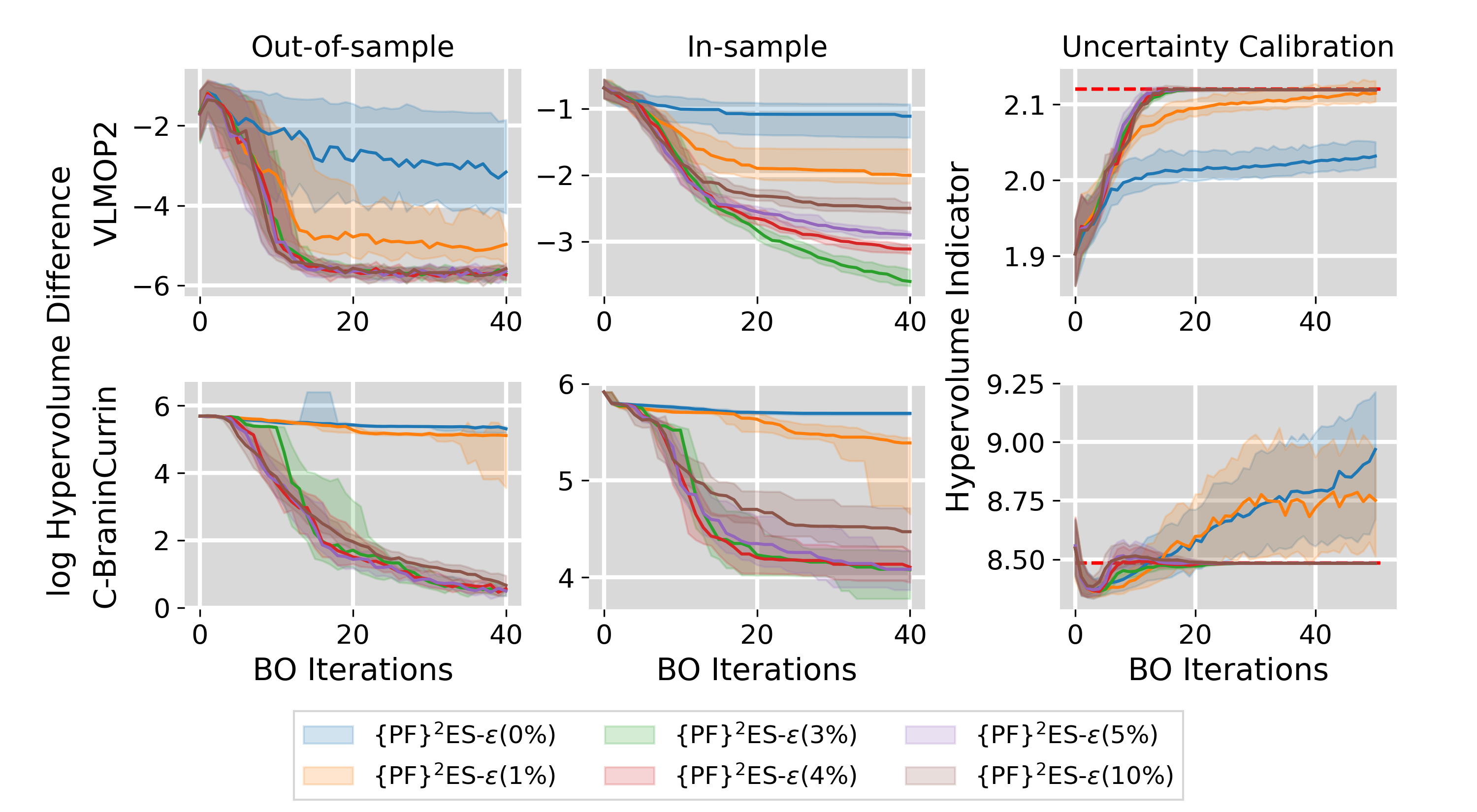}  
\caption{Sensitivity analysis of $\varepsilon$ on the performance of \{PF\}$^2$ES.}
\label{fig:sensitivity_epsilon}
\end{figure}

We conduct a sensitivity analysis for setting $\bm{\varepsilon}$ to determine its effect on the performance of the acquisition function. The results are depicted in Fig.~\ref{fig:sensitivity_epsilon}, with results of both in-sample, and out-of-sample regret, as well as the uncertainty calibration of the Pareto frontier.

 It can be seen that $\bm{\varepsilon}$ is less sensitive in terms of out-of-sample regret within the range of $3\% -5\%$. A too-small value will not mitigate the clustering issue, while a large value of $\varepsilon$ can cause too much exploration resulting in a deterioration of the in-sample regret. Importantly, using $\bm{\varepsilon}$ is helpful for a faster reduction of the uncertainty of the Pareto frontier.

\subsection{Number of Monte Carlo samples for q-\{PF\}$^2$ES}

\begin{figure}[h]
  \centering
  \includegraphics[width=1.0\linewidth]{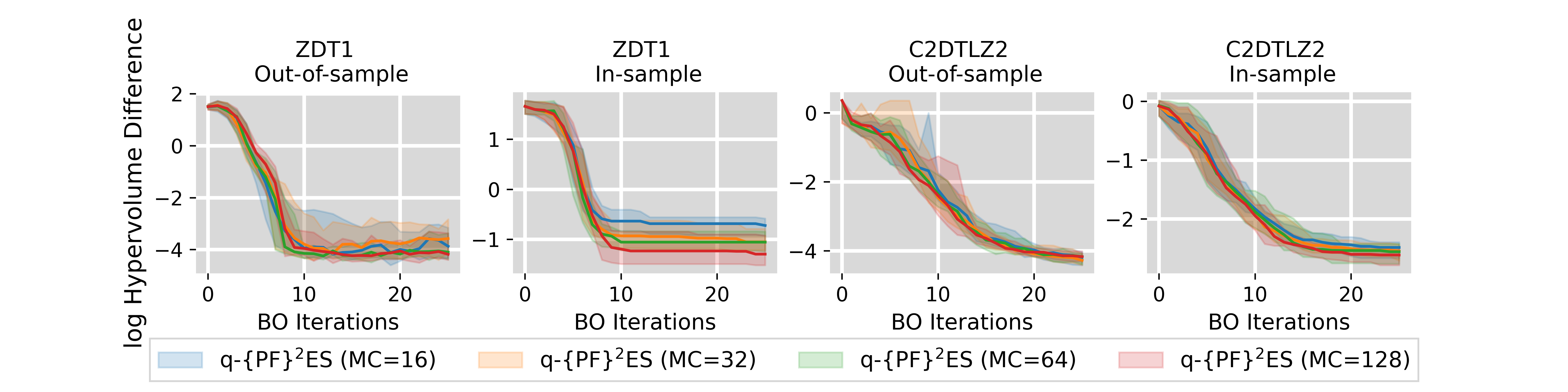}  
\caption{The performance difference of q-\{PF\}$^2$ES with respect to different MC sample sizes.}
\label{fig:mc_sensitivity_comparison}
\end{figure}

We conduct an empirical sensitivity analysis of the MC sample size used in approximating the q-\{PF\}$^2$ES (i.e., Eq.~12) and its effect on the performance. The results on two different benchmark functions are provided in Fig.~\ref{fig:mc_sensitivity_comparison}. As expected, it can be seen that with an out-of-sample recommendation strategy, a larger MC sample size can positively affect the performance of the acquisition function. The benefit is more obvious for the in-sample recommendation strategy.

\section{COMPUTATIONAL COMPLEXITY OF \{PF\}$^2$ES AND q-\{PF\}$^2$ ES} \label{sec:computation_complexity}
For the information-theoretic acquisition function, generating the Pareto frontier approximation (i.e., line 3-9 of Algorithm~\ref{alg: original},\ref{alg: prop1}) is treated separately as the \textit{initialization cost}, while the cost of evaluating the acquisition functions is termed as \textit{evaluation cost}.

\textbf{Initialization cost}  The initialization of \{PF\}$^2$ES consists of two parts: generation of $\tilde{\mathcal{F}}$ samples and calculate $\tilde{A}(\tilde{\mathcal{F}})$ by utilizing a hypervolume decomposition $\mathcal{P}$ on $\tilde{\mathcal{F}}$. The cost of the former, ignoring the non-dominated sorting complexity of NSGA2 which is $\mathcal{O}(M\vert\tilde{\mathcal{F}}\vert ^2)$ \citep{deb2002fast}, breaks down to the cost of evaluating the GP spectral samples, which is linear with the number of random Feature features.  The hypervolume decomposition strategy \citep{lacour2017box} utilized in this research has a complexity of $\mathcal{O}(\vert\tilde{\mathcal{F}}\vert^{\lfloor \frac{M}{2}\rfloor + 1})$. However, both \textbf{\{PF\}$^2$ES} and \textbf{q-\{PF\}$^2$ES} are agnostic to the choice of $\mathcal{P}$. 

\textbf{Evaluation cost}  

- \textbf{\{PF\}$^2$ES} By denoting the maximum number of cells across the set of Pareto frontier samples $\tilde{\bm{F}}$ as $max(N_p)$, the computational complexity of evaluate \{PF\}$^2$ES is: $\mathcal{O}\left(max(N_p)M\right)$ \footnote{Note that for both \{PF\}$^2$ES and q-\{PF\}$^2$ES, the complexity involved with constraint number $C$ is independent with the decomposed grid size $max(N_p)$ given independent assumption across each outcome.}.

- \textbf{q-\{PF\}$^2$ES}  We omit the cost of the batch reparameterization sampling (which is $\mathcal{O}((M+C)q)$) and hence, the evaluation cost of q-\{PF\}$^2$ES itself is:  $\mathcal{O}\left(max(N_p)qN_{MC}M\right)$.

\section{COMPARING RUNNING TIMES}\label{app:runtime}
\begin{figure}[!h]
  \centering
  \includegraphics[width=1.0\linewidth]{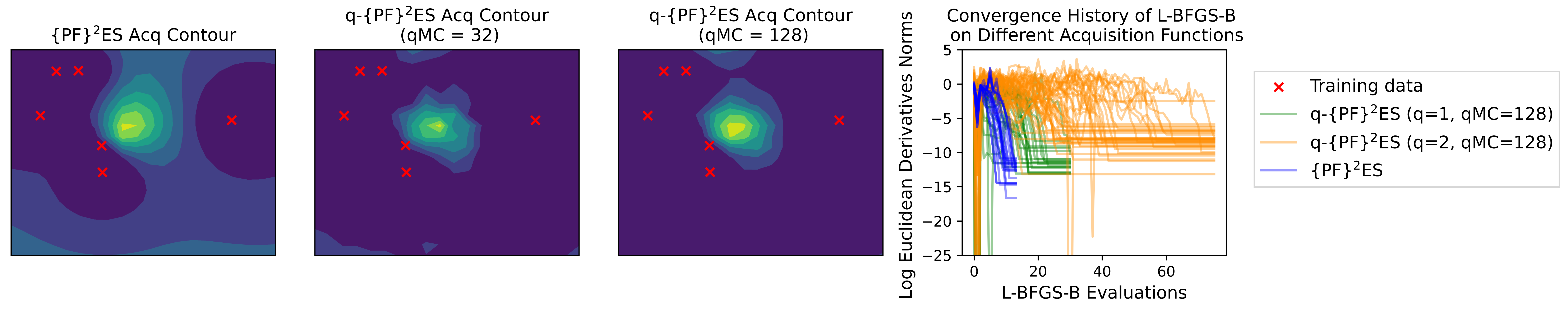}  
\caption{Comparison of \{PF\}$^2$ES and the Monte Carlo approximation of q-\{PF\}$^2$ES (with $q=1$) on the SRN benchmark function. The effect of the MC approximation on the gradient-based acquisition function optimizer can be seen in the last sub-figure: the multi-start gradient-based optimization generally requires larger function evaluations to converge (i.e., the norm of the derivatives reaches a lower value).}
\label{fig:Acq_opt_comparison}
\end{figure}

The following experiments are conducted in parallel per batch of ten on a Linux server with 256 GB RAM. The results are depicted in Table~\ref{tab: run_time}. In general, the joint batch acquisition function q-\{PF\}$^2$ES requires a longer time to query a batch of samples than \{PF\}$^2$ES. Besides the additional computational complexity introduced by the MC approximation, cfr. Section \ref{sec:computation_complexity}, this can be attributed to an increase in difficulty for the multi-start L-BFGS-B optimizer as illustrated in Fig.~\ref{fig:Acq_opt_comparison}. We also note that, though the \{PF\}$^2$ES-KB consumes similar time as other batch strategies, its computationally cost grows much faster w.r.t. $q$ as it requires additional sampling of the Pareto frontier $\tilde{\mathcal{F}}$. For PPESMOC, we generally observe a longer average query time than the original paper which reports the median (Table~1 of \cite{garrido2020parallel}), and the query time grows drastically with the number of iterations (e.g., the batch query time for C2DTLZ2 grows from around 20 minutes per batch sample to almost 3 hours when approaching the maximum number of batch iterations).

\begin{table*}[h]
\caption{Wall time for optimizing the acquisition function in seconds on a CPU (2x  Silver 4210 CPU @ 2.20GHz). The mean and 2 times the standard deviations are reported across 30 runs.
}
\begin{center}
\begin{tabular}{ l|l|l|l |l } 
\toprule
 & VLMOP2 & BraninCurrin &  ZDT1 & ZDT2 \\
 \midrule
 \{PF\}$^2$ES& $47.38 \pm 4.28$ & $47.01 \pm 2.58$  & $46.35 \pm 2.08$ & $45.86 \pm 2.23$  \\
 PFES & $63.68\pm 32.95 $ & $53.90 \pm 26.29$  & $84.02 \pm 63.6$ & $63.32 \pm 44.74$\\
 MESMO & $34.70 \pm 1.56$ & $34.85 \pm 1.50$  & $34.21 \pm 1.79$ & $34.00 \pm 1.64$\\
 PESMO &  $319.91 \pm 133.79$ & $321.83 \pm 135.60$ & $335.36 \pm 396.14$  & $432.59 \pm 334.89$\\
 EHVI & $0.29 \pm 1.19$ & $1.08 \pm 1.48$  & $0.43 \pm 1.39 $& $0.3226 \pm 1.35$\\
 \midrule
  q-\{PF\}$^2$ES (q=2)&  $113.96 \pm 85.54 $  & $128.08 \pm 115.16$& $132.23 \pm 215.71 $& $117.43 \pm 202.28$\\
  \{PF\}$^2$ES-KB (q=2)&  $89.96 \pm 79.00$ & $90.43 \pm 65.71$  & $94.80 \pm 80.17$& $94.80 \pm 80.17$ \\
    qEHVI (q=2)& $6.49 \pm 3.90 $ & $10.44 \pm 9.90$  & $4.98 \pm 9.83$ & $4.20 \pm 8.69$\\
 \midrule
 & C-BraninCurrin & Constr-Ex & SRN & C2DTLZ2\\
  \midrule
  \{PF\}$^2$ES& $63.05 \pm 4.58$ & $75.52 \pm 5.92$ & $75.81 \pm 6.39$ & $65.43 \pm 4.54$ \\
  EHVI-PoF& $7.46 \pm 2.9$ & $10.44 \pm 3.90$  & $8.47 \pm 3.32$ & $6.81 \pm 1.85$\\
  MESMOC& $46.41 \pm 2.59$ & $54.47 \pm 2.81$ & $52.22 \pm 2.93$ & $48.89 \pm 5.06$\\
 MESMOC+  & $58.52 \pm 38.62$ & $70.02 \pm 54.44$  & $70.78 \pm 72.39$ & $83.42 \pm 72.88$\\
   \midrule
PPESMOC (q = 2) & $2152.51 \pm 2981.10$ & $2735.67 \pm 3953.59$  & $2372.81 \pm 3791.25$ & $4276.52 \pm 7232.86$\\
  q-\{PF\}$^2$ES (q=2)& $186.72 \pm 177.68 $ & $259.27 \pm 167.52$  & $98.13 \pm 50.22$ & $217.71 \pm 217.48$\\
  \{PF\}$^2$ES-KB (q=2)& $170.56 \pm 126.00$ & $169.40 \pm 156.65$  & $196.04 \pm 160.60$ & $169.11 \pm 94.08$\\
    qEHVI (q=2)& $33.07 \pm 14.43 $ & $30.30 \pm 22.13$ & $35.57 \pm 14.25$ & $32.99 \pm 27.26$\\
\bottomrule
\end{tabular}
\end{center}
\label{tab: run_time}
\end{table*}

\begin{figure*}[!h]
\begin{subfigure}[t]{15cm}
  \centering
  \includegraphics[width=1.1\linewidth]{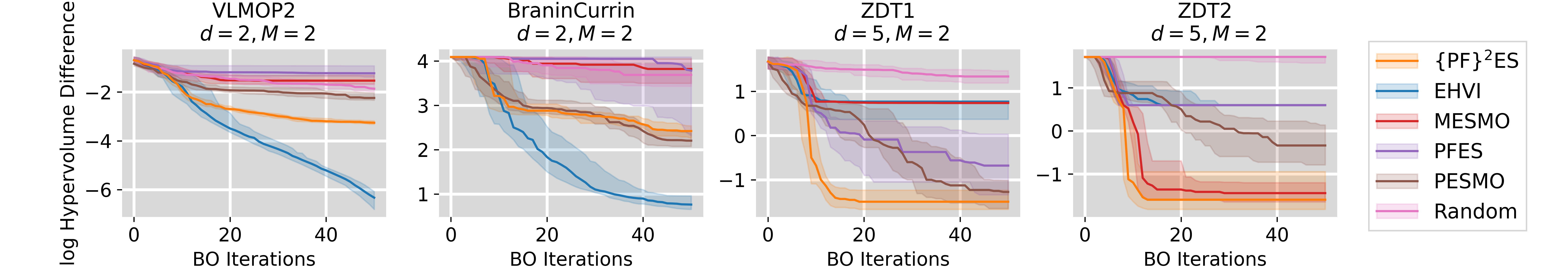}  
\end{subfigure}
\begin{subfigure}[t]{15cm}
\vspace*{-0.2cm}
  \centering
  \includegraphics[width=1.1\linewidth]{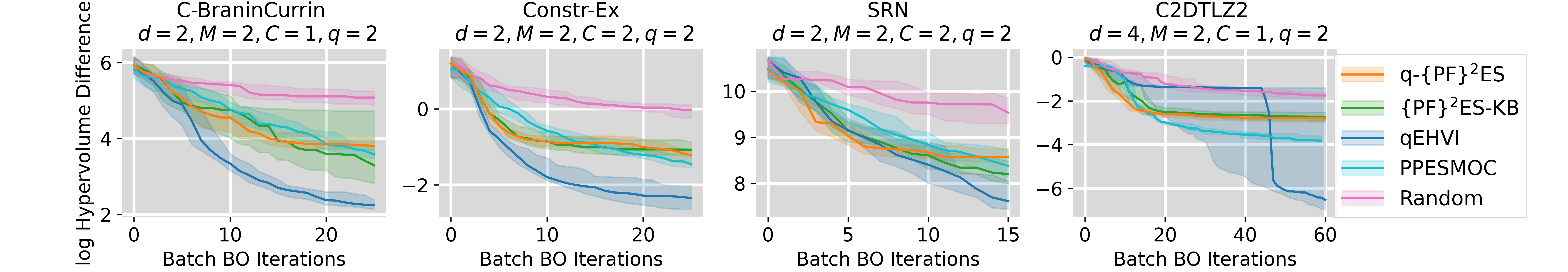}  
\end{subfigure}
\vspace*{-0.2cm}
\begin{subfigure}[t]{15cm}
  \centering
  \includegraphics[width=1.1\linewidth]{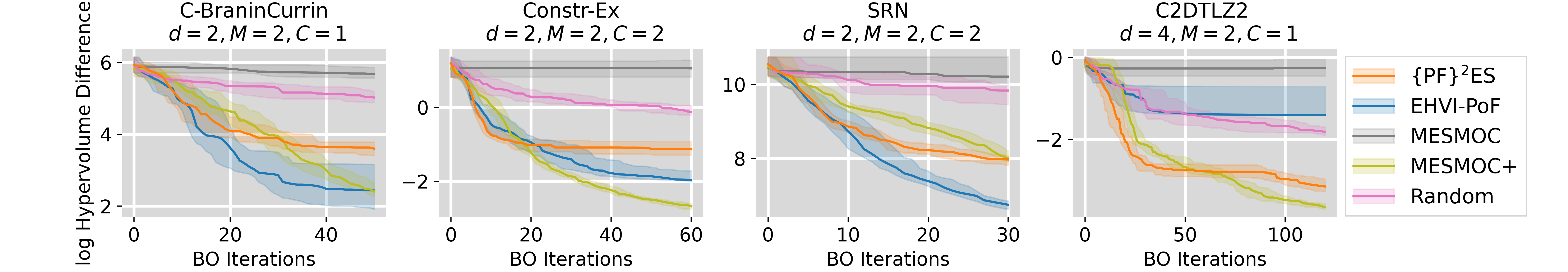}  
\end{subfigure}
\begin{subfigure}[t]{15cm}
\vspace*{-0.2cm}
  \centering
  \includegraphics[width=1.1\linewidth]{figures/Batch_CMOO_q2_is_res.png}  
\end{subfigure}
\caption{Comparison of different acquisition functions for multi-objective Bayesian optimization with respect to an in-sample recommendation strategy (In case of encountering a numerical issue (i.e., Cholesky decomposition issue), a different initial sample set is utilized.).}
\label{fig:in_sample_rec}
\end{figure*}

\section{ADDITIONAL EXPERIMENTAL RESULTS}\label{app:additional_exp_res}
\subsection{Experimental results for the in-sample recommendation strategy}\label{app:in_sample_res}

We report the experimental results of the in-sample recommendation strategy in the main paper, see Fig.~\ref{fig:in_sample_rec}, the in sampled Pareto frontiers are extracted from all of the data obtained so far in the current iteration. However, the in-sample recommendation strategy is not the intrinsic strategy for non-myopic information-theoretic acquisition functions.

\subsection{Experimental results for larger batch sizes}  \label{sec: 9.2}
\begin{figure}[!h]
  \centering
  \includegraphics[width=0.9\linewidth]{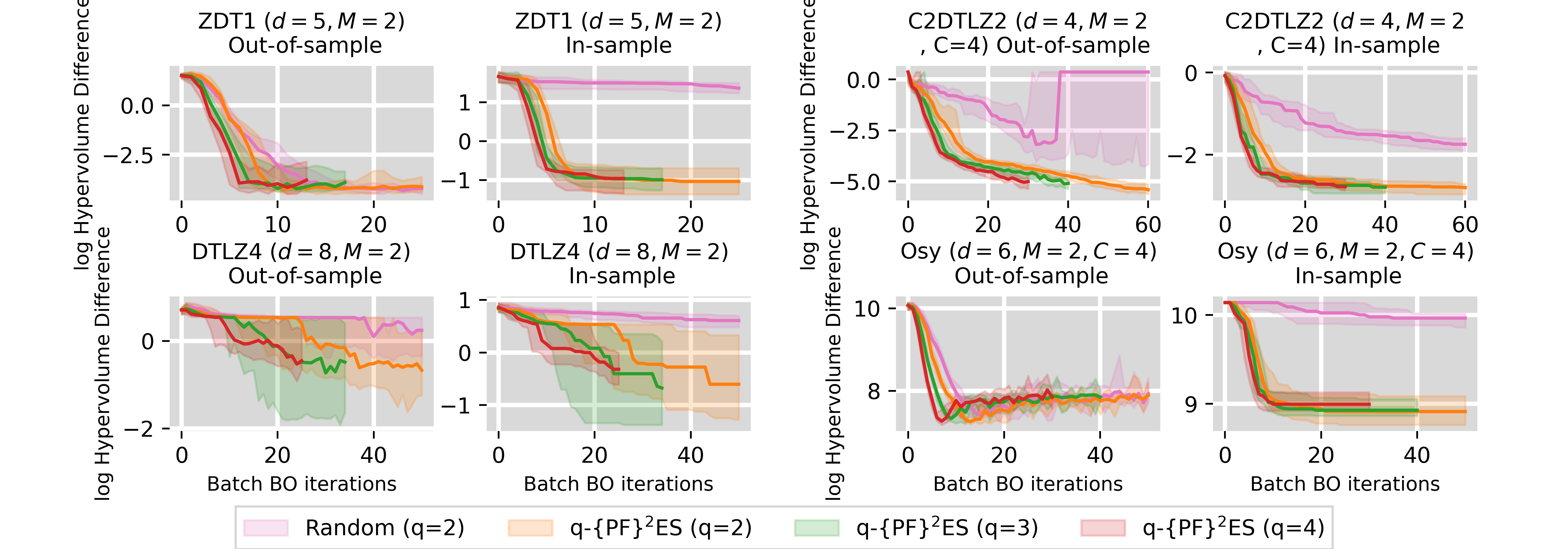}  
\caption{Numerical experiments of q-\{PF\}$^2$ES on larger batch sizes. Note all the batch methods terminate on roughly the same number of function evaluations.}
\label{fig:Acq_larger_batch_size}
\end{figure}

Additional experimental results for larger batch sizes $q$ is given in  Fig.~\ref{fig:Acq_larger_batch_size}. Besides the ZDT1 and C2DTLZ2 synthetic functions reported in the main paper, we also add the results of two new synthetic experiments, namely the DTLZ4 problem and the Osy problem, see Table~\ref{Tabular: reference_pts} for detailed settings. In general, with the same black box function evaluation budget, q-\{PF\}$^2$ES demonstrates similar performance. It is also expected that with an increase in batch size, the acquisition function performance can be degraded slightly due to the increasing complexity of the acquisition function optimization process.

\subsection{Experimental results for $M>2$ objectives} \label{sec: 9.3}
\begin{figure*}[!t]
\begin{subfigure}[t]{8.5cm}
  \centering
  \includegraphics[width=1.00\linewidth]{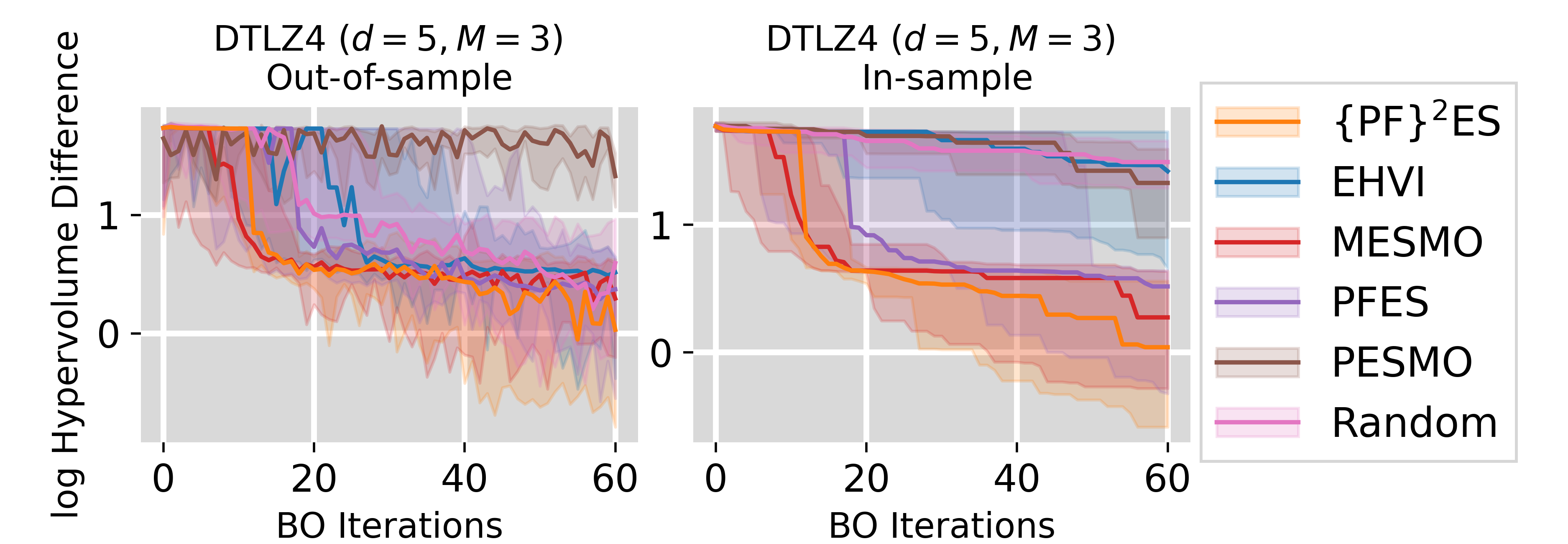}  
\end{subfigure}
\begin{subfigure}[t]{8.5cm}
  \centering
  \includegraphics[width=1.00\linewidth]{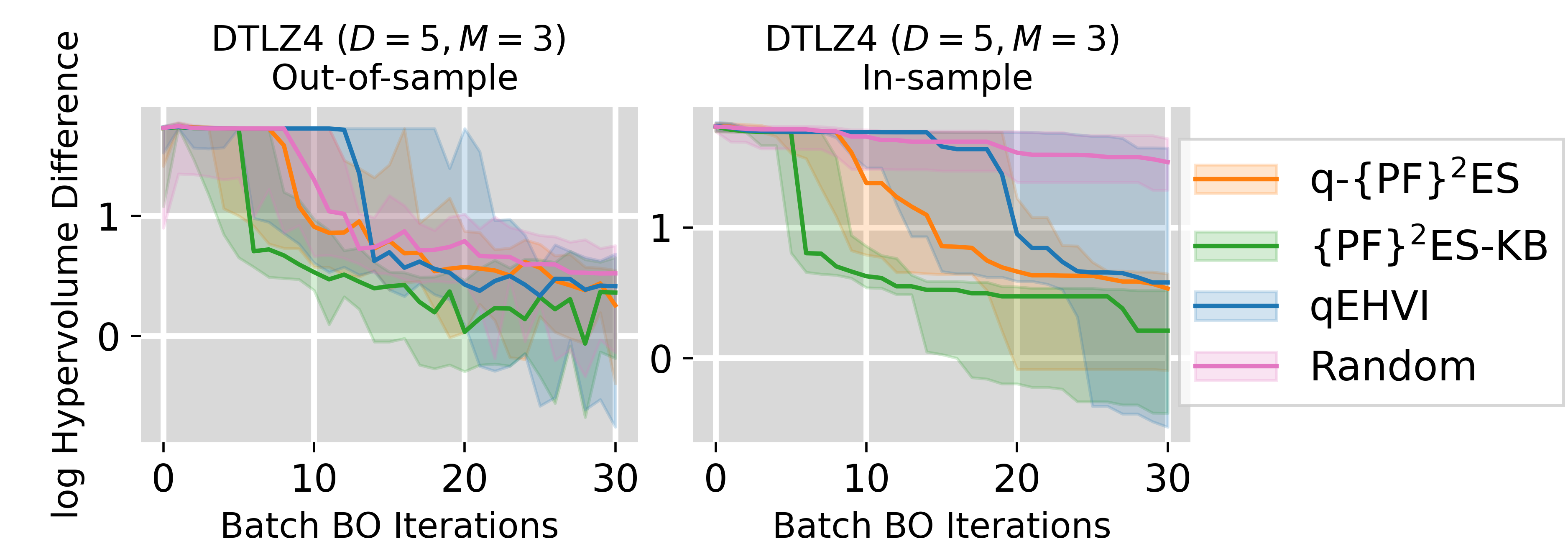}  
\end{subfigure}
\vspace*{-0.2cm}
\begin{subfigure}[t]{8.5cm}
  \centering
  \includegraphics[width=1.00\linewidth]{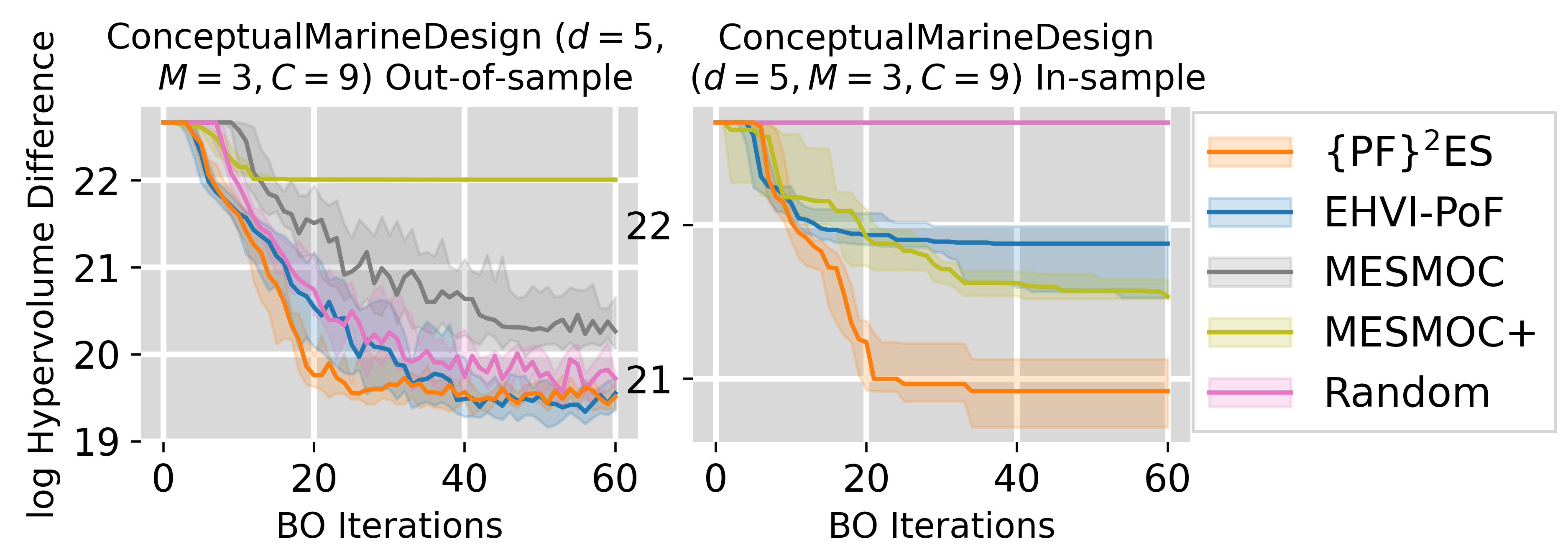}  
\end{subfigure}
\begin{subfigure}[t]{8.5cm}
  \centering
  \includegraphics[width=1.00\linewidth]{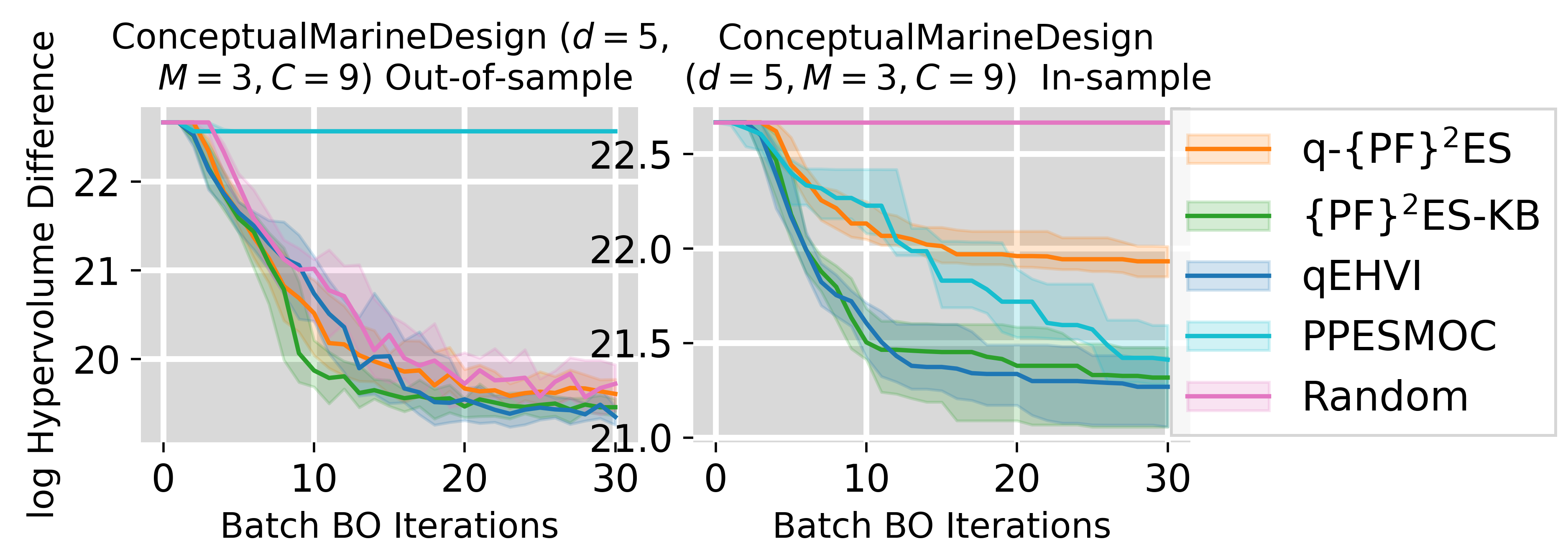}  
\end{subfigure}
\caption{Experimental comparison for larger output dimensionality.}
\vspace{128in}
\label{fig:bo_larger_m}
\end{figure*}

Two additional results for a larger number of objectives are seen in Fig.~\ref{fig:bo_larger_m} based on two new problems. For the unconstrained version, we use DTLZ4 with 3 objective functions. For the constraint case, we use a real-life conceptual Marine design problem \citep{parsons2004formulation, tanabe2020easy}. Details of the problem are also provided in Table~\ref{Tabular: reference_pts}. It can be observed that  \{PF\}$^2$ES and q-\{PF\}$^2$ES provide competitive performance. For the conceptual Marine design problem, we generally observe that \{PF\}$^2$ES-KB provides the fastest converge speed. We reason this is because given the relatively high constraint numbers, the approximated non-dominated feasible region $\tilde{A}$ is relatively small. For Monte Carlo-based acquisition functions like q-\{PF\}$^2$ES, this can impose difficulties for approximation and optimization since $Z_{\tilde{A}_{q}}$ (Eq.~\ref{eq: mc_batch_mopi}) is rather small. 

Note that, as discussed in Appendix.~\ref{sec:computation_complexity}, the computational complexity for providing the exact approximation of $A$ will increase exponentially with the number of objectives $M$, meaning that it will become the dominant factor in the calculation cost of \{PF\}$^2$ES as $M$ increases. This is a common problem for hypervolume partition strategy-based acquisition functions (e.g., EHVI, MOPI).


\vfill
\end{document}